\documentclass[acmlarge]{acmart}

\usepackage{multirow}
\usepackage{subcaption}
\usepackage{float}
\usepackage{enumitem}
\usepackage[table]{xcolor}

\definecolor{LightGreen}{rgb}{0.9,0.98,0.93}
\AtBeginDocument{%
  }

\setcopyright{none}
\copyrightyear{}
\acmYear{}
\acmDOI{}
\acmConference[]{}  
\acmVolume{}
\acmNumber{}
\acmArticle{}
\acmPrice{}
\acmISBN{}
\acmSubmissionID{}
\renewcommand{\acmArticleSeq}{} 

\acmISBN{}
\begin{document}

\title{MARAuder's Map: \underline{M}otion-\underline{A}ware \underline{R}eal-time \underline{A}ctivity Recognition with Layo\underline{u}t-Base\underline{d} Traj\underline{e}cto\underline{r}ies}

\author{Zishuai Liu}
\affiliation{%
  \institution{School of Computing, University of Georgia}
  \country{USA}
}
\author{Weihang You}
\affiliation{%
  \institution{School of Computing, University of Georgia}
  \country{USA}
}
\author{Jin Lu}
\affiliation{%
  \institution{School of Computing, University of Georgia}
  \country{USA}
}
\author{Fei Dou}
\affiliation{%
  \institution{School of Computing, University of Georgia}
  \country{USA}
}

\begin{abstract}
Ambient sensor-based human activity recognition (HAR) in smart homes remains challenging due to the need for real-time inference, spatially grounded reasoning, and context-aware temporal modeling. Existing approaches often rely on pre-segmented, within-activity data and overlook the physical layout of the environment, limiting their robustness in continuous, real-world deployments.
In this paper, we propose MARAuder’s Map, a novel framework for real-time activity recognition from raw, unsegmented sensor streams. Our method projects sensor activations onto the physical floorplan to generate trajectory-aware, image-like sequences that capture the spatial flow of human movement. These representations are processed by a hybrid deep learning model that jointly captures spatial structure and temporal dependencies. To  enhance temporal awareness, we introduce a learnable time embedding module that encodes contextual cues such as hour-of-day and day-of-week. Additionally, an attention-based encoder selectively focuses on informative segments within each observation window, enabling accurate recognition even under cross-activity transitions and temporal ambiguity.  Extensive experiments on multiple real-world smart home datasets demonstrate that our method outperforms strong baselines, offering a practical solution for real-time HAR in ambient sensor environments.
\end{abstract}


\renewcommand{\footnotetextcopyrightpermission}[1]{
  \footnotetext{This manuscript is under review.}
}

\maketitle
\thispagestyle{empty}
\pagestyle{empty}
\section{Introduction}
Human Activity Recognition (HAR) in smart home environments has become a cornerstone for enabling a wide range of human-centric applications, including healthcare monitoring \cite{islam2020development, enshaeifar2018health, chen2023digital}, elder care \cite{riboni2016smartfaber, do2018rish}, and ambient assisted living (AAL) \cite{guerra2023ambient, zakka2024action, bari2024advancements}. By leveraging ambient sensors, such as motion detectors, door contacts, and environmental indicators like temperature and pressure sensors, HAR system unobtrusively monitor activities of daily living (ADLs), including cooking, medication intake, personal hygiene, and housekeeping, without compromising user comfort and privacy or requiring active user participation\cite{tewell2019monitoring}. In contrast to camera-based or wearable solutions, ambient sensors offer a unique combination of non-intrusiveness, privacy preservation, and minimal maintenance requirements, making them particularly suitable for long-term monitoring while preserving residents’ comfort and autonomy \cite{mustafa2021iot}. With the rising demand for aging-in-place solutions and remote health monitoring, particularly in light of global demographic shifts and an aging population \cite{kanasi2016aging}, the need for accurate and robust HAR systems is more urgent than ever. Recent advances in deep learning have significantly improved recognition performance by capturing dependencies in ambient sensor systems \cite{ghadi2022improving, bouchabou2021survey, bouchabou2021using, li2019relation,liciotti2020sequential}.



Nevertheless, despite recent progress, significant challenges persist in deploying ADL recognition systems effectively in real-world environments. In practice, sensor logs are continuous, noisy, spatially structured, and temporally patterned-conditions that complicate recognition and are often overlooked in prior work.  These characteristics reveal critical limitations in current approaches, particularly in their ability to handle real-time data streams, manage cross-activity transitions, leverage physical layout information, and incorporate rich temporal context reflective of daily routines.



\textbf{Real-time recognition and cross-activity window ambiguity.} Most prior studies assume access to pre-segmented activity data, where "begin" and "end" annotations clearly mark the boundaries of each activity instance \cite{bouchabou2021using, liciotti2020sequential, kiranyaz20211d, gochoo2018unobtrusive}. Under this assumption, models are then trained and evaluated on clean activity segments, often divided into sliding windows or treated as whole activity instances, which are presumed to contain a single, or complete activity. As a result, existing approaches primarily focus on \textbf{within-activity recognition}, where the model's task is to classify behavior given a well-defined activity window. However, in real-time smart home deployments, such segmentation is rarely feasible \cite{yala2015feature, demongivert2021handling}. Instead, sensor data arrive as a continous, asynchronous stream without explicit activity boundaries. Consequently, a single observation window may span multiple activities or include partial activity fragments, creating what we refer to as the \textbf{cross-activity window ambiguity}. This ambiguity introduces significant complexity into the recognition process, as models must make predictions without knowing whether an activity has started, ended, or is in transition. While very few early studies have attempted to address this problem using traditional probabilistic models such as Hidden Markov Models (HMMs) and Bayesian approaches \cite{yala2015feature, yan2016real}, these methods typically depend on handcrafted features and lack the flexibility to generalize across divers sensor environments. In contrast, deep learning models offer the capacity to learn  complex, hierarchical representations directly from raw sensor streams, making them better suited for handling cross-activity transitions and real-time inference under continuous monitoring conditions.

\textbf{Underutilization of spatial context.} Despite the inherently spatial nature of human behavior, many ADL recognition approaches overlook or inadequately represent the physical layout of smart homes. A common approach is to treat sensor IDs as independent categorical variables, ignoring their spatial relationships and deployment positions  \cite{liciotti2020sequential}. Some methods attempt to impose artificial spatial structures by manually arranging sensors in a 1D space-for example, by aligning nearby sensors adjacently along the y-axis \cite{gochoo2018unobtrusive} through adjusted coordinates-while others model inter-sensor dependencies using graphs or encode room information using textual embeddings \cite{thukral2025layout, chen2024towards, plotz2024using}. However, these techniques often provide only coarse approximations and fail to capture the spatial continuity and flow of human movement across the environment. A principled layout-aware representation—one that explicitly incorporates floorplan and sensor placements, and models sensor activations as trajectories over space—is largely absent in current work. Such spatial encoding is essential for enabling models like CNNs to learn spatially grounded  patterns associated with human daily routines, and for improving both recognition accuracy and interpretability in real-world deployments.

\textbf{Neglect of time-sensitive contextual cues.} Although time plays a central role in human behavior, temporal information remains underutilized in many HAR approaches. While some studies incorporate basic features such as weekday indicators or coarse time-of-day bins\cite{chen2024towards, civitarese2024large}, most rely soley on the order of sensor events to capture temporal context\cite{liciotti2020sequential, gochoo2018unobtrusive, mohmed2020employing}. This limited perspective overlooks the rich temporal patterns in daily routines, such as consistent patterns tied to specific days of the week, hours, or broader periods like mornings, afternoons, and evenings. Fine-grained temporal encoding is particularly important to disambiguate activities that occur in the same spatial region but differ in temporal context (e.g., taking medication in the morning versus at night). Without modeling these recurring temporal structures, HAR systems are prone to missing critical  context cues, ultimately reducing their accuracy and reliability in real-world settings.

\begin{figure}[htbp]
    \centering
    \begin{subfigure}[b]{0.3\textwidth}
        \centering
        \includegraphics[height=0.25\textheight]{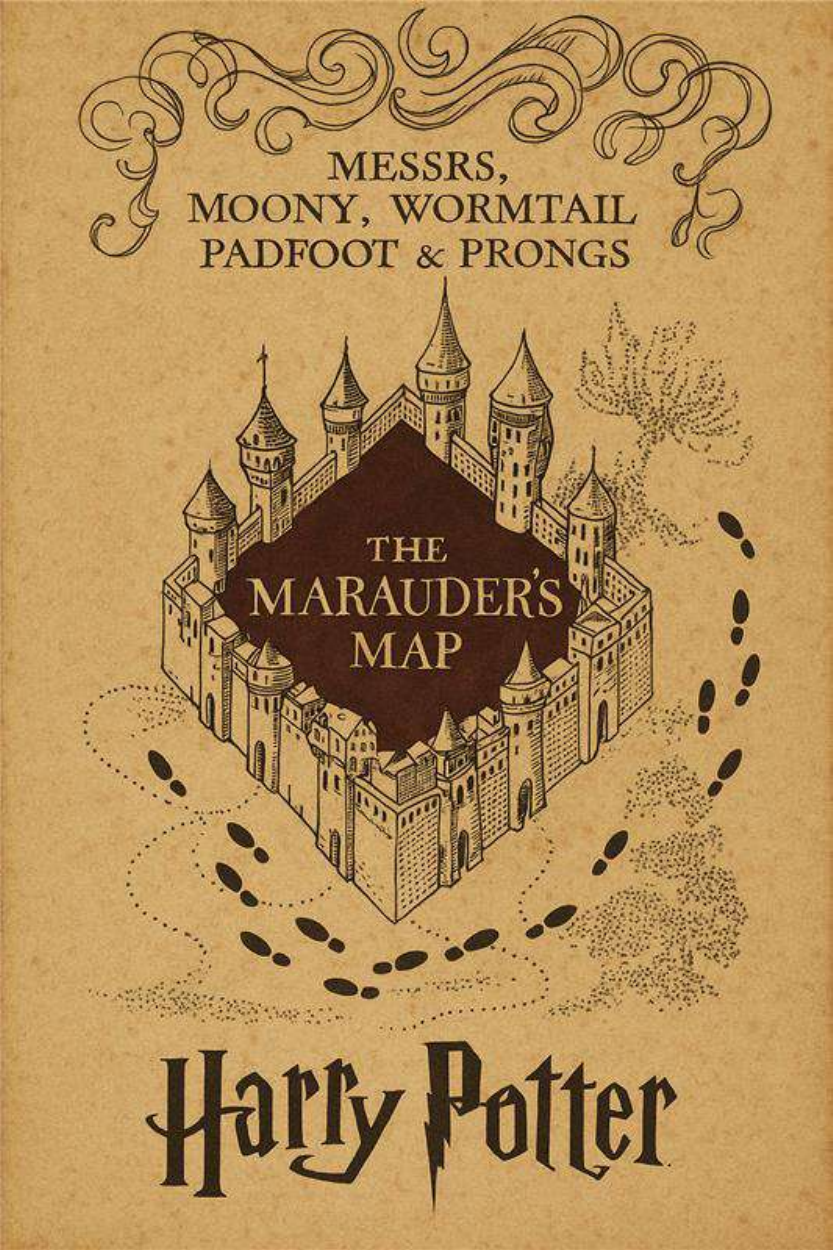}
        \caption{Marauder's Map in Harry Potter}
        \label{fig:marauds}
    \end{subfigure}
    \hspace{0.02\linewidth}
    \begin{subfigure}[b]{0.5\textwidth}
        \centering
        \includegraphics[height=0.25\textheight]{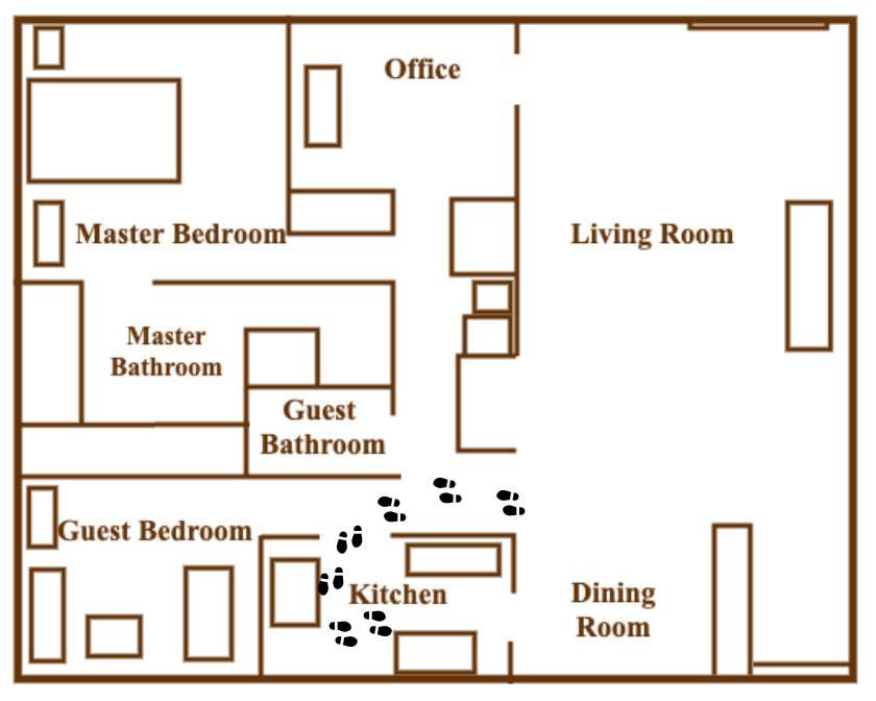}
        \caption{Marauder's Map in HAR}
        \label{fig:maradsh}
    \end{subfigure}
    \caption{Illustration of the inspiration behind our proposed Marauder’s Map framework. (a) The original Marauder’s Map from Harry Potter, which visualizes real-time movement within a physical space \cite{chatgpt_image2025}. (b) Our adaptation for Human Activity Recognition (HAR), depicting a floorplan with a trajectory example in the kitchen area.}
    \label{fig:maraudermap}
\end{figure}

We introduce MARAuder’s Map (\underline{M}otion-\underline{A}ware \underline{R}eal-time \underline{A}ctivity recognition with layo\underline{u}t-base\underline{d} traj\underline{e}cto\underline{r}ies)-a novel framework for real-time activity recognition from raw, unsegmented ambient sensor streams in smart home environments. Inspired by the magical Marauder’s Map from the Harry Potter series as shown in Fig~\ref{fig:maraudermap}, our approach directly addresses key limitations of existing HAR systems by integrating motion-aware modeling through spatial layout encoding, cross activity attention for ambiguous observation windows, and time-sensitive context to capture temporal variations in daily routines. The key contributions of this work are summarized as follows:
\begin{itemize}
    \item \textbf{Real-time recognition from unsegmented, cross-activity streams.} We explicitly move beyond the pre-segmented setting dominant in prior work, which assumes clean activity boundaries and within-activity recognition. Our framework is designed to operate on continuous sensor streams where activity transitions are gradual and boundaries are unknown, enabling real-time prediction even when observation windows contain multiple or incomplete activities. To support recognition under these challenging conditions, we incorporate an attention-based sequential encoder that selectively focuses on the most informative segments within each window, allowing accurate inference without relying on explicit activity boundary information.
    \item \textbf{Layout-aware spatial encoding:} We propose a layout-based trajectory representation that projects sensor activation sequences onto the physical floorplan of the smart home. In contrast to previous methods that rely on non-layout-based visualizations or abstract feature sequences, our approach generates trajectory-aware, image-like sequences that more accurately capture the real-time spatial flow of movement and enable models to learn patterns grounded in the physical context of the environment.
    \item \textbf{Contextual temporal embedding:} We introduce a temporal embedding module that encodes contextual cues such as hour-of-day and day-of-week, to capture time-sensitive behavioral variations. This allows the model to disambiguate activities that occur in the same spatial regions but differ by temporal context (e.g., morning vs. evening routines). 
\end{itemize}
\section{Motivation}
Traditional sliding window approaches often assume that each window aligns precisely with a single labeled activity, simplifying the learning task by providing clean, isolated segments. For example, a "prepare meal" activity may be captured with a window consisting of 30 log data that exactly starts when the user enters the kitchen and ends when cooking stops, with no other activity mixed in. This assumption works well for curated datasets but rarely holds in practical deployment. In smart homes, sensor streams would be collected continuously and passively, without human supervision, leading to messier data structures. In real environments, residents naturally multitask, switch between activities, and perform tasks in overlapping or fragmented ways. For instance, a resident might start reading (activity A), go to check the laundry (activity B), and then return to finish reading. In our setting, we segment the continuous sensor stream into fixed-size windows (e.g., 60 events with a step size of 6) to emulate real-time recognition. However, these windows rarely align with true activity start and end points \cite{yan2016real}, resulting in mixed or partial activity segments. We refer to this as \textbf{cross-activity window ambiguity}, where a single window may span multiple transitions or fragmented behaviors. This label ambiguity complicates learning and calls for models that are robust to noisy, composite inputs.
\begin{figure}[h!]
  \centering
   \includegraphics[width=\linewidth]{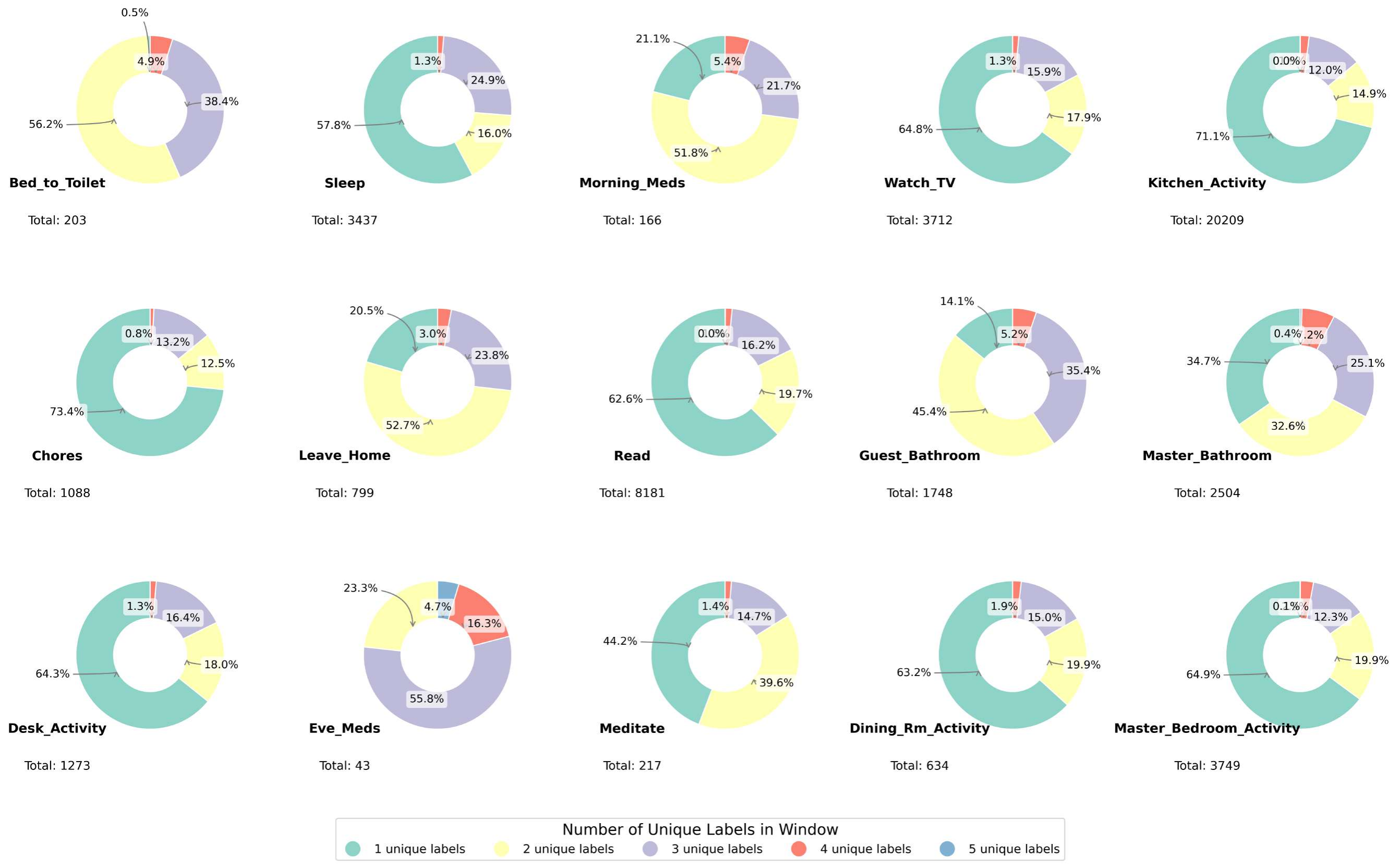}
   \caption{Distribution of the number of unique activity labels contained within each sliding window for all activity classes in the Milan dataset (window size = 60, step size = 6). Each donut chart represents a specific activity and shows the proportion of windows containing 1 to 5 distinct activity labels. Higher proportions of multi-label windows indicate greater label ambiguity, especially for activities like Bed\_to\_Toilet, Leave\_Home, and Eve\_Meds.}
   \label{fig:lable_unique}
\end{figure}

Fig.\ref{fig:lable_unique} visualizes the extent of label mixing within these windows across different activity classes. Each donut chart corresponds to a primary activity label (e.g., Bed\_to\_Toilet, Sleep, Morning\_Meds) and shows the proportion of windows that also include additional unique labels. The colored segments indicate the number of extra labels present, ranging from zero to four. Notably, the presence of substantial yellow, purple, red, and blue segments across various charts reflects a non-trivial frequency of label co-occurrence within a single window. For example, while a large portion of windows labeled as Sleep contain only that activity (light green segment), a meaningful fraction includes other activities, suggesting these windows often span transitions into or out of sleep. Similarly, activities such as Bed\_to\_Toilet and Leave\_Home frequently appear alongside other labels, indicating a high degree of behavioral interleaving within the window size used.

This activity co-occurrence in one window introduces label ambiguity, which poses a significant obstacle for learning clean, discriminative features. When a window contains mixed activity segments, the model may extract misleading representations, ultimately degrading classification performance. These findings highlight the importance of careful windowing design and suggest the potential value of alternative segmentation strategies or models that can handle noisy labels. The analysis in Fig.\ref{fig:lable_unique} provides a clear depiction of the complexity introduced by the fixed-window approach and underscores the need for robust modeling techniques in the presence of temporal label ambiguity.

\begin{figure}[h!]
    \centering
    \begin{subfigure}[b]{0.25\textwidth}
        \centering
        \includegraphics[width=\textwidth]{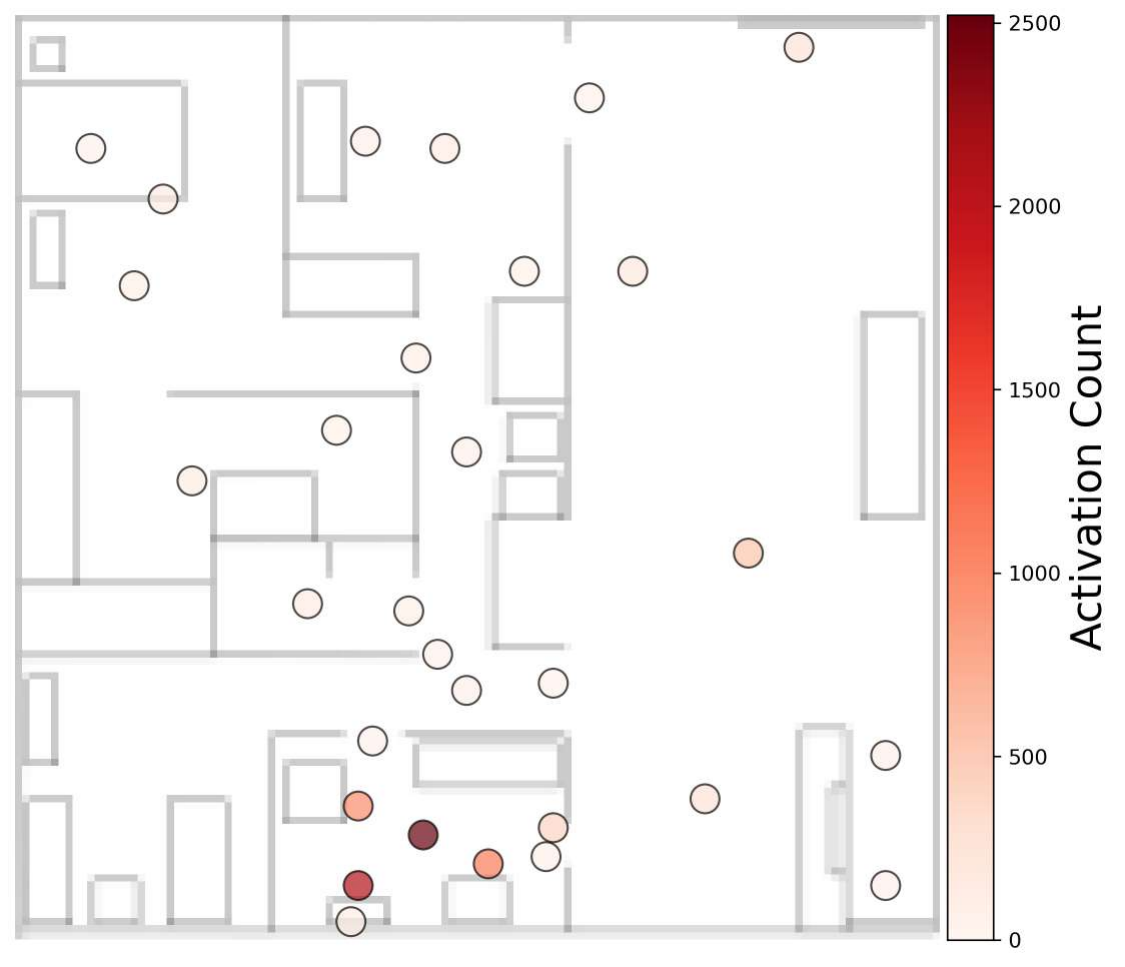}
        \caption{Morning Meds}
        \label{fig:mornmeds}
    \end{subfigure}
    \hspace{0.02\linewidth}
    \begin{subfigure}[b]{0.26\textwidth}
        \centering
        \includegraphics[width=\textwidth]{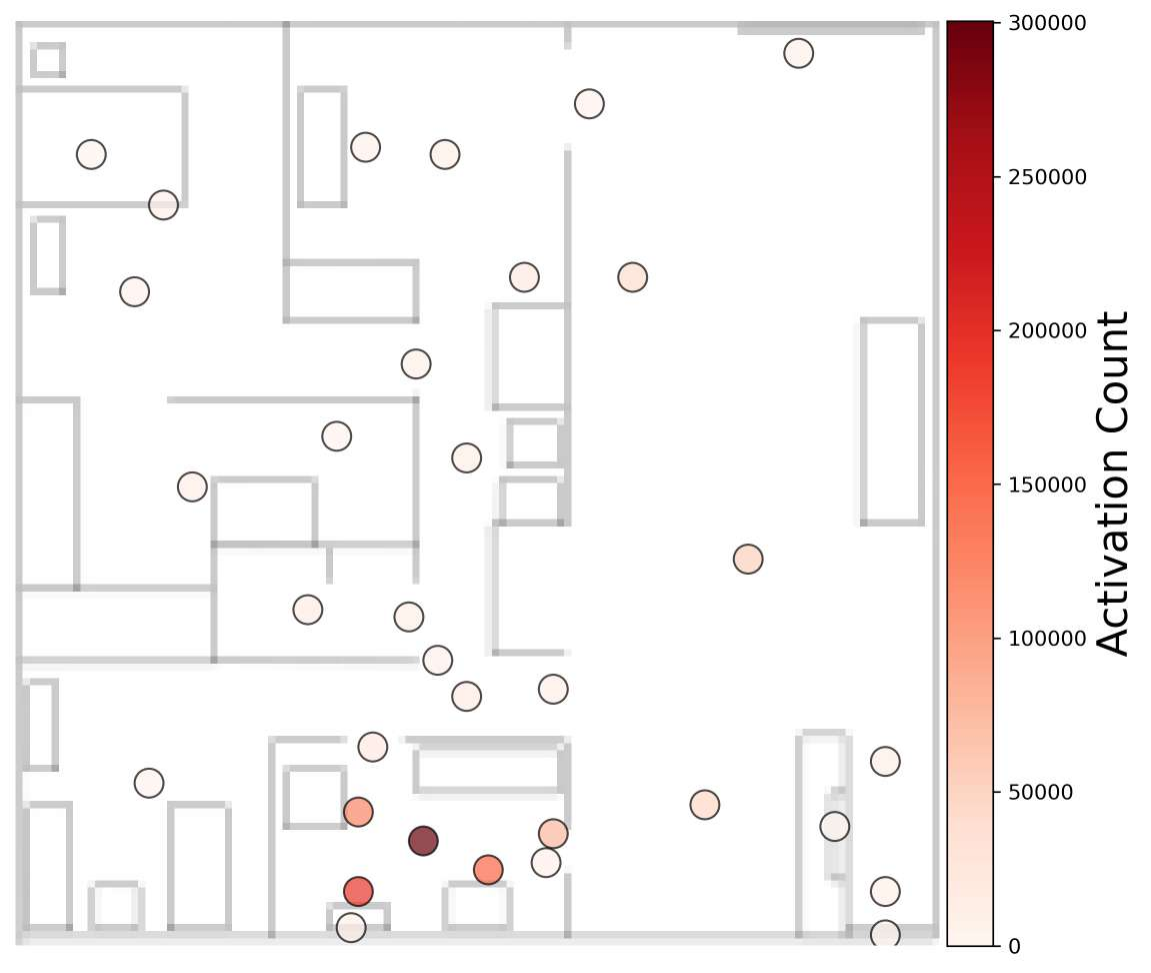}
        \caption{Kitchen Activity}
        \label{fig:kitchacti}
    \end{subfigure}
    \hspace{0.02\linewidth}
    \begin{subfigure}[b]{0.25\textwidth}
        \centering
        \includegraphics[width=\textwidth]{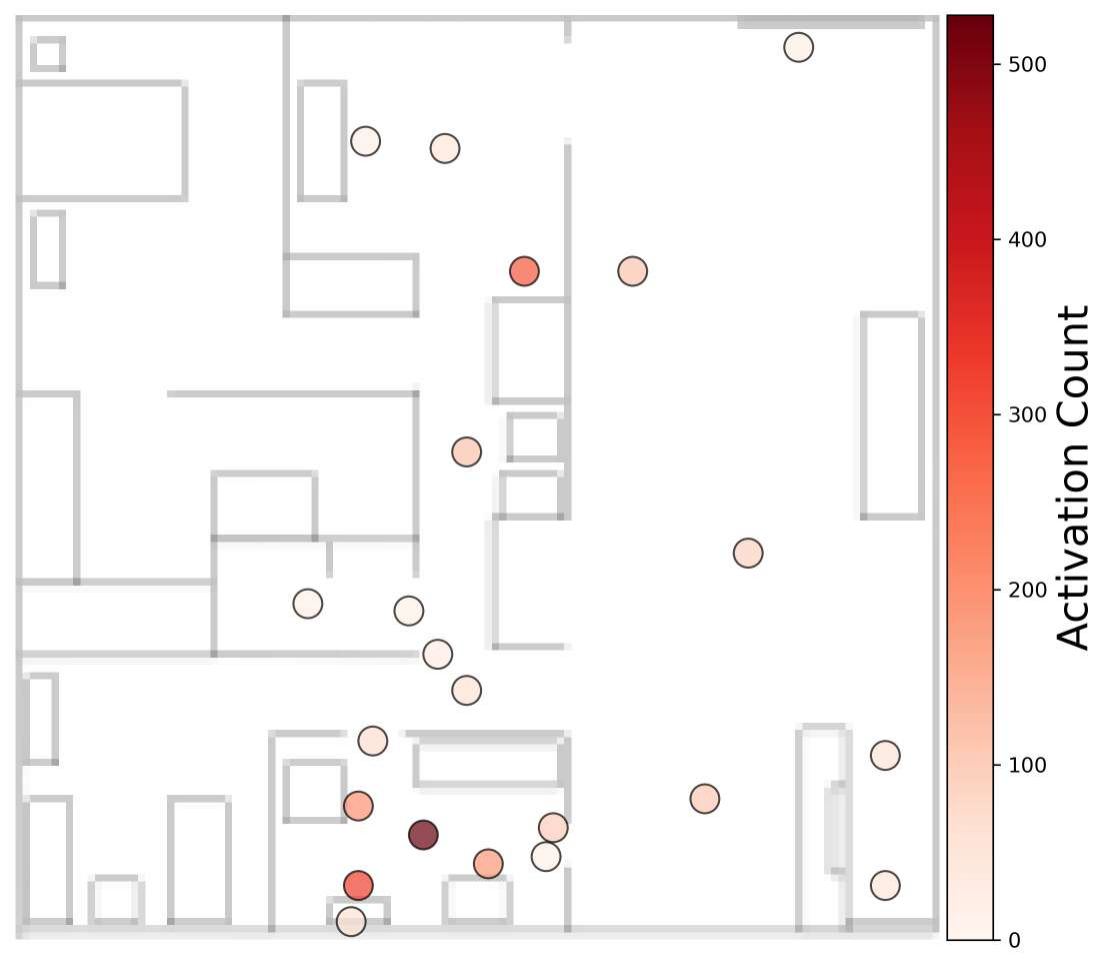}
        \caption{Eve Meds}
        \label{fig:evemed}
    \end{subfigure}
    \caption{Spatial heatmaps illustrating the distribution and intensity of sensor activations during three specific activities: (a) Morning\_Meds, (b) Kitchen\_Activity, and (c) Eve\_Meds. The color intensity of each circle represent the frequency of sensor triggers at corresponding locations, highlighting patterns of movement and activity hotspots within the home environment.}
    \label{fig:heatmaps}
\end{figure}

Besides, another fundamental challenge in accurately classifying human activities within smart home environments arises from the fact that sensor activations are often not uniquely tied to a single activity. As illustrated in Fig.\ref{fig:heatmaps}, which presents heatmaps of sensor trigger frequencies during three distinct labeled activities – Morning Meds, Kitchen Activity, and Eve Meds – a significant degree of spatial overlap in sensor activation patterns is evident, particularly within the kitchen area. Notably, all three activities exhibit a concentration of sensor triggers in this region, suggesting that the sensors located in or around the kitchen are frequently activated during medication intake (both morning and evening) as well as during general kitchen activity. This shared trajectory of sensor activations makes it inherently difficult for classification algorithms to effectively discriminate between these seemingly related yet distinct activities based solely on the patterns of sensor firings. The ambiguity introduced by such overlapping sensor signatures underscores the need for sophisticated feature engineering and advanced machine learning models capable of discerning subtle temporal or contextual cues that might differentiate these activity groups.

\begin{figure}[h!]
    \centering
    \includegraphics[width=0.50\linewidth]{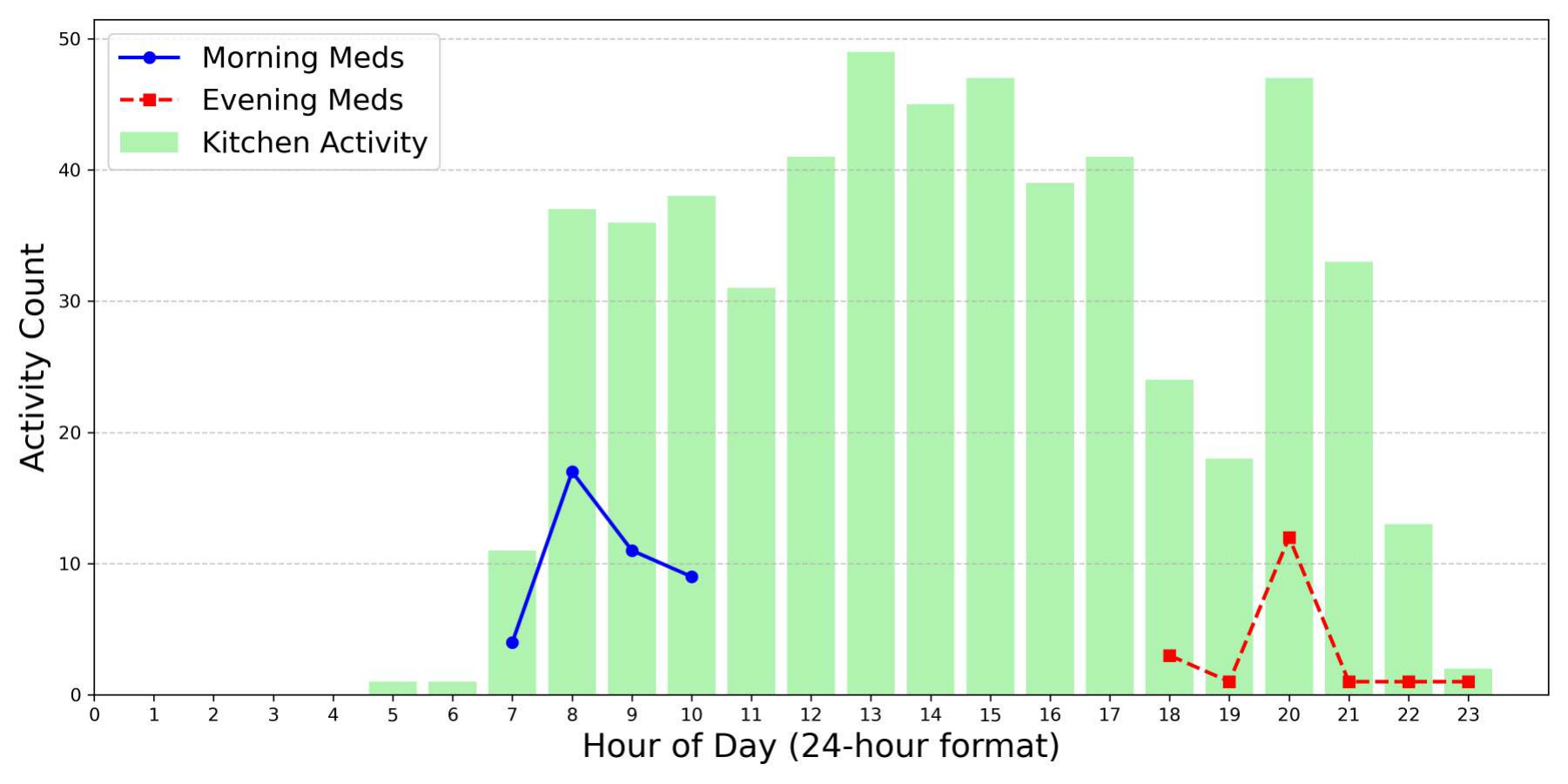}
    \caption{Hourly distribution of activity start times for Morning\_Meds, Evening\_Meds, and Kitchen\_Activity. Line plots represent the start times of medication-related activities, highlighting clear temporal boundaries—Morning\_Meds typically begins between 7–10 AM, while Evening\_Meds starts around 7–9 PM. The green bars show the distribution of Kitchen\_Activity start times, which occur throughout the day but peak around midday and early evening.}
    \label{fig:kitchhour}
\end{figure}

To resolve such spatial ambiguity, we argue that temporal context is critical.
Fig.\ref{fig:kitchhour} illustrates the start-time distributions for Morning Meds, Evening Meds, and Kitchen Activity. The histograms reveal distinct temporal patterns that provide critical discriminative cues, especially when spatial trajectories overlap, as in the kitchen area shown in Fig.\ref{fig:heatmaps}. Morning Meds (blue) and Evening Meds (red) exhibit strong temporal regularity. The former is concentrated in the early morning (7–10 AM), while the latter peaks in the evening (7–8 PM). In contrast, Kitchen Activity (green) is distributed across the day, with increased frequency during typical meal times, including late morning, midday, and early evening. These differences in temporal profiles highlight the value of encoding time-related features—such as hour-of-day or day-of-week—into the model. Temporal context can help disambiguate activities with similar spatial patterns. For example, even if Morning Meds and Kitchen Activity share sensor activations, knowing that an event starts during early morning hours increases the likelihood that it corresponds to Morning Meds.

Together, these analyses motivate the core design of our framework. We propose a model that can handle cross-activity window ambiguity, encodes layout-aware spatial trajectories, and integrates contextual time features such as hour-of-day and day-of-week. This spatiotemporal integration allows the model to reason beyond rigid segmentation boundaries and improve real-time activity recognition in complex, naturalistic settings.

\section{Related Work}
Activity of Daily Living Recognition in smart home environments typically involves two essential components: preprocessing and representation of raw sensor data, followed by the design of effective models for activity classification\cite{thukral2025layout}. The first stage focuses on transforming continuous sensor logs—often noisy, irregular, and diverse—into structured, machine-interpretable formats. The second stage concentrates on constructing models capable of learning discriminative patterns from the processed data. These models range from traditional machine learning classifiers using handcrafted features to advanced deep learning architectures that capture temporal dynamics, spatial dependencies, or multi-modal information. In the following sections, we first review preprocessing and representation strategies, then discuss the development of classification models, and finally examine recent approaches that embed spatial information, temporal information, or a combination of both to enhance activity recognition performance.
\subsection{User Behavior Processing through Ambient Sensor Logs}
In the segmentation phase of the ADL recognition, windowing techniques play a crucial role in partitioning continuous sensor streams into meaningful segments for subsequent feature extraction and classification\cite{bouchabou2021survey, najeh2022dynamic}. Commonly adopted methods include Explicit Windowing, Time Windowing, Sensor Event Windowing, and Dynamic Windowing \cite{quigley2018comparative}. Explicit Windowing (EW) follows a two-step process by first segmenting data into activity-specific portions before classification. However, variability in activity duration across individuals and the sequential dependency between steps can limit its effectiveness \cite{krishnan2014activity}. Time Windowing (TW) segments data into fixed-duration intervals, making it suitable for continuous streams such as accelerometer data \cite{medina2018ensemble, hamad2021dilated}. Yet, this method often struggles with binary sensor data in smart homes, where sparse or overlapping events challenge the temporal window’s representativeness \cite{banos2014window}. Fuzzy Time Windows (FTW) were introduced by Medina et al. \cite{medina2018ensemble} to enhance temporal encoding for multi-variate binary sensor sequences. Unlike fixed-size windows, FTWs employ a trapezoidal-shaped segmentation of sensor streams that captures short, medium, and long-term dependencies in a time interval \cite{hwang2021deep, hamad2019efficient}. Sensor Event Windowing (SEW) divides data based on a fixed number of sensor activations, enabling variable temporal granularity. While adaptive to sensor frequency, it may fail to provide adequate context during periods of low or high activity density \cite{krishnan2014activity}. Dynamic Windowing (DW) adjusts window sizes based on predefined rules to better capture activity boundaries. Although potentially more accurate, DW requires reliable annotations and may face challenges in modeling complex or overlapping activities \cite{al2016windowing, al2016improving, al2017activity}. These windowing strategies each present trade-offs that must be carefully considered depending on the characteristics of the dataset and target application.

\subsection{Classification Models for Smart Home Activity Recognition}
Early work on activity recognition often relied on rule-based or knowledge-driven methods. These systems encoded expert knowledge or heuristics about human behavior, sensor layouts, or event sequences to manually define activity patterns. For instance, logic-based models and handcrafted temporal rules were used to infer activities from sequences of sensor events \cite{chen2008logical, yamada2007applying}. While interpretable and relatively lightweight, these approaches struggled with scalability, adaptability to new environments, and robustness to noisy activities \cite{bouchabou2021survey}.

To overcome the limitations of knowledge-based systems, researchers turned to data-driven machine learning techniques. Methods such as Support Vector Machines (SVMs)\cite{yala2015feature}, decision trees, and probabilistic models like Hidden Markov Models (HMMs)\cite{asghari2020online} were widely applied. These methods learned patterns from labeled data and handled temporal dependencies more flexibly. Hybrid models further improved performance by combining probabilistic reasoning with feature extraction stages informed by latent structures, such as the use of Latent Dirichlet Allocation (LDA) to learn activity-feature patterns offline and feed them into online classifiers~\cite{yan2016real}. Recurrent Neural Networks (RNNs), Long Short-Term Memory (LSTM) networks, and attention-based architectures have demonstrated strong performance on sequence classification tasks. Liciotti et al. \cite{liciotti2020sequential} demonstrated that LSTM models surpass traditional ADL classification methods by automatically learning temporal features without relying on handcrafted features. Similarly, studies such as \cite{singh2017human} and \cite{alshammari2018evaluating} confirmed that LSTM outperforms algorithms like HMMs, AdaBoost, and multilayer perceptrons in classification accuracy. Besides, Park et al.\cite{park2018deep}, integrated residual connections and an attention mechanism into a multi-layer LSTM model, improving gradient flow and highlighting salient events in time series. 

Gochoo et al. \cite{gochoo2018unobtrusive} introduced a binary image encoding approach that maps sensor activations into spatial formats suitable for 2D CNN processing. This was later extended in their subsequent work \cite{tan2018multi}, which incorporated colored pixel values to encode richer sensor modalities, including continuous values such as temperature and inter-sensor relationships. Similarly, Mohmed et al. \cite{mohmed2020employing} employed grayscale images where pixel intensity reflects the duration of sensor activation, and utilized AlexNet for feature extraction, followed by conventional classifiers for activity prediction. These methods demonstrate the viability of visual representations in capturing spatial-temporal patterns in smart home environments. Singh et al. \cite{singh2017human} applied 1D CNN architectures on unprocessed sequences, showing comparable performance to traditional CNN approaches while preserving the temporal resolution of events. More recently, Thukral et al. \cite{thukral2025layout} introduced TDOST, a layout-agnostic framework for HAR that encodes sensor readings into textual descriptions enriched with contextual information, such as coarse-grained layout and sensor attributes. By leveraging large pre-trained language models, TDOST enables transferable activity recognition across different smart home environments with minimal prior knowledge.

\subsection{Review of Spatial and Temporal Embedding Techniques in ADL Recognition}
Effectively modeling spatial and temporal information is critical for accurate activity recognition in smart home environments. However, prior studies vary significantly in how they incorporate these two dimensions.

For spatial information, early works such as \cite{liciotti2020sequential} and \cite{oguntala2021passive} ignored explicit spatial layout information entirely, treating sensor activations as purely sequential events. Other methods, such as in \cite{gochoo2018unobtrusive, mohmed2020employing, arrotta2022dexar}, attempted to encode implicit spatial relations by mapping sensor IDs along the y-axis of a matrix based on physical proximity—assigning adjacent coordinates to sensors in nearby rooms and larger gaps to distant ones. While this captures some neighborhood relationships, it fails to represent full resident trajectories across the environment. Graph-based models, such as \cite{ye2023graph, plotz2024using}, constructed sensor graphs based on inferred interaction dependencies, capturing functional connections like those between kitchen appliances and motion sensors. More recently, \cite{chen2024towards, thukral2025layout} explored using large language models to embed room-level semantic information (e.g., bedroom, kitchen) into the model. However, these approaches either oversimplify the rich spatial dynamics of human movement or discard fine-grained path information that is crucial for activity recognition.

For temporal information, many prior approaches rely on sliding window segmentation to capture sequential dependencies \cite{liciotti2020sequential, gochoo2018unobtrusive, mohmed2020employing, lesani2021smart, arrotta2022dexar}. Some methods, such as \cite{chen2024towards, ye2023graph, thukral2025layout}, additionally embed coarse-grained time features, such as weekday indicators or discretized hour bins, into the model. Although these features provide basic routine context, they do not fully capture the periodic and cyclical structure of human behavior. Furthermore, few studies explicitly combine multiple temporal components—such as hour, weekday, and minute—to enhance time-aware representations across different granularities. As a result, these models often struggle to disambiguate activities that share spatial similarities but differ temporally, such as distinguishing morning and evening medication routines within the same kitchen region.

Overall, these limitations motivate the need for frameworks that explicitly model both spatial trajectories and rich temporal patterns, providing stronger contextual grounding for real-time activity recognition in noisy and ambiguous sensor streams.
\section{Methodology}
In this section, we present a trajectory-based deep learning framework for human activities of daily living recognition based on ambient sensor data. Our approach is designed to capture both spatial and temporal patterns inherent in sensor activation sequences while maintaining flexibility and robustness in real-world smart home environments. 

We consider the task of human activity recognition in smart home environments using ambient sensor data. Given a time-ordered sequence of sensor events
\begin{equation}
E = \{e_1, e_2, \ldots, e_T\}, \quad \text{where} \quad e_t = [\text{sensor}_t, \text{status}_t, \text{timestamp}_t],
\label{eq:event_sequence}
\end{equation}
our goal is to predict the activity label associated with segments of the sequence. We segment the event stream into overlapping sliding windows
\begin{equation}
W_i = \{e_i, e_{i+1}, \ldots, e_{i+w-1}\}
\label{eq:window}
\end{equation}
of fixed size $w$ and stride $s$, producing a set of windows $\{W_1, W_2, \ldots, W_n\}$.
Each window $W_i$ is assigned a label $y_i \in \mathcal{Y}$ from a predefined activity set, where the label is determined by the activity tag of the last event within the window.

Formally, we aim to learn a mapping
\begin{equation}
f: W_i \mapsto y_i,
\label{eq:classifier}
\end{equation}
where $f$ is a classifier that predicts the activity label based on the observed sequence of sensor events.
\begin{figure}[h!]
  \centering
   \includegraphics[width=0.8\linewidth]{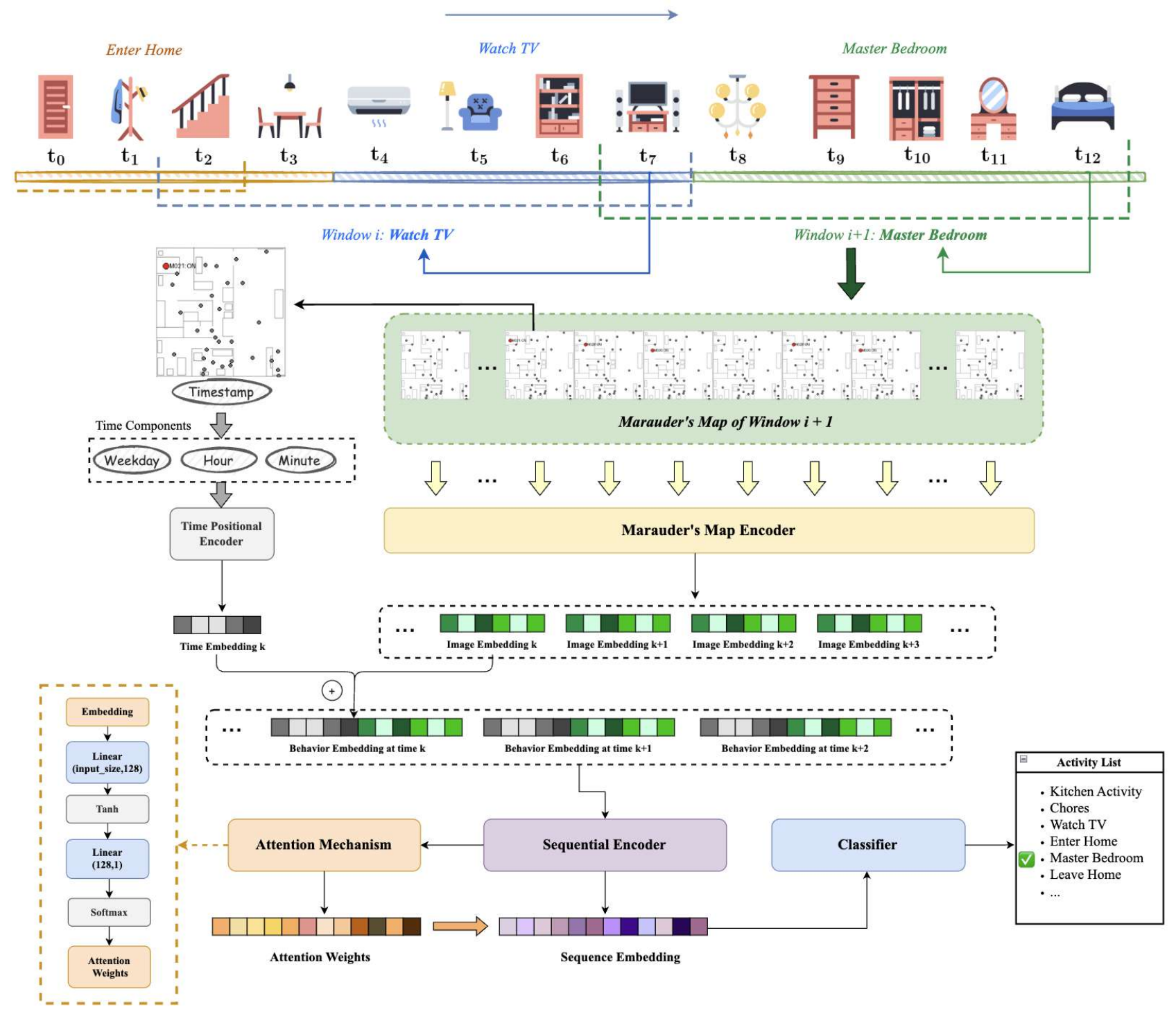}
   \caption{Overview of the proposed Marauder’s Map framework for cross-activity human activity recognition. Sensor activations are collected within temporal windows and transformed into spatially structured trajectory images. Simultaneously, timestamp information is encoded through a time positional encoder using weekday, hour, and minute components. Both image and time embeddings are fused into behavior embeddings at each timestep and then sequentially processed by an RNN encoder. An attention mechanism is applied to the RNN outputs, dynamically assigning weights to highlight salient temporal features across the sequence. The resulting context vector is passed to a classifier for final activity prediction. The architecture of the attention mechanism is shown on the left, consisting of a two-layer fully connected network with Tanh activation and softmax normalization to generate attention scores.}
   \label{fig:overarch}
\end{figure}

The proposed methodology consists of four main stages, shown in Fig.~\ref{fig:overarch}: (1) image sequence generation, where temporal windows of sensor activations are transformed into layout-based trajectory images; (2) temporal encoding, where timestamps associated with each window are embedded through a time positional encoder; (3) behavior sequence construction, where image features extracted via a convolutional neural network (CNN) are concatenated with time embeddings to form fused behavior embeddings; and (4) sequence modeling and classification, where an LSTM encoder captures temporal dependencies across the behavior sequence, an attention mechanism highlights informative timesteps, and a lightweight multilayer perceptron (MLP) predicts the final activity label. We first illustrate the image sequence generation process in Section~\ref{img_gene}, followed by the temporal embedding generation in Section~\ref{tim_emb}. Finally, in Section~\ref{pipline}, we introduce the behavior sequence construction and the overall sequence modeling and classification pipeline.

\subsection{Spatiotemporal Representation via MARAuder’s Map}
\subsubsection{Spatial Encoding via Trajectory Image Sequences}
\label{img_gene}
To incorporate spatial priors from the smart home environment into our model, we propose a method to convert sequences of raw sensor activations into structured image representations based on the actual floorplan layout of the smart home. This enables the model to learn spatial patterns of user trajectories over time, while preserving the physical relationships between sensor locations. Each sensor in the environment is assigned a unique coordinate $(x_s, y_s)$ on the original floorplan, typically defined in pixels. We scale these coordinates to fit a fixed-size image grid of resolution $R \times R$ according to the original width and height of the floorplan. Each sensor type is associated with a distinct RGB color $\boldsymbol{c}_t \in \mathbb{R}^3$, defined in a pre-assigned color palette.
\begin{figure}[h!]
  \centering
   \includegraphics[width=\linewidth]{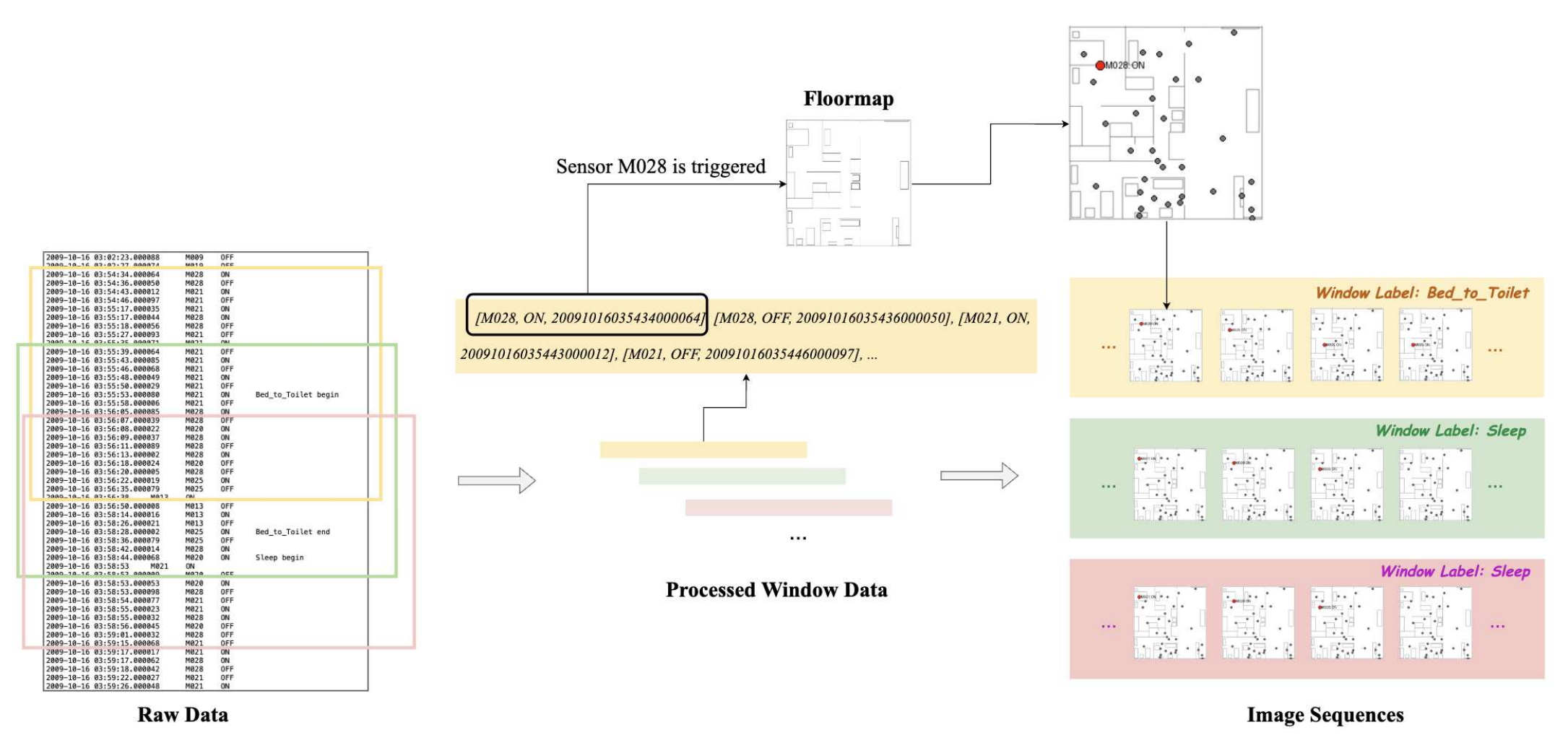}
   \caption{Pipeline for generating layout-based image sequences from smart home sensor logs. The process begins by slicing the raw sensor event logs into overlapping time windows. For each window, sensor triggers are identified, and their corresponding spatial locations are mapped onto a floorplan. Each triggered sensor is visualized as a plotted marker in the image, resulting in a sequence of floorplan-based frames that reflect spatial-temporal activity patterns. Each image sequence is labeled based on the activity occurring within the window, forming the input for activity recognition models.}
   \label{fig:img_seq}
\end{figure}

For each timestep $t$, the floorplan is loaded as background, and for every active sensor (\texttt{status = 1}) or temperature-type sensor, we render a small colored footprint centered at the scaled coordinates $(x_s', y_s')$.
This results in a 3D tensor representation of shape $(3, R, R)$ per timestep. Given a batch of sequences with shape $(B, T, 2)$---where each item contains a sensor ID and status---we obtain an output tensor of shape $(B, T, 3, R, R)$, which encodes the spatial footprint of each activation over time. The overview of image sequence generation is shown as Fig.\ref{fig:img_seq}. By plotting the active sensors directly on the floorplan-aligned canvas, this approach retains key spatial context and allows the model to infer movement trajectories and activity hotspots. These structured image sequences are particularly well-suited for spatial-temporal modeling using convolutional or attention-based architectures.

\subsubsection{Temporal Context Embedding with Structured Time Decomposition}
\label{tim_emb}
To capture periodic and contextual temporal patterns from sensor event sequences, we construct a continuous and interpretable representation of time based on decomposed timestamp components. Each sensor event is represented as:
\[
e_i = [s_i, v_i, t_i],
\]
where $s_i$ is the sensor ID, $v_i$ is the sensor status (e.g., ON/OFF), and $t_i$ is the timestamp in the format \texttt{YYYYMMDDHHMMSS}.

To model temporal dependencies, the raw timestamp $t_i$ is decomposed into discrete components:
\begin{itemize}
    \item \textbf{Month} ($M$): $M \in [1, 12]$
    \item \textbf{Day of Month} ($D$): $D \in [1, 31]$
    \item \textbf{Day of Week} ($W$): $W \in [0, 6]$ (Monday to Sunday)
    \item \textbf{Hour of Day} ($H$): $H \in [0, 23]$
    \item \textbf{Minute of Hour} ($m$): $m \in [0, 59]$
\end{itemize}

\textbf{Example of timestamp decomposition:}
\begin{itemize}
    \item $t_i = \texttt{20091202202730}$ $\rightarrow$ $M=12$, $D=2$, $W=2$ (Wednesday), $H=20$, $m=27$
    \item $t_i = \texttt{20100315110610}$ $\rightarrow$ $M=3$, $D=15$, $W=1$ (Monday), $H=11$, $m=6$
    \item $t_i = \texttt{20211009074245}$ $\rightarrow$ $M=10$, $D=9$, $W=5$ (Saturday), $H=7$, $m=42$
\end{itemize}

\textbf{Cyclic Encoding of Components:}
The set of selected components is denoted as $\mathcal{C}$. Each component $x \in \mathcal{C}$ is normalized by its predefined range $R_x$ and encoded into a 2D cyclic representation:
\begin{equation}
\theta_x = 2\pi \cdot \frac{x}{R_x},
\quad
\text{feat}_x = [\sin(\theta_x), \cos(\theta_x)].
\label{eq:cyclic}
\end{equation}

The 2D features are then projected into a higher-dimensional embedding space:
\begin{equation}
\text{emb}_x = \mathbf{W}_x \cdot \text{feat}_x + \mathbf{b}_x,
\label{eq:linear_projection}
\end{equation}
where $\mathbf{W}_x \in \mathbb{R}^{d_x \times 2}$ and $\mathbf{b}_x \in \mathbb{R}^{d_x}$ are learnable parameters, and $d_x$ is the embedding dimension allocated to component $x$.

Given a total time embedding size $d$, we evenly divide it across all components:
\begin{equation}
d_x = \left\lfloor \frac{d}{|\mathcal{C}|} \right\rfloor,
\end{equation}
assigning any remaining dimensions to the final component.
The final time embedding vector $\mathbf{e}_{\text{time}}$ is obtained by concatenating all projected embeddings:
\begin{equation}
\mathbf{e}_{\text{time}} = \left[ \text{emb}_M, \text{emb}_D, \text{emb}_W, \dots \right] \in \mathbb{R}^d.
\label{eq:final_time_emb}
\end{equation}

This continuous, cyclically encoded representation enables the model to learn both periodic patterns (e.g., daily routines) and contextual temporal cues from ambient sensor data.

\subsection{Layout and Time-Aware Activity Recognition Pipeline using Marauder’s Map}
\label{pipline}
As shown in Fig. \ref{fig:overarch}, after constructing the sequence of sensor activation images and extracting their associated timestamp information, the model proceeds to encode each input into a rich behavior representation. Each image in the sequence is first processed independently by a convolutional neural network (CNN) to extract spatial feature embeddings that capture the distribution and intensity of sensor activations within the home layout. In parallel, the corresponding timestamp of each image is encoded into a low-dimensional time embedding using a time positional encoder that encodes components such as weekday, hour, and minute through periodic functions. The CNN image features and the time embeddings are then concatenated at each timestep to form a sequence of fused behavior embeddings. This behavior sequence is subsequently fed into a sequence encoder, which can be flexibly configured as a GRU, LSTM, or BiLSTM depending on task requirements. In this work, we employ a simple LSTM network to model the temporal dependencies within the behavior sequence, illustrating the potential of our framework independent of architectural complexity. An attention mechanism is applied to the LSTM outputs to dynamically emphasize the most informative parts of the sequence. The resulting context vector is finally classified by a lightweight multilayer perceptron (MLP) to predict the target activity. This modular encoding pipeline enables the model to capture both fine-grained spatial-temporal patterns and long-term dependencies in smart home sensor data.

\section{Experiments}
We structure our experimental evaluation around the following four core research questions:

\begin{itemize} 
\item \textbf{Q1:} Does the proposed method outperform existing approaches? 
\item \textbf{Q2:} How does the method perform under real-time conditions, particularly with cross-activity windows? 
\item \textbf{Q3:} What is the impact of key architectural components in the model? 
\item \textbf{Q4:} How do the window size and overlapping portion affect performance? 
\end{itemize}

To address Q1, we compare our method against state-of-the-art baselines across multiple smart home datasets. Results show that our approach consistently achieves higher classification accuracy, demonstrating superior activity recognition in sequential sliding windows.

For Q2, we simulate real-time conditions where input windows may span multiple activities. These experiments evaluate the model’s robustness to ambiguous transitions without precise temporal segmentation. Our method exhibits greater resilience and outperforms baseline models under these challenging streaming scenarios.

To investigate Q3, we conduct an ablation study analyzing the contributions of major architectural components. We assess the effects of different image emulator designs based on floorplans and varying combinations of temporal components in the time embedding module. Results confirm that both spatial and temporal encoding strategies are crucial for achieving strong performance.

Finally, for Q4, we vary the window size and overlapping portion to study their influence on model effectiveness. Findings reveal that performance is sensitive to these hyperparameters, and optimal settings depend on the complexity of activity patterns and the sensor density of the environment.

\subsection{Experimental Setup}
\subsubsection{Dataset}
To evaluate the effectiveness of our proposed mechanism, we conduct experiments on three datasets from CASAS dataset\cite{cook2012casas}, which are Milan, Kyoto7, and Aruba. To provide context and highlight the distinct characteristics of each dataset, we briefly describe Milan, Kyoto7, and Aruba individually below.

\begin{figure}[h!]
    \centering
    \begin{tabular}{ccc}
        \begin{subfigure}[t]{0.24\linewidth}
            \centering
            \includegraphics[width=\linewidth]{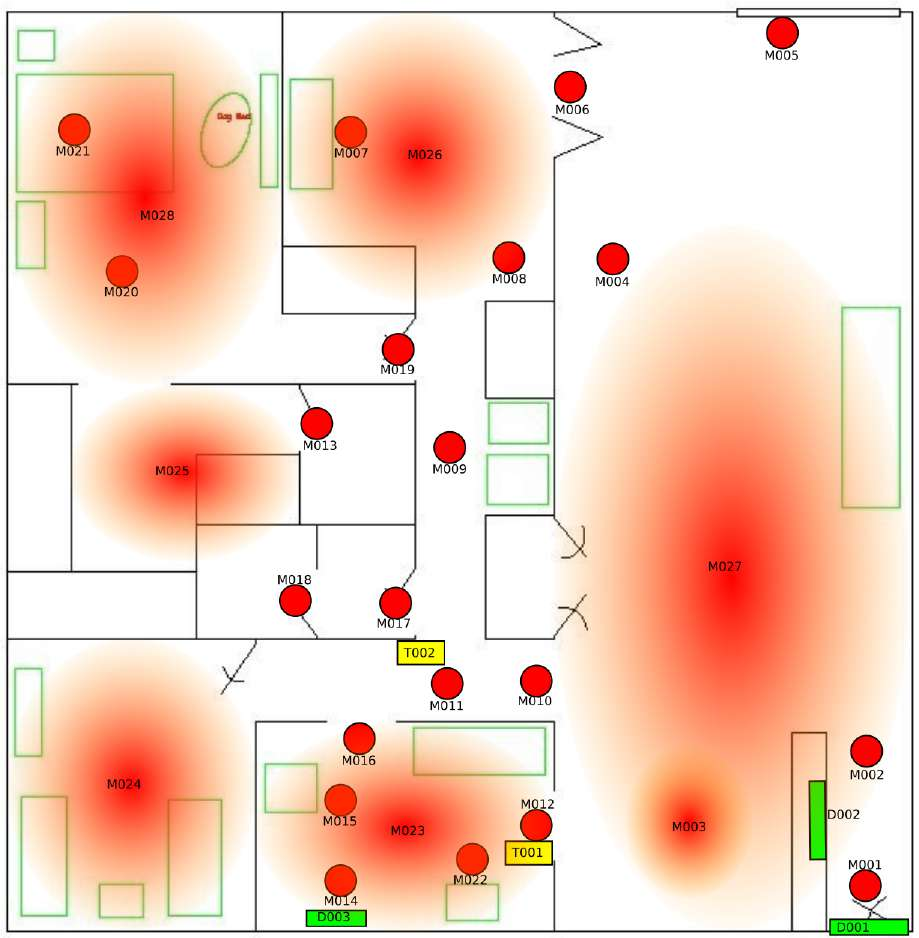}
            \caption{Milan}
            \label{fig:laymilan}
        \end{subfigure}
        \hfill
        \begin{subfigure}[t]{0.34\linewidth}
            \centering
            \includegraphics[width=\linewidth]{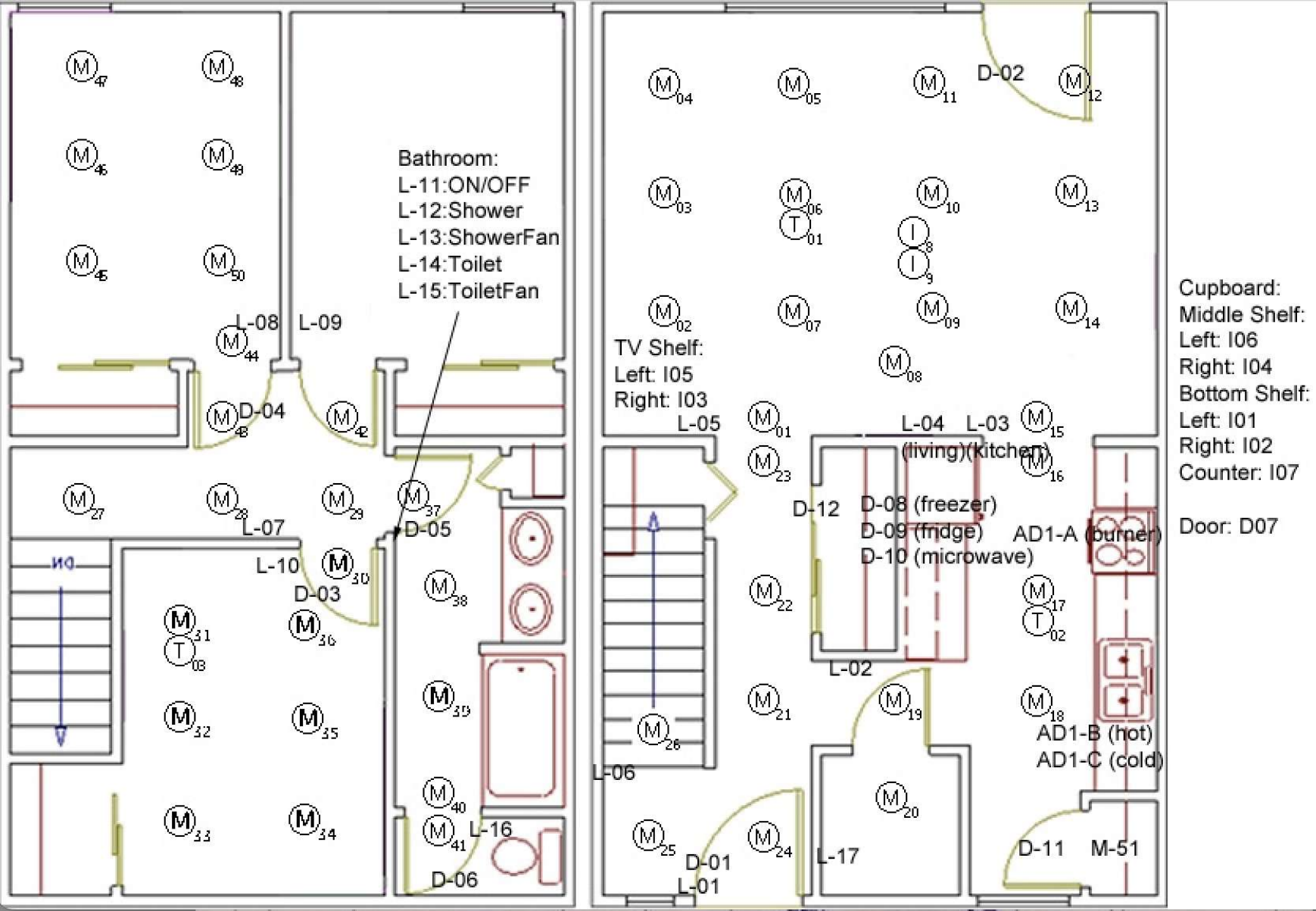}
            \caption{Kyoto7}
            \label{fig:laykyoto}
        \end{subfigure} 
        \hfill
        \begin{subfigure}[t]{0.28\linewidth}
            \centering
            \includegraphics[width=\linewidth]{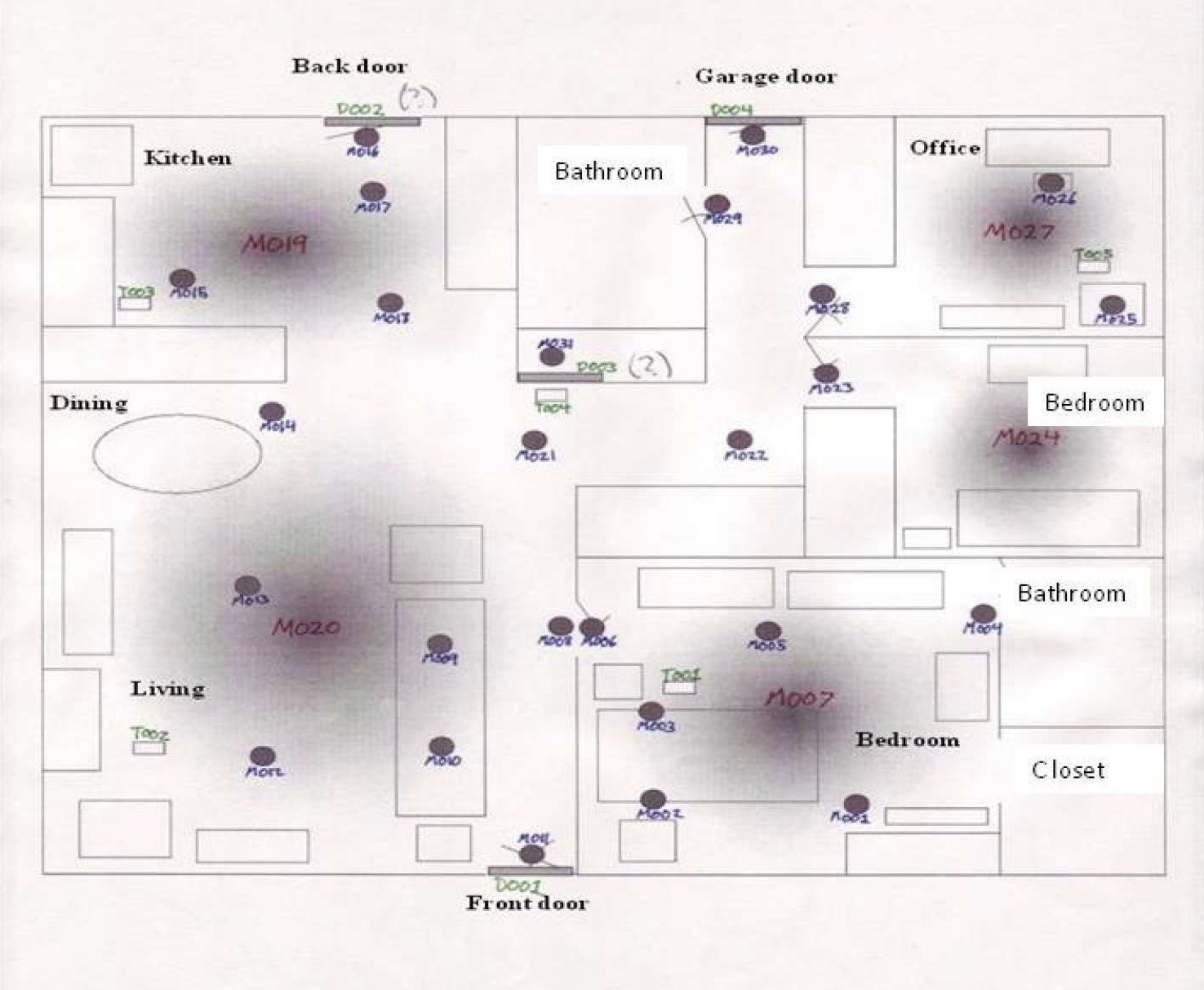}
            \caption{Aruba}
            \label{fig:layaruba}
        \end{subfigure} 
    \end{tabular}
    \caption{Floorplan layouts and sensor distributions for three smart homes: Milan, Kyoto7, and Aruba.}
    \label{fig:floormap}
\end{figure}

The Milan dataset records single-resident activity in a smart home equipped with motion ("M"), door ("D"), and temperature ("T") sensors, as illustrated in Fig.~\ref{fig:laymilan}. Fifteen distinct activities are annotated, ranging from frequent activities such as Kitchen\_Activity (554 instances) and Guest\_Bathroom (330 instances) to rare events like Meditate (17 instances). Milan provides a rich temporal distribution for studying personalized activity recognition.

The Kyoto7 dataset captures multi-resident activity from an apartment shared by two individuals (R1 and R2), shown in Fig.~\ref{fig:laykyoto}. In addition to motion, door, and temperature sensors, it includes appliance sensors (e.g., burner, water flow) and item sensors ("I"). Activities are annotated at the resident level (e.g., R1\_Work, R2\_Sleep) and shared level (e.g., Clean, Study), with Meal\_Preparation being the most frequent. Kyoto7 enables investigation of complex, interleaved, and resident-specific activity patterns.

The Aruba dataset records single-resident behavior in a home visited regularly by family members, with the sensor layout shown in Fig.~\ref{fig:layaruba}. It includes 11 annotated activities, dominated by Relax (2910 instances) and Meal\_Preparation (1606 instances), while rare activities such as Resperate (6 instances) pose class imbalance challenges.

\begin{figure}[h!]
    \centering
    \begin{tabular}{ccc}
        \begin{subfigure}[b]{0.33\textwidth}
        \centering
        \includegraphics[width=\textwidth]{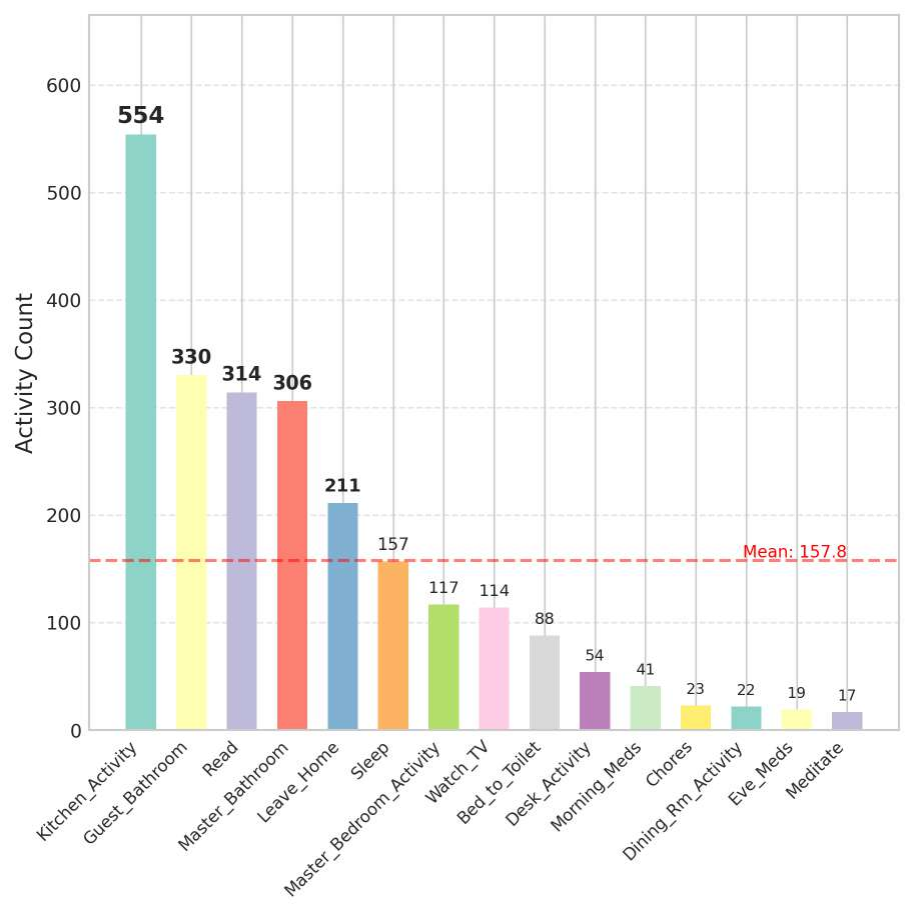}
        \caption{Milan}
        \label{fig:labeldistmilan}
    \end{subfigure}
    \hfill
    \begin{subfigure}[b]{0.33\textwidth}
        \centering
        \includegraphics[width=\textwidth]{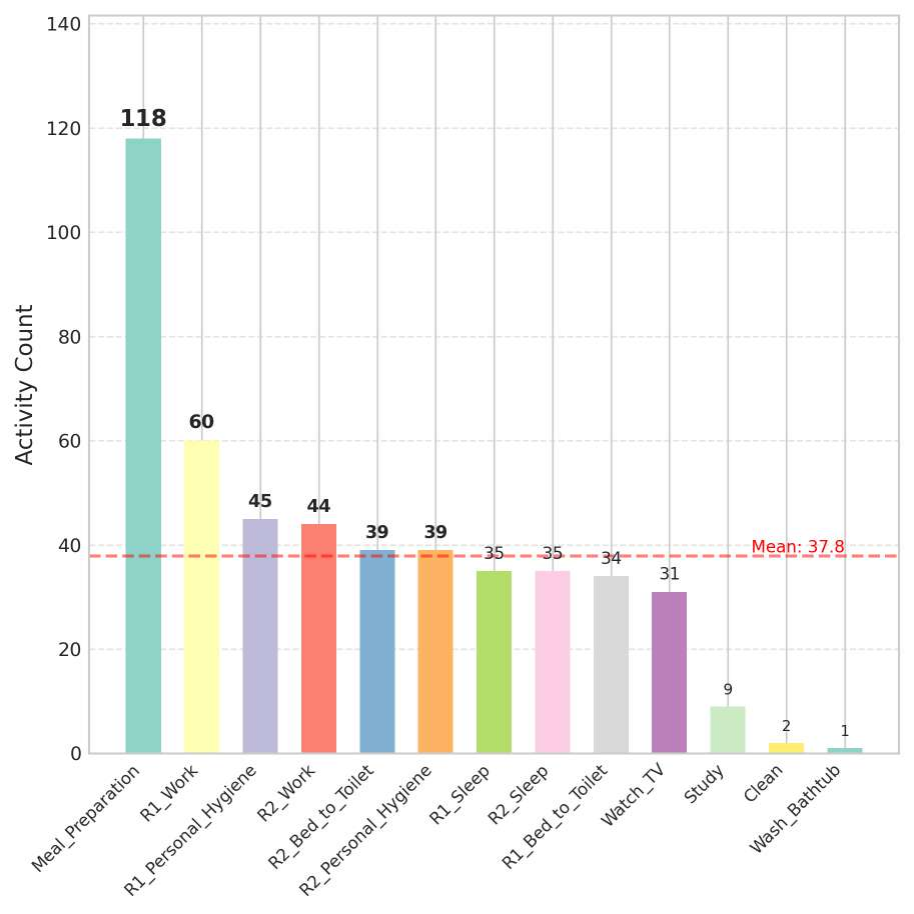}
        \caption{Kyoto7}
        \label{fig:labeldistkyoto}
    \end{subfigure}
    \hfill
    \begin{subfigure}[b]{0.33\textwidth}
        \centering
        \includegraphics[width=\textwidth]{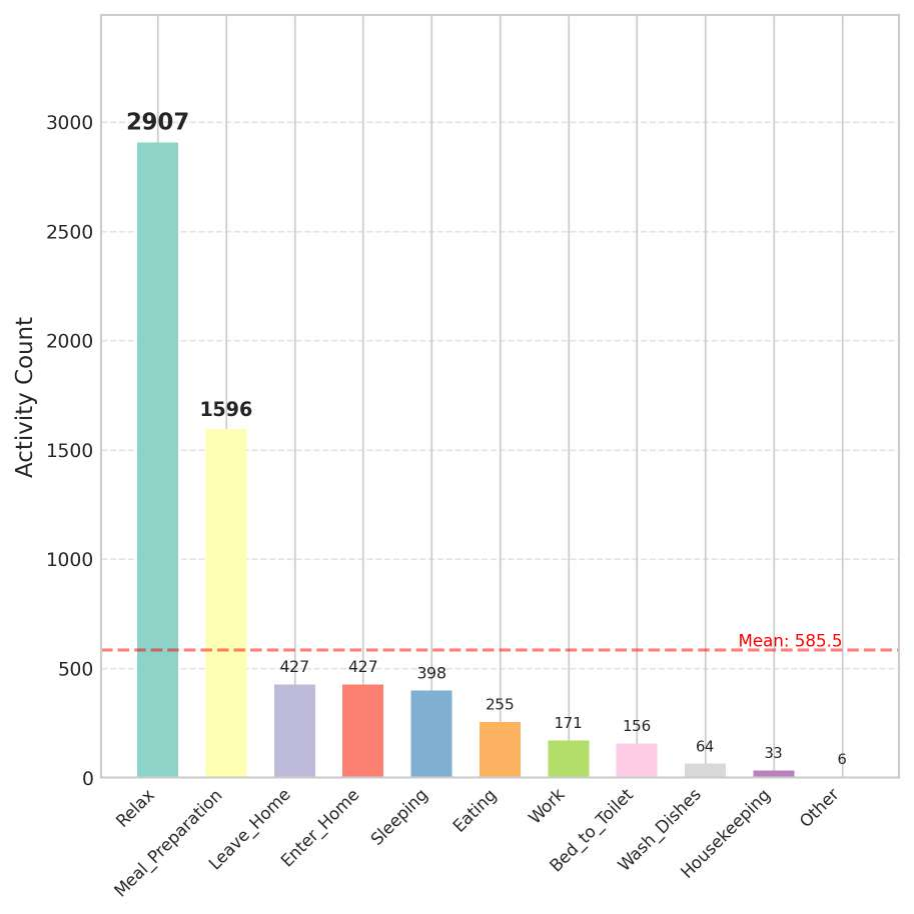}
        \caption{Aruba}
        \label{fig:labeldistaruba}
    \end{subfigure}
    \end{tabular}
    \caption{Histogram of activity distributions across three smart home datasets: (a) Milan, (b) Kyoto7, and (c) Aruba.}
    \label{fig:activity_histograms}
\end{figure}

These datasets cover a range of smart home scenarios—from single-resident (Milan, Aruba) to multi-resident (Kyoto7) settings—providing diverse temporal patterns, class imbalance, and sensor modalities. By analyzing the raw dataset, the resulting activity distribution is shown in Fig.~\ref{fig:activity_histograms}, which exhibits a mix of frequent and rare activities—critical for evaluating model robustness under class imbalance. Furthermore, the inclusion of timestamped annotations and fine-grained sensor events enables realistic sliding window-based evaluation, while the physical and sensor diversity supports testing of spatiotemporal sequence learning models.
\subsubsection{Baselines}
To evaluate the overall effectiveness of the proposed method, we begin by comparing Marauder's Map method against several state-of-the-art ADL modeling algorithms.
\begin{itemize}[leftmargin=*]
\item[$\bullet$] \textbf{DeepCasas} \cite{liciotti2020sequential}:
DeepCasas converts each event in the sensor log into an integer, which is subsequently embedded into a continuous vector space. The embedded sequences are fed into an LSTM-based architecture, followed by a fully connected block, to perform activity classification. This model captures temporal dependencies in the sensor sequences through recurrent processing.
\item[$\bullet$] \textbf{1DCNN} \cite{kiranyaz20211d}:
The 1D CNN model is designed for processing sequential data such as time-series sensor streams. It applies one-dimensional convolutional filters across time to extract localized temporal patterns. This approach emphasizes efficient learning of short-term dependencies in the data without relying on recurrent connections.
\item[$\bullet$] \textbf{DCNN} \cite{gochoo2018unobtrusive}:
The DCNN model transforms sequences of sensor events into binary images, where the x-axis corresponds to the chronological order of events and the y-axis represents sensor IDs. The resulting images are processed using a deep convolutional network consisting of three convolutional layers followed by three fully connected layers. This method leverages spatial features in the image representation to model sensor activity patterns.
\item[$\bullet$] \textbf{greyDCNN} \cite{mohmed2020employing}:
GreyDCNN is an extension of DCNN that incorporates the temporal duration of activities into the image representation. While the layout of the binary image remains unchanged, the color intensity of each dot reflects the duration of the corresponding event—brighter dots indicate longer durations. This additional temporal cue provides the model with more contextual information for improved classification.
\item[$\bullet$] \textbf{LSTM-CNN} \cite{xia2020lstm}:
The LSTM-CNN hybrid model combines the strengths of both recurrent and convolutional architectures. Initially, LSTM layers are used to embed the sequential sensor data, capturing long-range dependencies. Subsequently, one-dimensional convolutional layers are applied to extract local temporal features from the embedded sequence. This combination enables robust modeling of complex activity patterns.
\item[$\bullet$] \textbf{Transformer} \cite{vaswani2017attention}:
A Transformer-based architecture is employed to capture temporal dependencies and complex patterns in sensor event sequences. Raw sensor data are first converted into image representations using a CNN-based encoder, followed by positional encoding and Transformer encoders to model the sequential nature of activities. This model is an adaptation of the Transformer framework tailored for the sensor-based ADL classification task.
\item[$\bullet$] \textbf{TCN} \cite{bai2018empirical}:
Temporal Convolutional Networks (TCNs) use dilated 1D convolutions with residual connections to model sequential data efficiently. Sensor event indices are first embedded into dense vectors, which are processed through stacked temporal blocks composed of dilated convolutions, Chomp1d layers to maintain sequence length, ReLU activations, and dropout. The exponentially increasing dilation enables capturing both short- and long-term dependencies. Finally, the last timestep’s output is passed through a fully connected layer for activity classification.
\end{itemize}
The examined baselines reveal diverse modeling strategies for activity recognition in smart home environments, with varying emphases on temporal and spatial features. Models such as DeepCasas, 1DCNN, LSTM-CNN, and TCN leverage sequential information from sensor event logs without incorporating explicit time embeddings (e.g., timestamps or durations as embeddings). Among these, DeepCasas and LSTM-CNN utilize recurrent layers to capture long-range dependencies, whereas 1DCNN and TCN rely on convolutional mechanisms—TCN in particular employs dilated convolutions to model longer temporal dependencies while enabling parallel computation. In contrast, models like Transformer, DCNN, and greyDCNN aim to exploit spatial characteristics by transforming event sequences into image-like formats. However, unlike methods that utilize actual physical layout or positional metadata, DCNN and greyDCNN construct image representations where the x-axis reflects event order and the y-axis maps sensor IDs, thus relying on artificial spatial structures rather than real-world sensor arrangements. greyDCNN adds a temporal nuance by varying the pixel intensity based on event duration. Notably, none of these baselines integrate explicit timestamp embeddings, and thus may struggle with subtle temporal misalignments or irregular sampling.

\subsection{Overall performance}
In this experiment, we evaluate the effectiveness of Marauder’s Map across three smart home datasets: Milan, Kyoto7, and Aruba. To evaluate the performance of our proposed method and baselines, we employ four standard metrics: accuracy, precision, recall, and F1 score. We denote the full version of our proposed framework as Marauder-Time+Attn, which incorporates both temporal embeddings and an attention mechanism. In contrast, Marauder-Basic represents the ablated variant that utilizes image sequences to capture human activity trajectories but excludes the temporal and attention components. The results presented in Table \ref{tab:overallperf} demonstrate the superiority of Marauder-Time+Attn over a comprehensive set of baseline models across all datasets and evaluation metrics. Compared to traditional CNN-based models (e.g., DCNN, 1DCNN, greyDCNN) and sequential deep learning models (e.g., LSTM-CNN, Transformer, TCN), our proposed method consistently achieves the highest accuracy, F1-score, precision, and recall.  In particular, the performance improvement over Marauder-Basic highlights the critical contribution of modeling temporal dependencies and emphasizing salient contextual features through attention. For example, on the Milan dataset, Marauder-Time+Attn improves the F1-score from 81.75\% to 86.27\% and accuracy from 88.00\% to 92.74\%, confirming the effectiveness of the temporal-aware and attention-enhanced design. Moreover, Marauder-Time+Attn exhibits greater robustness and generalization, as reflected in its lower standard deviations across most metrics, indicating stable predictions under varying user behavior patterns. Among all baselines, TCN achieves the strongest performance across all three datasets, followed by DeepCasas, demonstrating the effectiveness of sequential modeling under consecutive window settings. For CNN-based approaches, DCNN performs best, suggesting that encoding sensor sequences into spatial representations better aligns with CNN architectures. These observations further validate the efficacy of our framework, which leverages both spatial encoding and temporal modeling to advance sensor-based activity recognition.

\begin{table}[h!]
  \caption{Performance Comparison of Different Methods on CASAS Dataset}
  \label{tab:overallperf}
  \centering
  \resizebox{1\columnwidth}{!}{
    \setlength{\tabcolsep}{1mm}{
  \begin{tabular}{ccccccccccccc}
    \toprule
    \multirow{2}{*}{Model} & \multicolumn{4}{c}{Milan} & \multicolumn{4}{c}{Kyoto7} & \multicolumn{4}{c}{Aruba} \\
    \cmidrule(lr){2-5} \cmidrule(lr){6-9} \cmidrule(lr){10-13}
    & Acc & F1 & Prec & Rec & Acc & F1 & Prec & Rec & Acc & F1 & Prec & Rec \\
    \midrule
    DeepCasas\cite{liciotti2020sequential} 
      & 75.93 $\pm$ 0.57 & 63.81 $\pm$ 1.81 & 68.46 $\pm$ 2.70 & 62.48 $\pm$ 1.72 
      & 88.61 $\pm$ 2.61 & 83.93 $\pm$ 3.17 & 86.39 $\pm$ 2.19 & 82.20 $\pm$ 3.93 
      & 88.22 $\pm$ 0.34 & 69.66 $\pm$ 1.15 & 75.71 $\pm$ 1.05 & 66.44 $\pm$ 1.34 \\
    1DCNN\cite{kiranyaz20211d}
      & 42.07 $\pm$ 1.23 & 10.03 $\pm$ 1.11 & 12.67 $\pm$ 2.34 & 11.16 $\pm$ 0.97
      & 50.50 $\pm$ 1.07 & 17.53 $\pm$ 0.66 & 18.97 $\pm$ 1.68 & 20.07 $\pm$ 0.63
      & 54.14 $\pm$ 1.00 & 12.83 $\pm$ 0.87 & 16.94 $\pm$ 2.25 & 12.94 $\pm$ 0.69 \\
    DCNN\cite{gochoo2018unobtrusive} 
      & 67.21 $\pm$ 0.69 & 51.30 $\pm$ 1.10 & 53.52 $\pm$ 1.59 & 50.72 $\pm$ 1.21 
      & 87.57 $\pm$ 0.80 & 78.90 $\pm$ 1.04 & 83.13 $\pm$ 1.57 & 76.22 $\pm$ 0.77 
      & 82.92 $\pm$ 0.16 & 54.44 $\pm$ 1.38 & 58.14 $\pm$ 2.72 & 53.03 $\pm$ 1.84 \\
    greyDCNN\cite{mohmed2020employing} 
      & 64.41 $\pm$ 0.21 & 46.76 $\pm$ 0.59 & 50.98 $\pm$ 1.90 & 47.27 $\pm$ 1.07 
      & 82.31 $\pm$ 0.60 & 71.46 $\pm$ 1.96 & 76.64 $\pm$ 2.99 & 69.23 $\pm$ 1.82 
      & 82.02 $\pm$ 0.13 & 52.81 $\pm$ 0.85 & 57.11 $\pm$ 1.86 & 50.77 $\pm$ 1.08 \\
    LSTM-CNN\cite{xia2020lstm}
      & 62.36 $\pm$ 0.39 & 41.05 $\pm$ 1.58 & 47.30 $\pm$ 2.41 & 39.55 $\pm$ 1.87
      & 72.97 $\pm$ 0.56 & 52.16 $\pm$ 1.95 & 56.54 $\pm$ 3.90 & 51.29 $\pm$ 1.80
      & 81.73 $\pm$ 0.30 & 53.92 $\pm$ 1.10 & 56.04 $\pm$ 1.14 & 52.97 $\pm$ 1.11 \\
    Transformer\cite{vaswani2017attention} 
      & 64.82 $\pm$ 0.41 & 45.84 $\pm$ 1.49 & 47.51 $\pm$ 1.45 & 46.86 $\pm$ 1.95
      & 83.94 $\pm$ 0.53 & 73.11 $\pm$ 1.15 & 81.96 $\pm$ 0.99 & 71.76 $\pm$ 1.40 
      & 81.31 $\pm$ 0.22 & 51.56 $\pm$ 1.28 & 52.59 $\pm$ 1.72 & 51.83 $\pm$ 3.43 \\
    TCN\cite{bai2018empirical} 
      & \underline{84.73 $\pm$ 2.30} & \underline{68.30 $\pm$ 3.70} & \underline{73.48 $\pm$ 7.04} & \underline{67.29 $\pm$ 2.78} 
      & \underline{92.22 $\pm$ 2.61} & \underline{87.19 $\pm$ 3.00} & \underline{89.01 $\pm$ 2.61} & \underline{85.98 $\pm$ 2.92} 
      & \underline{93.23 $\pm$ 0.61} & \underline{78.98 $\pm$ 1.31} & \textbf{83.88 $\pm$ 0.84} & \underline{75.67 $\pm$ 1.99} \\
    \midrule
    Marauder-Basic 
      & \textbf{88.00 $\pm$ 1.12} & \textbf{81.75 $\pm$ 1.08} & \textbf{83.99 $\pm$ 1.88} & \textbf{80.58 $\pm$ 1.05} 
      & \textbf{92.78 $\pm$ 1.25} & \textbf{89.86 $\pm$ 0.97} & \textbf{91.41 $\pm$ 0.31} & \textbf{88.72 $\pm$ 1.45} 
      & \textbf{93.51 $\pm$ 0.10} & \textbf{80.51 $\pm$ 0.11} & \underline{82.00 $\pm$ 0.29} & \textbf{80.16 $\pm$ 0.26} \\
    Marauder-Time+Attn
      & \cellcolor{green!20}\textbf{92.74 $\pm$ 0.90} & \cellcolor{green!20}\textbf{86.27 $\pm$ 2.51} & \cellcolor{green!20}\textbf{87.04 $\pm$ 1.34} & \cellcolor{green!20}\textbf{86.08 $\pm$ 3.36} 
      & \cellcolor{green!20}\textbf{96.30 $\pm$ 0.87} & \cellcolor{green!20}\textbf{92.08 $\pm$ 1.13} & \cellcolor{green!20}\textbf{92.35 $\pm$ 1.51} & \cellcolor{green!20}\textbf{91.91 $\pm$ 0.80} 
      & \cellcolor{green!20}\textbf{95.80 $\pm$ 0.10} & \cellcolor{green!20}\textbf{85.12 $\pm$ 0.37} & \cellcolor{green!20}\textbf{85.68 $\pm$ 0.47} & \cellcolor{green!20}\textbf{84.94 $\pm$ 0.26} \\
    \bottomrule
  \end{tabular}}}
\end{table}
\begin{figure*}[h!]
    \centering
    \begin{tabular}{ccccc}
        \begin{subfigure}[t]{0.22\linewidth}
            \centering
            \includegraphics[width=\linewidth]{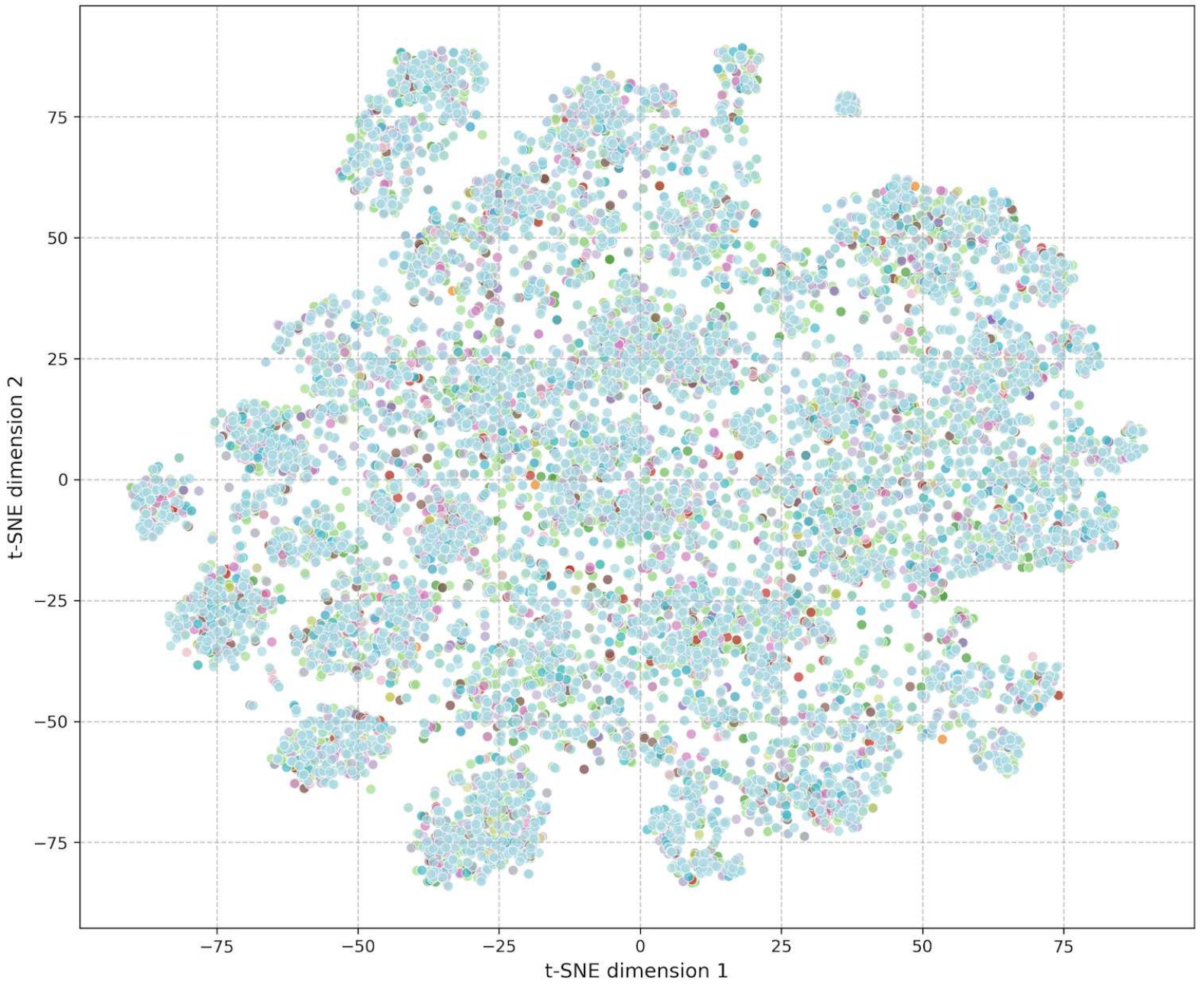}
            \caption{DeepCASAS}
            \label{fig:lstm_tsne}
        \end{subfigure}
        \hfill
        \begin{subfigure}[t]{0.22\linewidth}
            \centering
            \includegraphics[width=\linewidth]{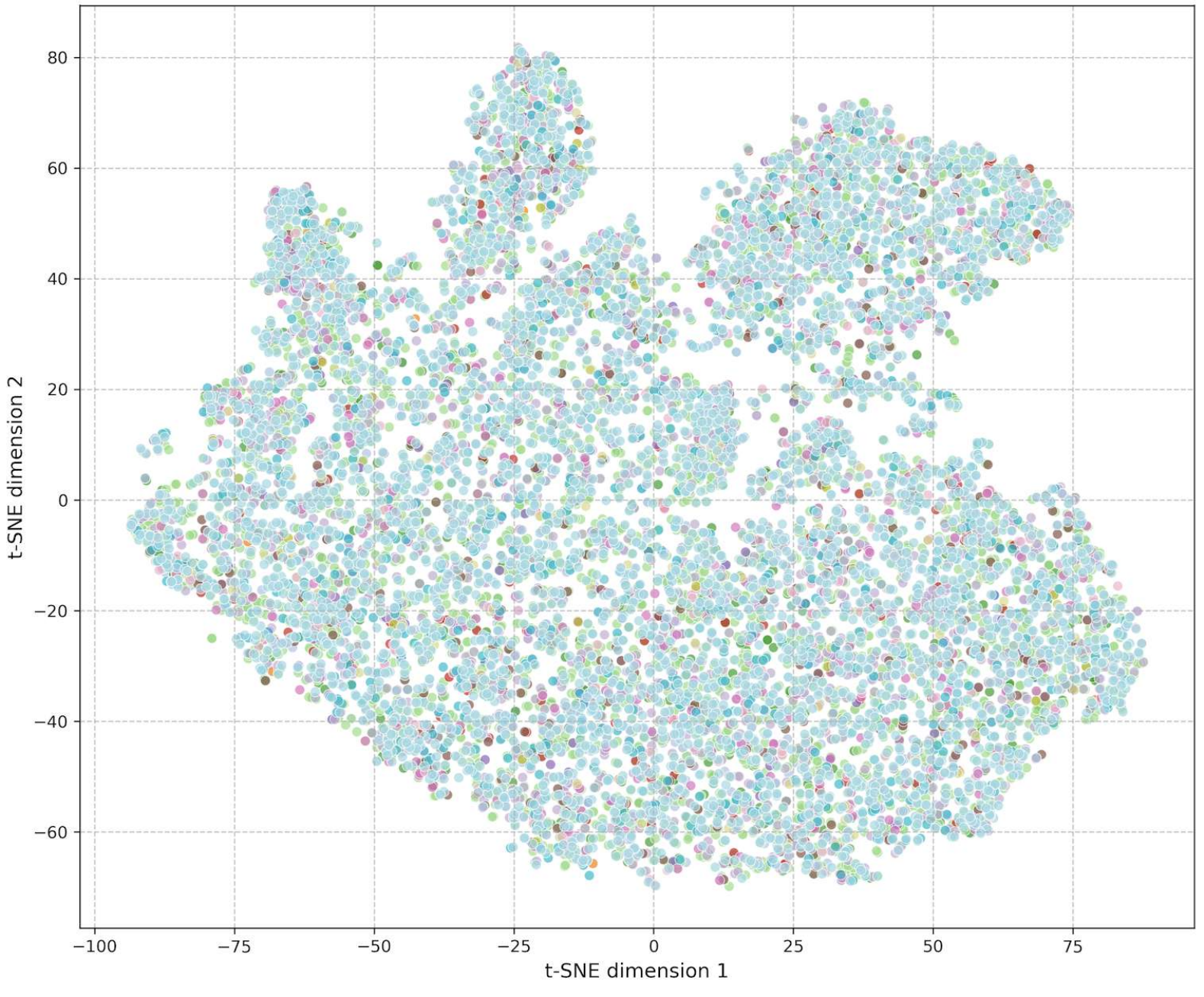}
            \caption{DCNN}
            \label{fig:dcnn_tsne}
        \end{subfigure} 
        \hfill
        \begin{subfigure}[t]{0.22\linewidth}
            \centering
            \includegraphics[width=\linewidth]{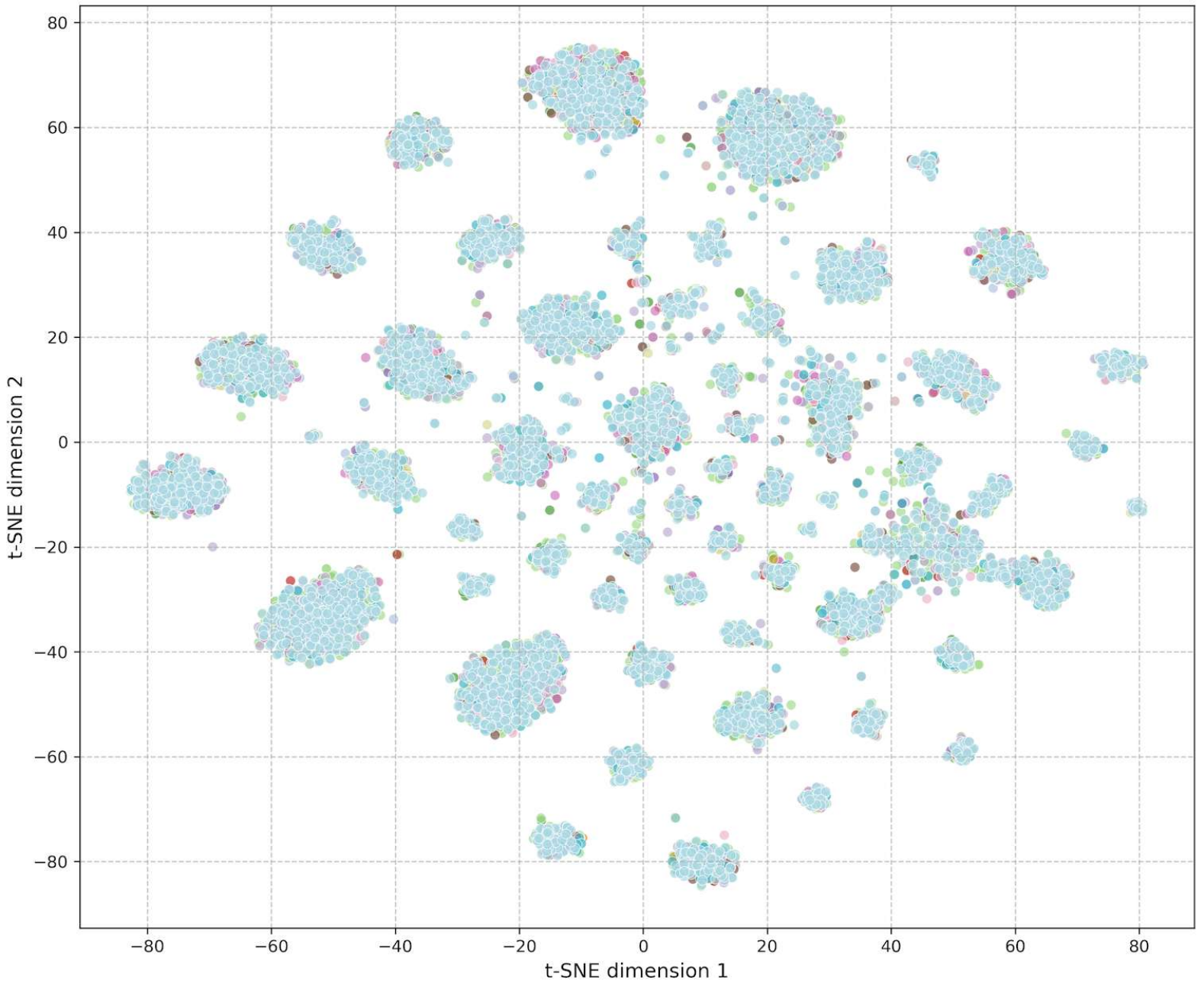}
            \caption{TCN}
            \label{fig:tcn_tsne}
        \end{subfigure} 
        \hfill
        \begin{subfigure}[t]{0.22\linewidth}
            \centering
            \includegraphics[width=\linewidth]{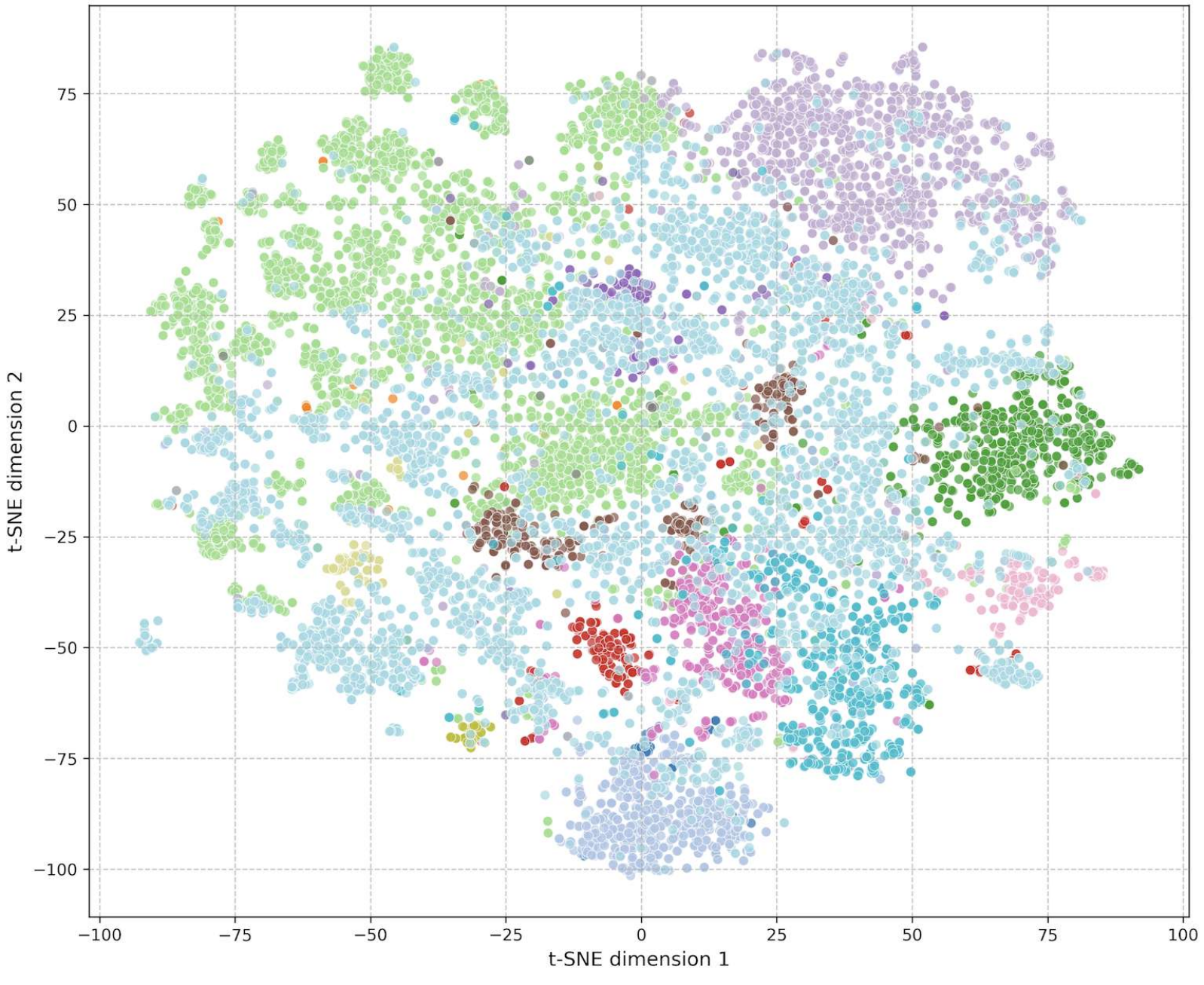}
            \caption{Marauder's Map}
            \label{fig:ours_tsne}
        \end{subfigure} 
        \hfill
        \begin{subfigure}[t]{0.12\linewidth}
            \centering
            \includegraphics[width=\linewidth]{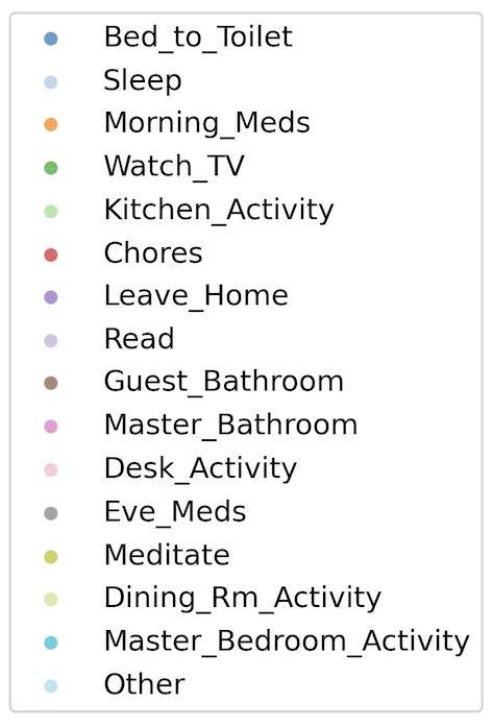}
            \label{fig:legend}
        \end{subfigure}\\
        \begin{subfigure}[t]{0.22\linewidth}
            \centering
            \includegraphics[width=\linewidth]{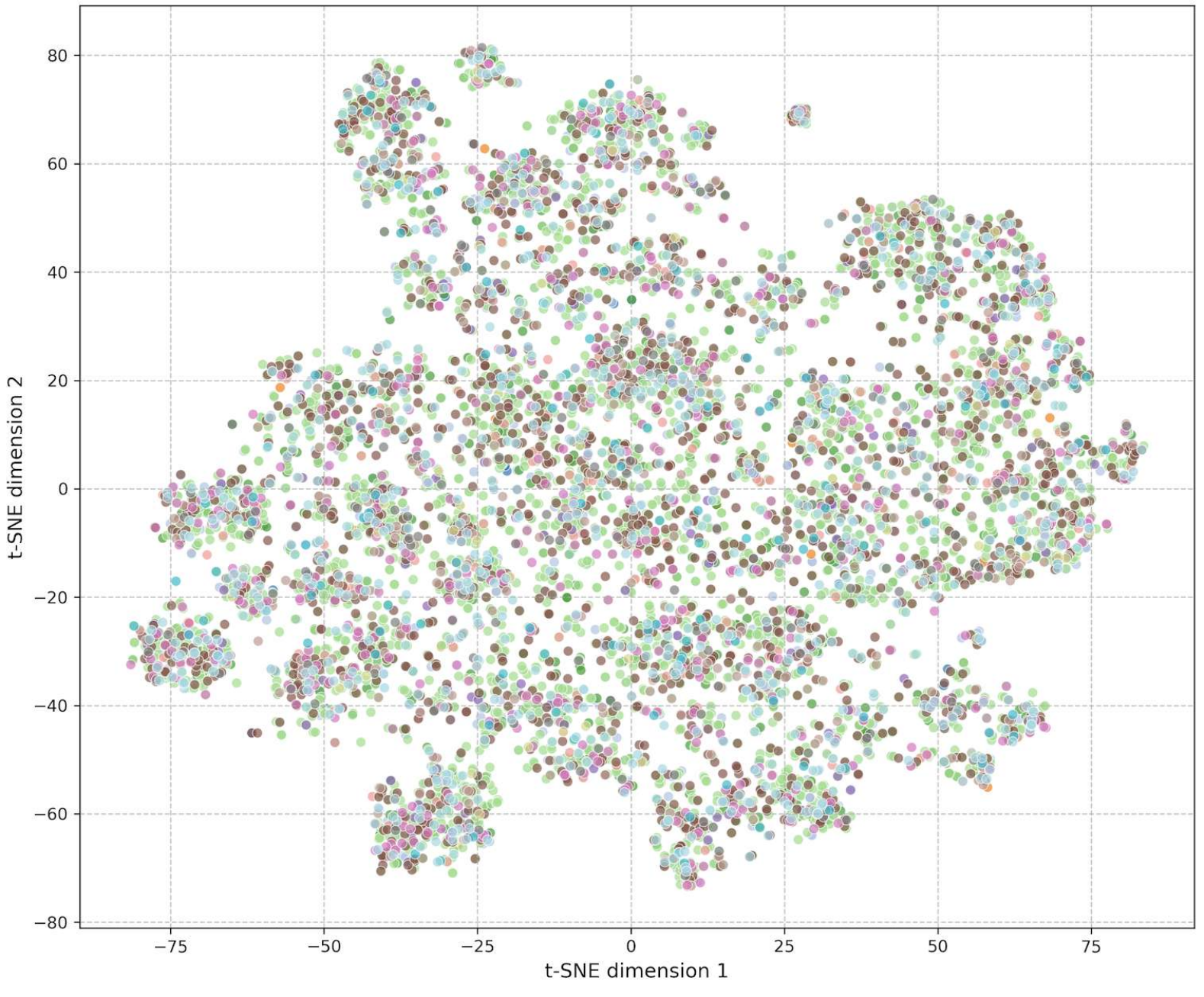}
            \caption{DeepCASAS}
            \label{fig:lstm_wo15}
        \end{subfigure}
        \hfill
        \begin{subfigure}[t]{0.22\linewidth}
            \centering
            \includegraphics[width=\linewidth]{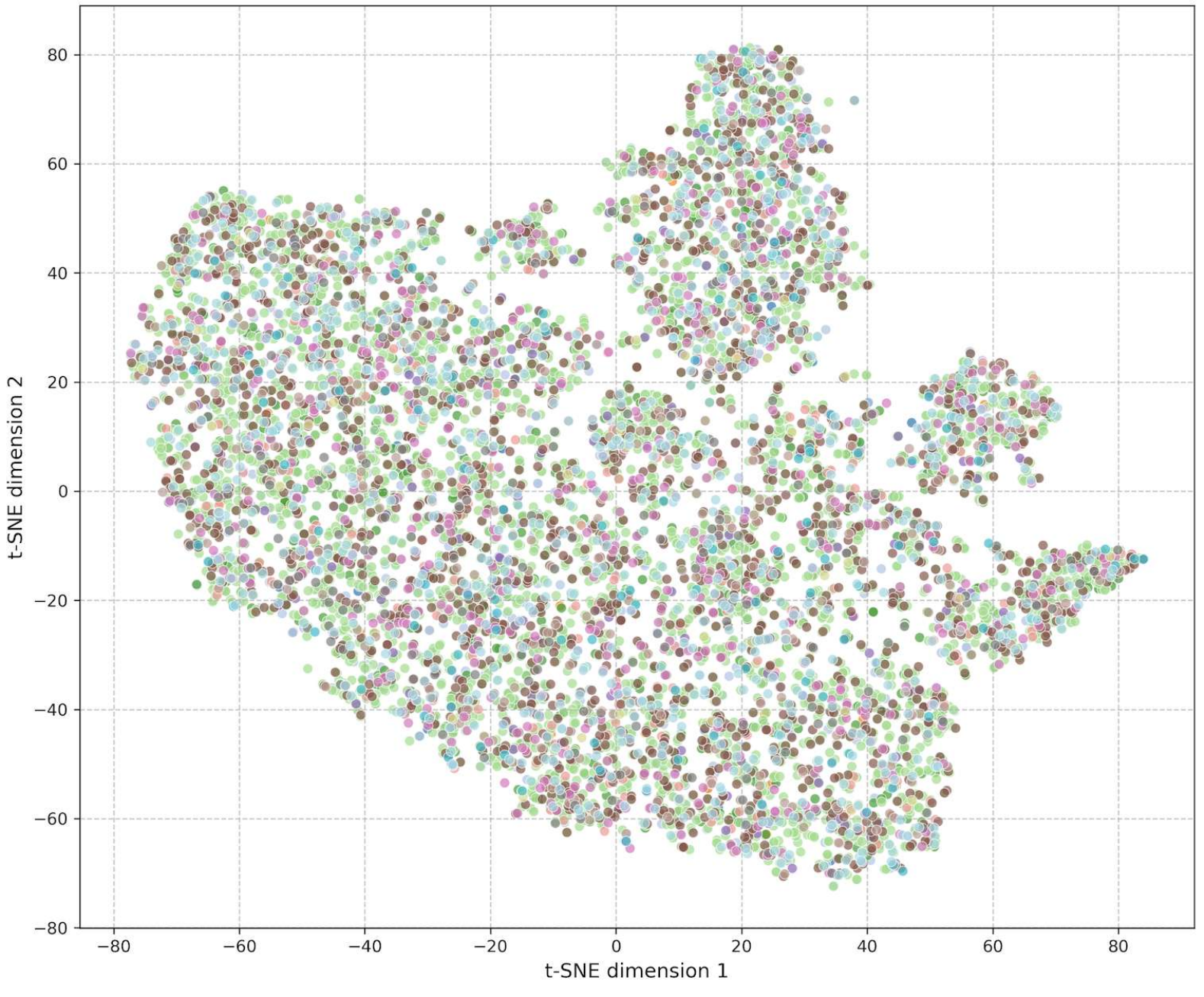}
            \caption{DCNN}
            \label{fig:dcnn_wo15}
        \end{subfigure} 
        \hfill
        \begin{subfigure}[t]{0.22\linewidth}
            \centering
            \includegraphics[width=\linewidth]{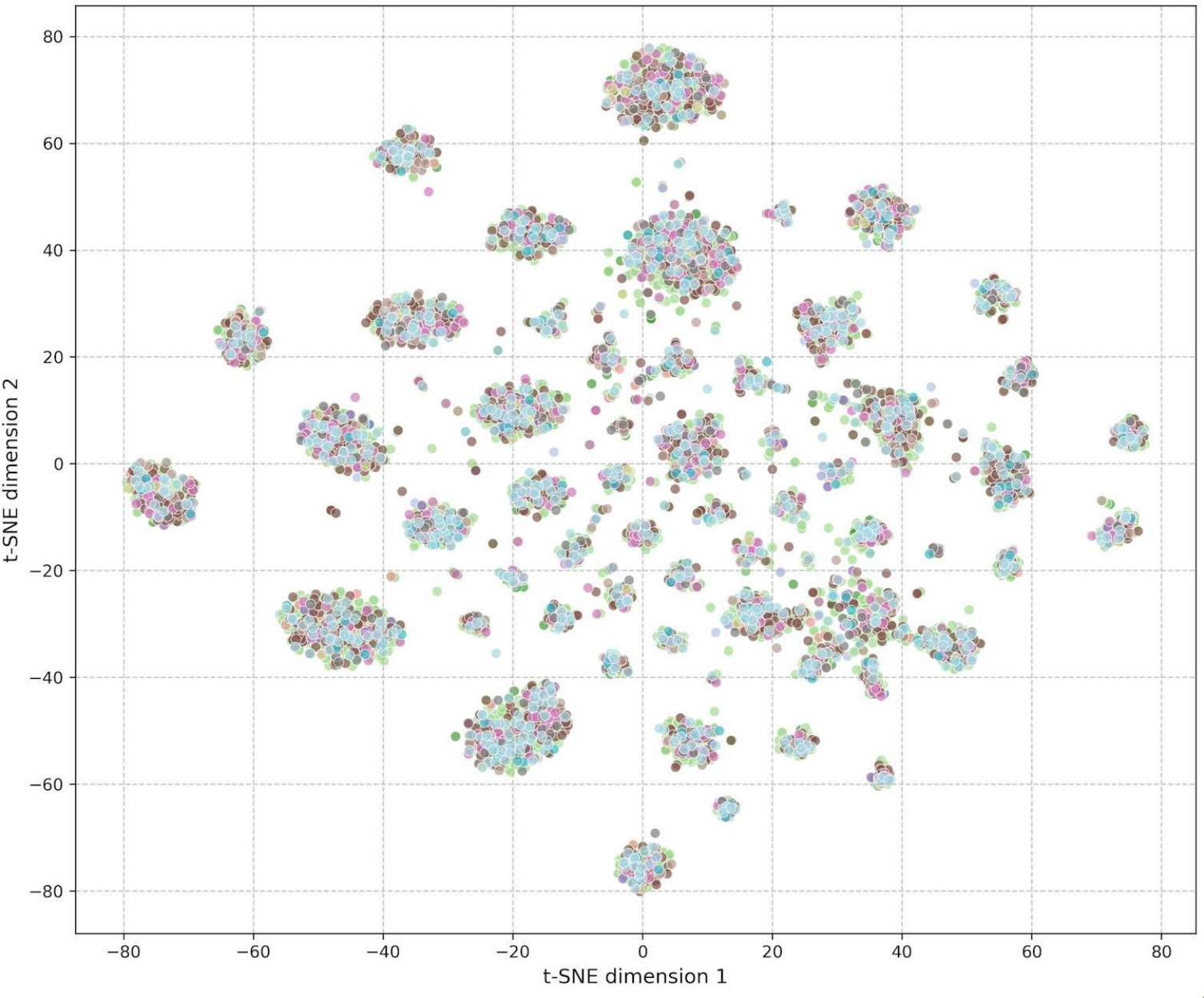}
            \caption{TCN}
            \label{fig:tcn_wo15}
        \end{subfigure} 
        \hfill
        \begin{subfigure}[t]{0.22\linewidth}
            \centering
            \includegraphics[width=\linewidth]{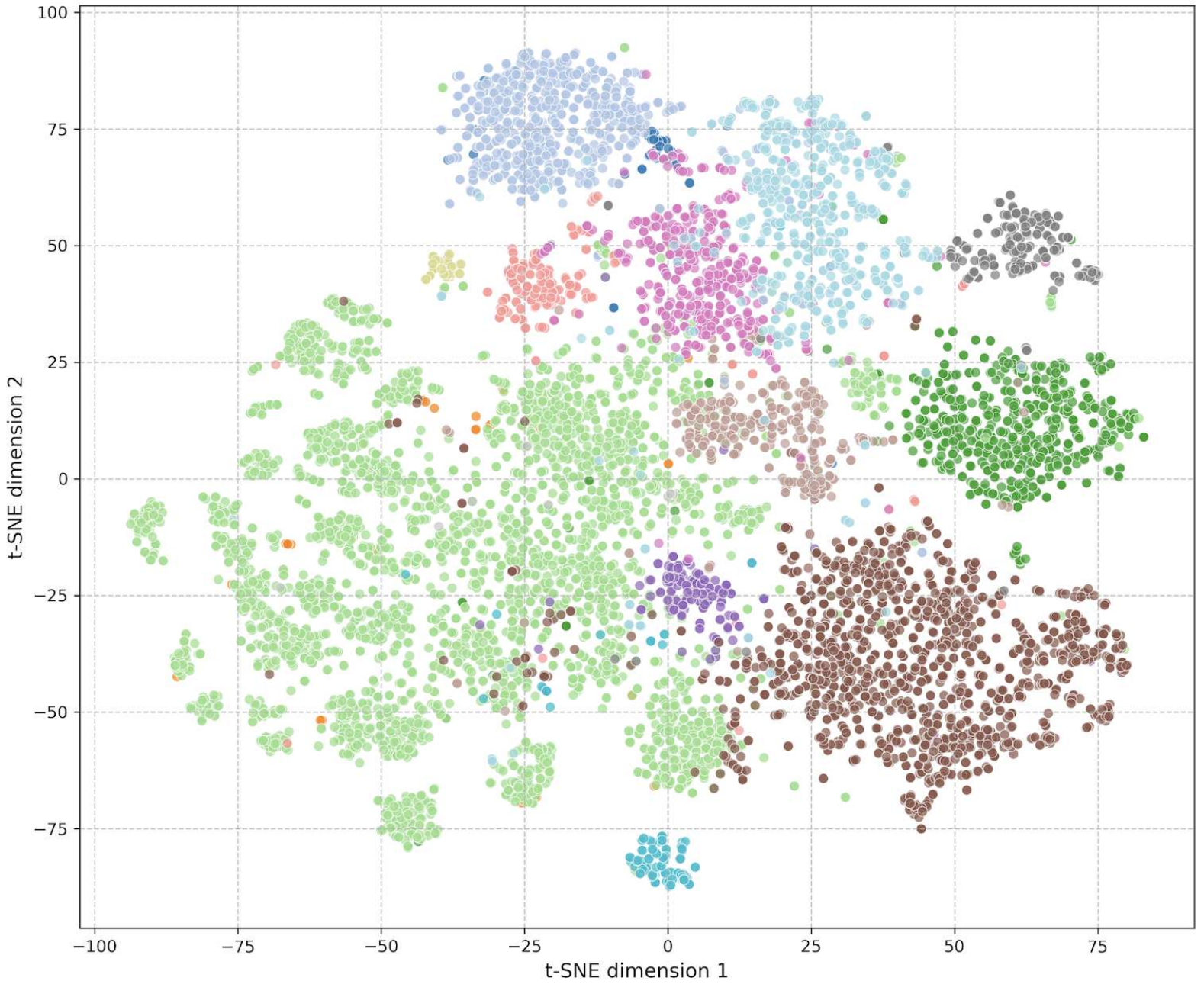}
            \caption{Marauder's Map}
            \label{fig:ours_wo15}
        \end{subfigure}
        \hfill
        \begin{subfigure}[t]{0.12\linewidth}
            \centering
            \includegraphics[width=\linewidth]{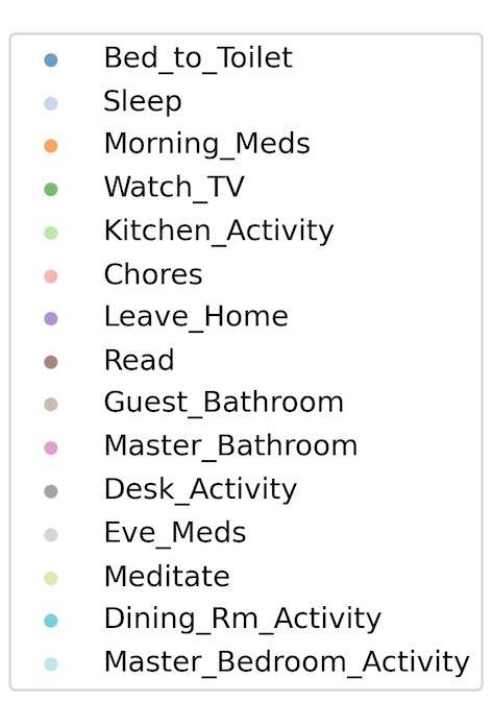}
            \label{fig:legend}
        \end{subfigure}
    \end{tabular}
    \caption{t-SNE visualizations of the learned feature representations on Milan dataset. The first row (a–d) shows the feature distributions for all 16 activity classes, including the "Other" class. The second row (e–h) excludes the "Other" class to emphasize the separation of meaningful activities. Compared methods include DeepCASAS (a, e), DCNN (b, f), TCN (c, g), and our proposed Marauder’s Map (d, h).}
    \label{fig:tsne_test}
\end{figure*}
\subsection{Cross-Activity Performance}
\begin{figure*}[h!]
    \centering
    \begin{tabular}{ccccc}
        \begin{subfigure}[t]{0.22\linewidth}
            \centering
            \includegraphics[width=\linewidth]{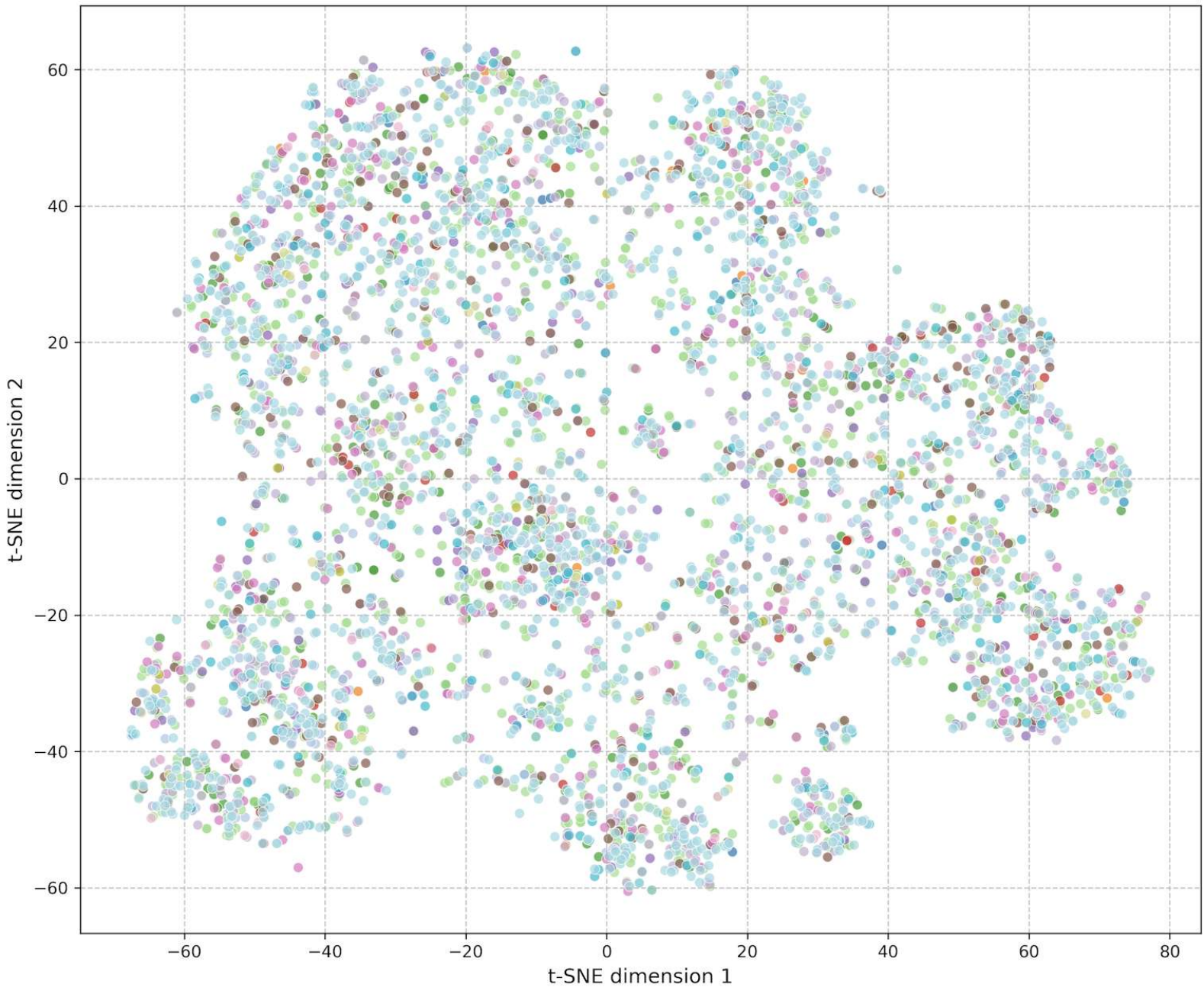}
            \caption{DeepCASAS}
            \label{fig:lstm_cross}
        \end{subfigure}
        \hfill
        \begin{subfigure}[t]{0.22\linewidth}
            \centering
            \includegraphics[width=\linewidth]{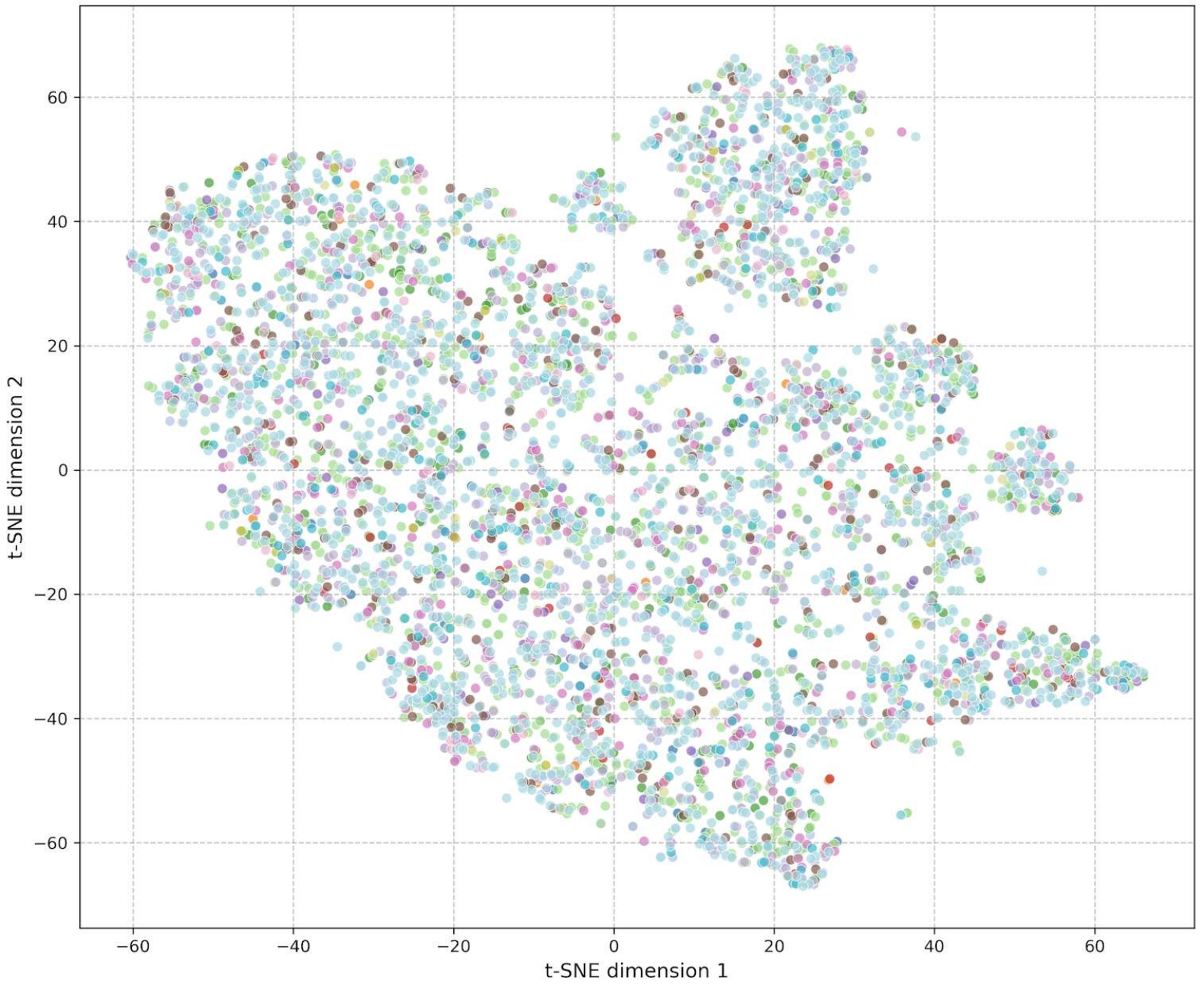}
            \caption{DCNN}
            \label{fig:dcnn_cross}
        \end{subfigure} 
        \hfill
        \begin{subfigure}[t]{0.22\linewidth}
            \centering
            \includegraphics[width=\linewidth]{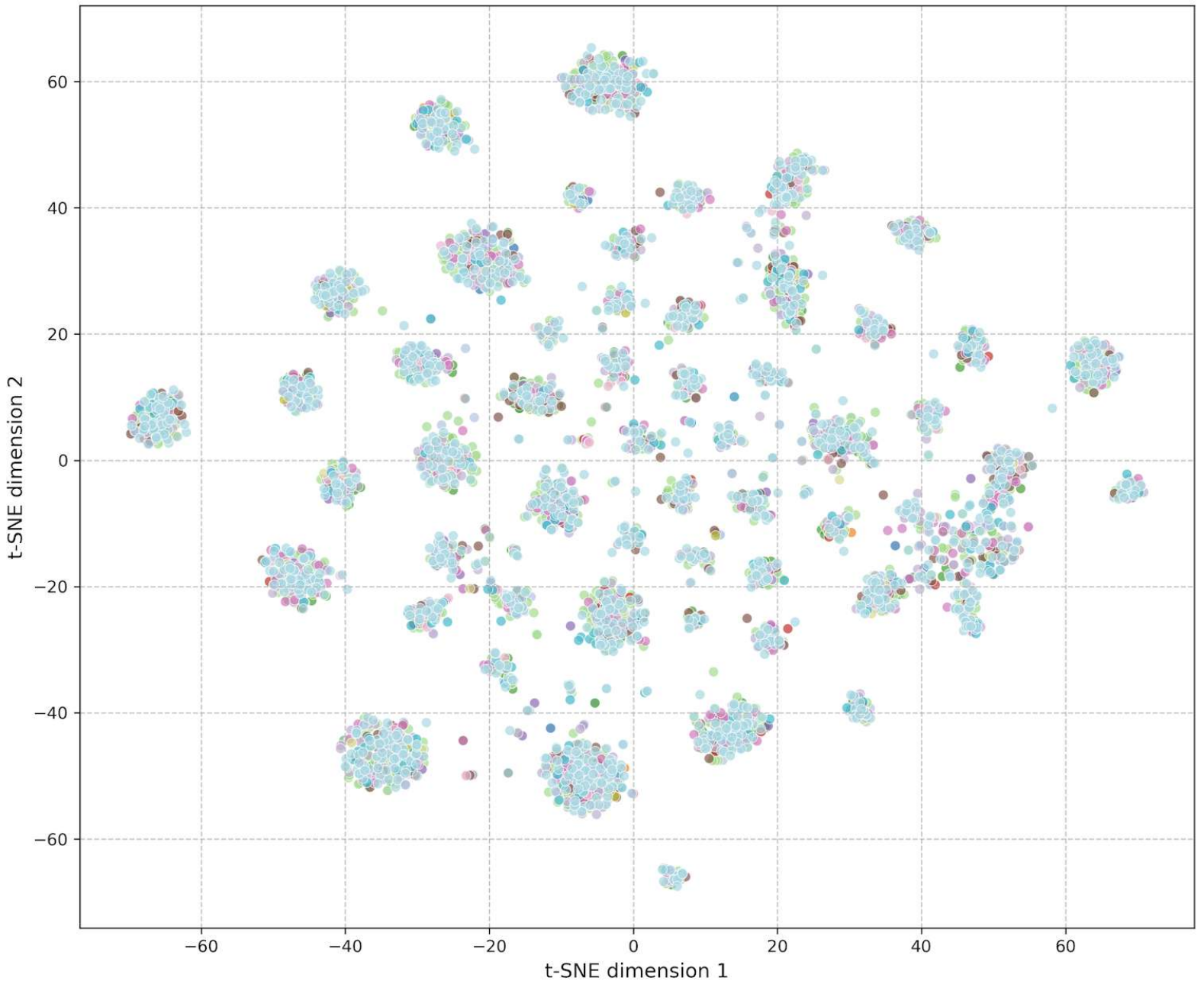}
            \caption{TCN}
            \label{fig:tcn_cross}
        \end{subfigure} 
        \hfill
        \begin{subfigure}[t]{0.22\linewidth}
            \centering
            \includegraphics[width=\linewidth]{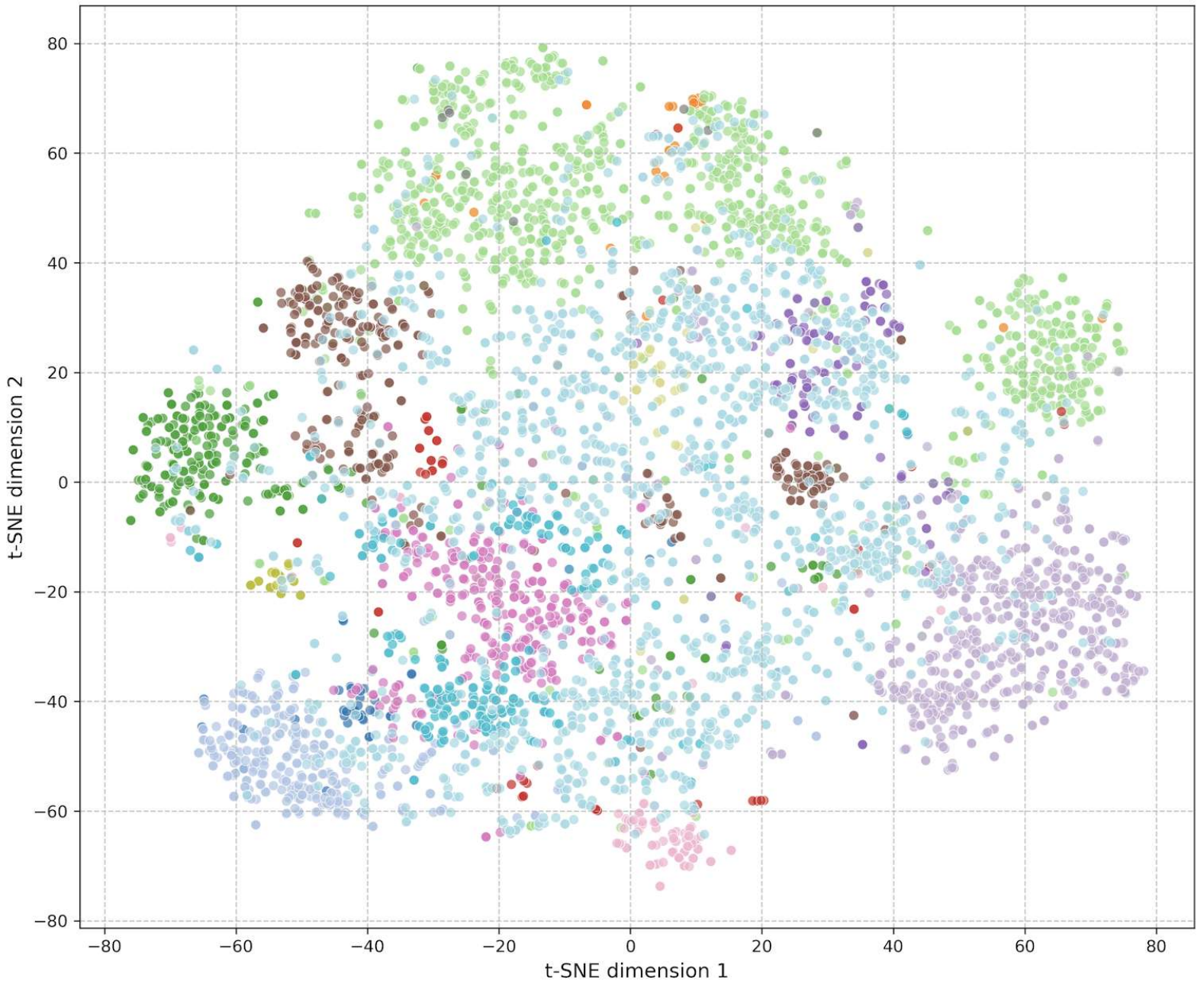}
            \caption{Marauder's Map}
            \label{fig:ours_cross}
        \end{subfigure} 
        \hfill
        \begin{subfigure}[t]{0.12\linewidth}
            \centering
            \includegraphics[width=\linewidth]{figures/tsne_legend.pdf}
            \label{fig:legend}
        \end{subfigure}\\
        \begin{subfigure}[t]{0.22\linewidth}
            \centering
            \includegraphics[width=\linewidth]{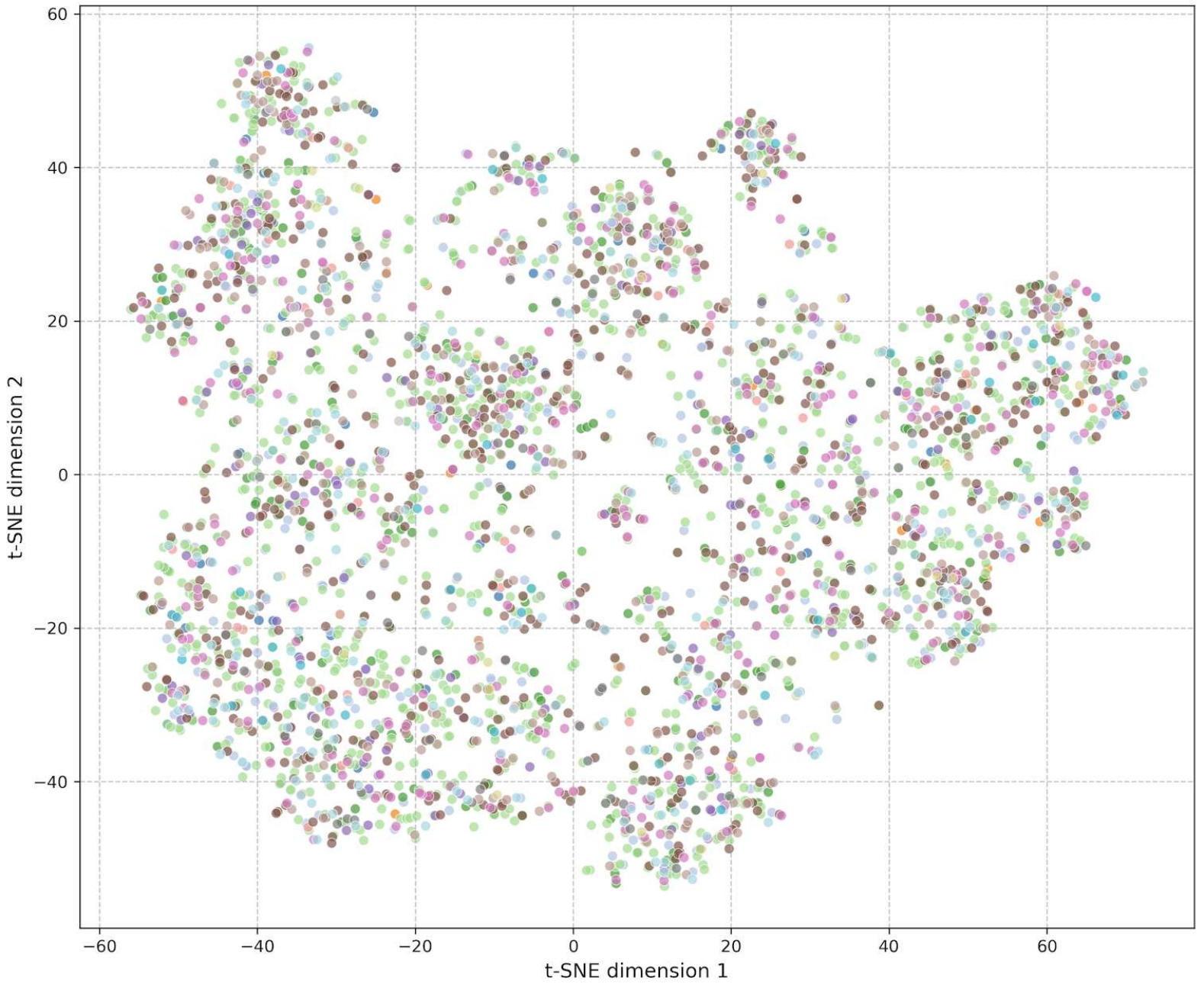}
            \caption{DeepCASAS}
            \label{fig:lstm_cross_wo15}
        \end{subfigure}
        \hfill
        \begin{subfigure}[t]{0.22\linewidth}
            \centering
            \includegraphics[width=\linewidth]{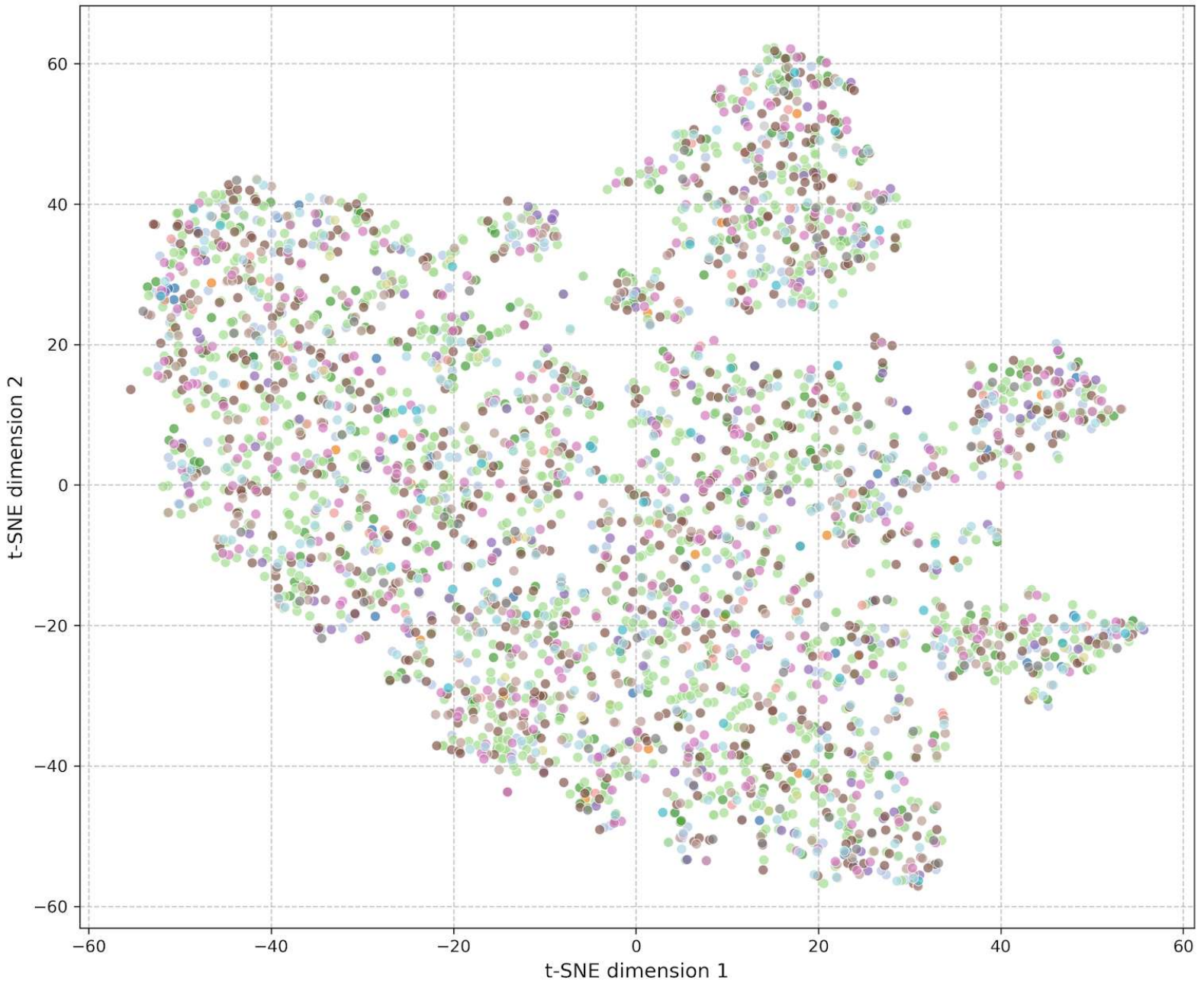}
            \caption{DCNN}
            \label{fig:dcnn_cross_wo15}
        \end{subfigure} 
        \hfill
        \begin{subfigure}[t]{0.22\linewidth}
            \centering
            \includegraphics[width=\linewidth]{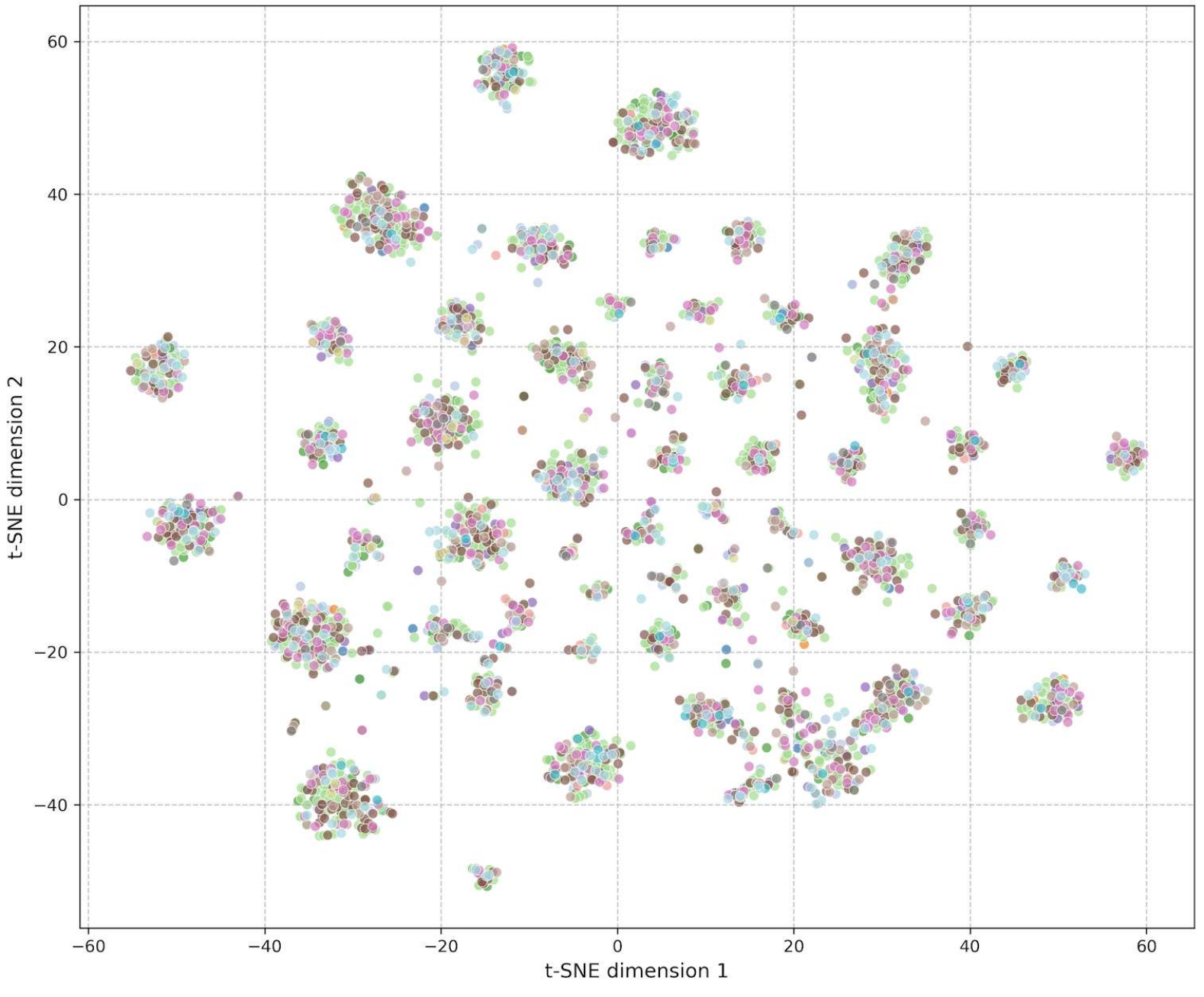}
            \caption{TCN}
            \label{fig:tcn_cross_wo15}
        \end{subfigure} 
        \hfill
        \begin{subfigure}[t]{0.22\linewidth}
            \centering
            \includegraphics[width=\linewidth]{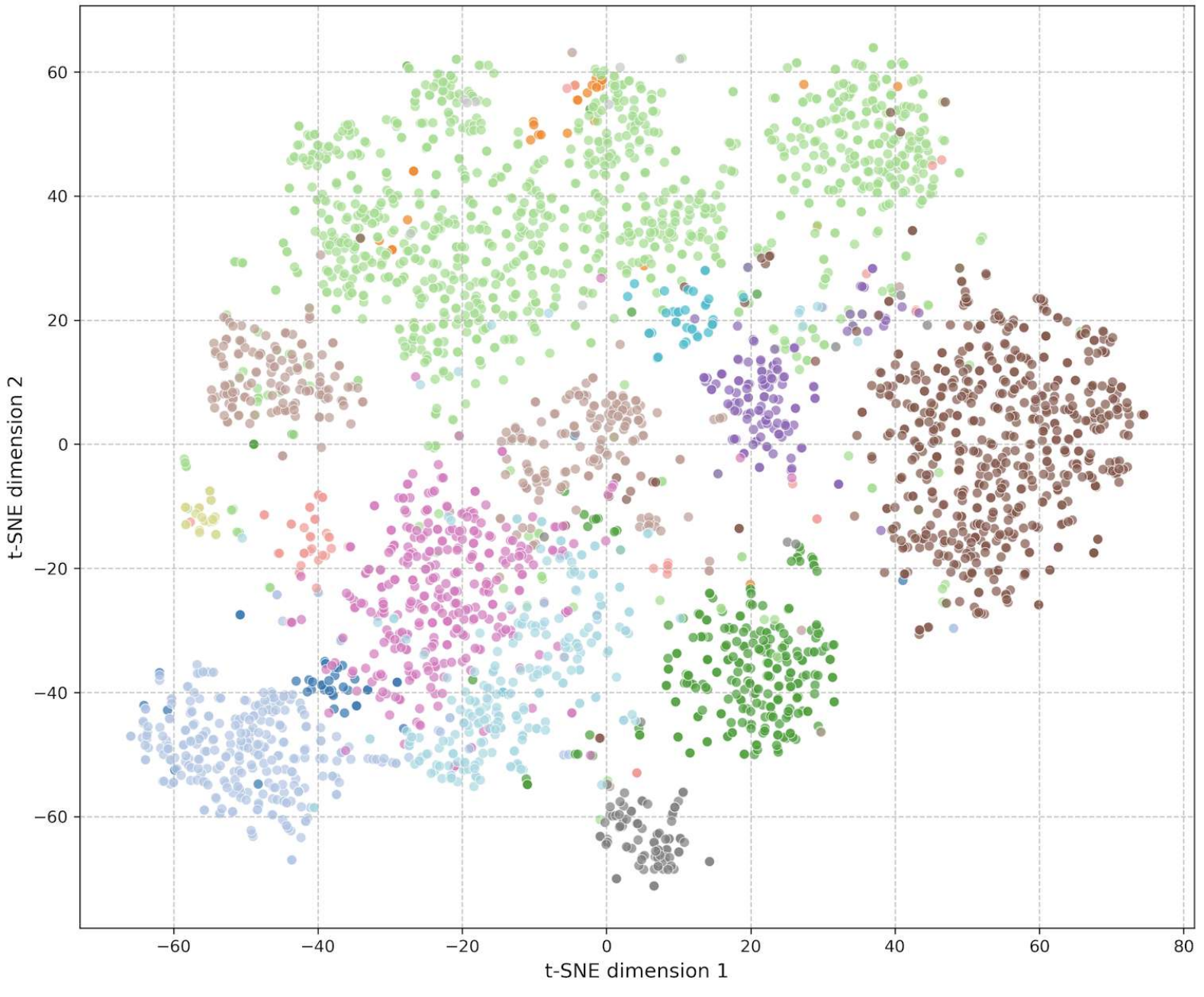}
            \caption{Marauder's Map}
            \label{fig:ours_cross}
        \end{subfigure}
        \hfill
        \begin{subfigure}[t]{0.12\linewidth}
            \centering
            \includegraphics[width=\linewidth]{figures/tsne_wo15_legend.pdf}
            \label{fig:legend}
        \end{subfigure}
    \end{tabular}
    \caption{t-SNE visualizations of the learned feature representations on the Milan dataset under cross-activity window settings. The first row (a–d) shows the feature distributions for all 16 activity classes, including the "Other" class. The second row (e–h) excludes the "Other" class to emphasize the separation of meaningful activities. Compared methods include DeepCASAS (a, e), DCNN (b, f), TCN (c, g), and our proposed Marauder’s Map (d, h).}
    \label{fig:tsne_cross}
\end{figure*}

To evaluate the robustness of different models under cross-activity windows where multiple activities may coexist within a single window, we conducted comprehensive performance comparisons and feature space analysis. The cross-activity dataset isolates windows that contain evidence of multiple concurrent or sequential activities, representing a more realistic and challenging evaluation scenario.  As shown in Table \ref{tab:crossperf}, our proposed methods, Marauder-Basic and Marauder-Time+Attn, consistently outperform all baseline approaches, including DeepCASAS, DCNN, and TCN, across the Milan, Kyoto7, and Aruba datasets. Specifically, Marauder-Time+Attn achieves the highest accuracy, F1 score, precision, and recall on all datasets, demonstrating strong generalization in complex multi-activity scenarios. To further investigate feature separability, we visualized the learned representations using t-SNE plots, presented in Fig. \ref{fig:tsne_cross}. In the first row of visualizations, which include all 16 classes, the feature spaces learned by baseline models are largely dominated by the "Other" class, resulting in overlapping and entangled clusters. The DeepCASAS and DCNN models (Figures 10a, 10b, 10e, and 10f) exhibit highly entangled feature spaces, where activity classes are heavily overlapped, suggesting that these models struggle to learn sufficiently discriminative features, especially in the presence of ambiguous activities. Although the TCN model (Figures 10c and 10g) demonstrates slightly better localized clusters, there remains noticeable scattering and overlap among classes. In stark contrast, the proposed Marauder’s Map model (Figures 10d and 10h) produces significantly more compact, well-separated clusters, indicating superior feature learning and robustness against inter-class ambiguity. Even with the challenging "Other" class included, Marauder’s Map maintains clear decision boundaries between activities. When excluding the "Other" class, its ability to distinctly map different activities becomes even more pronounced. These results highlight the effectiveness of incorporating temporal awareness and attention-guided representations in our framework, enabling better discrimination of nuanced activity patterns compared to traditional CNN- or LSTM-based baselines.
\begin{table}[h!]
  \caption{Performance Comparison of Different Methods on Cross-Activity Windows}
  \label{tab:crossperf}
  \centering
  \resizebox{1\columnwidth}{!}{
    \setlength{\tabcolsep}{1mm}{
  \begin{tabular}{ccccccccccccc}
    \toprule
    \multirow{2}{*}{Model} & \multicolumn{4}{c}{Milan} & \multicolumn{4}{c}{Kyoto7} & \multicolumn{4}{c}{Aruba} \\
    \cmidrule(lr){2-5} \cmidrule(lr){6-9} \cmidrule(lr){10-13}
    & Acc & F1 & Prec & Rec & Acc & F1 & Prec & Rec & Acc & F1 & Prec & Rec \\
    \midrule
    DeepCasas\cite{liciotti2020sequential} 
      & 72.37 $\pm$ 0.45 & 61.25 $\pm$ 1.36 & 68.04 $\pm$ 2.66 & 58.75 $\pm$ 1.27 
      & 86.17 $\pm$ 2.12 & 83.27 $\pm$ 3.83 & 85.09 $\pm$ 2.85 & 83.13 $\pm$ 3.83 
      & 84.50 $\pm$ 0.33 & 70.08 $\pm$ 1.45 & 81.80 $\pm$ 1.43 & 64.33 $\pm$ 1.45 \\
    1DCNN\cite{kiranyaz20211d}
      & 32.28 $\pm$ 0.80 & 7.80 $\pm$ 0.85 & 11.32 $\pm$ 2.45 & 8.96 $\pm$ 0.59
      & 34.28 $\pm$ 1.01 & 13.92 $\pm$ 0.56 & 15.15 $\pm$ 1.46 & 18.00 $\pm$ 0.71
      & 46.88 $\pm$ 0.70 & 9.95 $\pm$ 0.66 & 16.14 $\pm$ 3.32 & 10.98 $\pm$ 0.37 \\
    DCNN\cite{gochoo2018unobtrusive} 
      & 64.23 $\pm$ 0.87 & 49.42 $\pm$ 0.92 & 54.29 $\pm$ 1.79 & 47.58 $\pm$ 1.29 
      & 80.73 $\pm$ 0.46 & 67.00 $\pm$ 3.14 & 67.90 $\pm$ 3.69 & 66.96 $\pm$ 2.89 
      & 78.77 $\pm$ 0.42 & 53.91 $\pm$ 1.90 & 65.13 $\pm$ 5.87 & 49.55 $\pm$ 2.03 \\
    greyDCNN\cite{mohmed2020employing} 
      & 62.21 $\pm$ 0.45 & 45.53 $\pm$ 0.34 & 51.12 $\pm$ 2.87 & 44.61 $\pm$ 1.10 
      & 75.29 $\pm$ 0.37 & 62.23 $\pm$ 2.69 & 64.06 $\pm$ 2.29 & 61.85 $\pm$ 2.78 
      & 77.23 $\pm$ 0.27 & 51.40 $\pm$ 1.22 & 62.14 $\pm$ 3.89 & 47.02 $\pm$ 1.23 \\
    LSTM-CNN\cite{xia2020lstm}
      & 59.00 $\pm$ 0.38 & 39.54 $\pm$ 1.70 & 49.93 $\pm$ 3.26 & 36.67 $\pm$ 1.81
      & 61.31 $\pm$ 1.59 & 43.59 $\pm$ 2.10 & 46.29 $\pm$ 4.42 & 43.84 $\pm$ 1.76
      & 76.49 $\pm$ 0.74 & 52.05 $\pm$ 1.10 & 61.78 $\pm$ 3.67 & 48.15 $\pm$ 1.37 \\
    Transformer\cite{vaswani2017attention} 
      & 65.10 $\pm$ 0.72 & 45.90 $\pm$ 1.49 & 49.70 $\pm$ 0.96 & 45.04 $\pm$ 1.83
      & 83.66 $\pm$ 1.04 & 80.16 $\pm$ 2.24 & 83.44 $\pm$ 1.81 & 79.41 $\pm$ 2.76 
      & 77.63 $\pm$ 0.74 & 51.18 $\pm$ 2.49 & 58.10 $\pm$ 1.28 & 47.80 $\pm$ 4.23 \\  
    TCN\cite{bai2018empirical} 
      & \underline{77.21 $\pm$ 1.80} & \underline{63.70 $\pm$ 3.28} & \underline{71.72 $\pm$ 7.52} & \underline{61.63 $\pm$ 2.56} 
      & \underline{86.26 $\pm$ 2.20} & \underline{83.10 $\pm$ 1.46} & \underline{85.90 $\pm$ 1.90} & \underline{82.16 $\pm$ 1.42} 
      & \underline{88.68 $\pm$ 0.87} & \underline{77.13 $\pm$ 1.50} & \textbf{87.05 $\pm$ 0.78} & \underline{71.59 $\pm$ 2.14} \\    
    \midrule
    Marauder-Basic 
      & \textbf{88.00 $\pm$ 1.12} & \textbf{81.75 $\pm$ 1.08} & \textbf{83.99 $\pm$ 1.88} & \textbf{80.58 $\pm$ 1.05} 
      & \textbf{91.94 $\pm$ 0.15} & \textbf{90.66 $\pm$ 1.19} & \textbf{90.76 $\pm$ 1.04} & \textbf{90.96 $\pm$ 1.18} 
      & \textbf{90.45 $\pm$ 0.48} & \textbf{82.12 $\pm$ 0.14} & \underline{85.60 $\pm$ 0.78} & \textbf{79.97 $\pm$ 0.00} \\
    Marauder-Time+Attn 
      & \cellcolor{green!20}\textbf{88.43 $\pm$ 0.93} & \cellcolor{green!20}\textbf{83.85 $\pm$ 2.67} & \cellcolor{green!20}\textbf{84.65 $\pm$ 1.67} & \cellcolor{green!20}\textbf{83.78 $\pm$ 3.46} 
      & \cellcolor{green!20}\textbf{92.54 $\pm$ 1.30} & \cellcolor{green!20}\textbf{91.49 $\pm$ 1.11} & \cellcolor{green!20}\textbf{92.36 $\pm$ 1.55} & \cellcolor{green!20}\textbf{90.96 $\pm$ 0.77} 
      & \cellcolor{green!20}\textbf{92.81 $\pm$ 0.25} & \cellcolor{green!20}\textbf{84.96 $\pm$ 0.26} & \cellcolor{green!20}\textbf{88.88 $\pm$ 0.65} & \cellcolor{green!20}\textbf{83.00 $\pm$ 0.49} \\
    \bottomrule
  \end{tabular}}}
\end{table}

\subsection{Ablation Study}
To better understand the contributions of each component in our proposed architecture, we conduct a series of ablation studies aimed at evaluating their individual and combined effects on overall performance. Specifically, we investigate how various configurations of the image emulator impact the quality of the generated representations. This includes evaluating the effects of different fading schemes, image resolutions, spot sizes, spot types, and background styles. Additionally, we assess the influence of temporal encoding by isolating and combining different time components—such as month, day, weekday, hour, and minute—to determine which elements are most informative for activity recognition. Finally, we examine the role of the attention mechanism in our sequential classification pipeline, exploring its effectiveness in capturing dependencies across sensor activations over time. These experiments provide comprehensive insights into the design choices behind our framework and highlight the key factors driving its performance.

\subsubsection{Layout-based Image Emulator Effectiveness}
To encode spatiotemporal information in smart home activity logs, we propose a layout-based image emulator that transforms sequences of sensor activations into visual representations over a 2D floorplan. Within each fixed-size temporal window, individual sensor events are projected as graphical elements (e.g., spots or shapes) onto their corresponding physical locations on the layout. The appearance of each element—including its color, shape—is designed to encode various contextual attributes such as sensor type, trigger intensity, activation frequency, or role in the activity.

In our default configuration, all sensor activations within a window are represented with uniform intensity, resulting in clear, spatially faithful images that emphasize the distribution and intensity of sensor activity. We refer to this as the no-fading baseline, which focuses solely on spatial and categorical encoding. To explore the potential benefits of temporal encoding, we also implement a fading-based variant, where earlier events are displayed with reduced brightness or opacity to convey event recency. We evaluate both approaches using the same model architecture across a standard test set and a cross-activity generalization set. As shown in Table~\ref{tab:fading_scheme}, the no-fading method consistently outperforms the fading-based variant across all metrics. Specifically, it achieves higher accuracy (92.74\% vs. 89.81\%) and F1-score (86.27 vs. 84.22) in the standard setting, with similar trends observed in the cross-activity evaluation. These results suggest that while fading intuitively encodes temporal information, it may introduce visual noise that obscures critical patterns, ultimately hindering model performance. Notably, fading-based representations yield lower standard deviations, implying more stable but suboptimal learning dynamics. Overall, retaining uniform intensity in sensor activations leads to more discriminative and effective trajectory representations for activity recognition.
\begin{table*}[h!]
  \centering
  \caption{Comparison of Different Image Emulator}
  \resizebox{1\columnwidth}{!}{
    \setlength{\tabcolsep}{1mm}{
  \begin{tabular}{l|cccc|cccc}
    \toprule
    \multicolumn{1}{c|}{\textbf{Image Emulator}} & \multicolumn{4}{c|}{\textbf{Test Setting}} & \multicolumn{4}{c}{\textbf{Cross-Activity Test Setting}} \\
    \cmidrule(lr){2-5} \cmidrule(lr){6-9}
    & Accuracy & F1 & Precision & Recall & Accuracy & F1 & Precision & Recall \\
    \midrule
    fading image sequence
      & 89.81 $\pm$ 0.23 & 84.22 $\pm$ 0.78 & 85.99 $\pm$ 1.35 & 83.44 $\pm$ 0.74
      & 85.28 $\pm$ 0.25 & 81.58 $\pm$ 0.61 & 84.23 $\pm$ 0.60 & 80.35 $\pm$ 0.64\\
    no-fading image sequence 
      & \textbf{92.74 $\pm$ 0.90} & \textbf{86.27 $\pm$ 2.51} & \textbf{87.04 $\pm$ 1.34} & \textbf{86.08 $\pm$ 3.36}
      & \textbf{88.43 $\pm$ 0.93} & \textbf{83.85 $\pm$ 2.67} & \textbf{84.65 $\pm$ 1.67} & \textbf{83.78 $\pm$ 3.46} \\
    \bottomrule
  \end{tabular}}} 
  \label{tab:fading_scheme}
\end{table*}

Regardless of the fading mechanism, we conducted a series of controlled experiments to evaluate how different visual encoding configurations affect model performance within our image emulator framework. These experiments were grouped into four categories, each isolating a specific parameter dimension: spot type, image resolution, and background style. By varying one factor at a time while holding others constant, we aimed to gain insight into the role of visual design choices in shaping the quality of spatial-temporal representations learned by the model.\\
\begin{figure*}[h!]
    \centering
    \begin{tabular}{ccc}
        \begin{subfigure}[t]{0.3\linewidth}
            \centering
            \includegraphics[width=\linewidth]{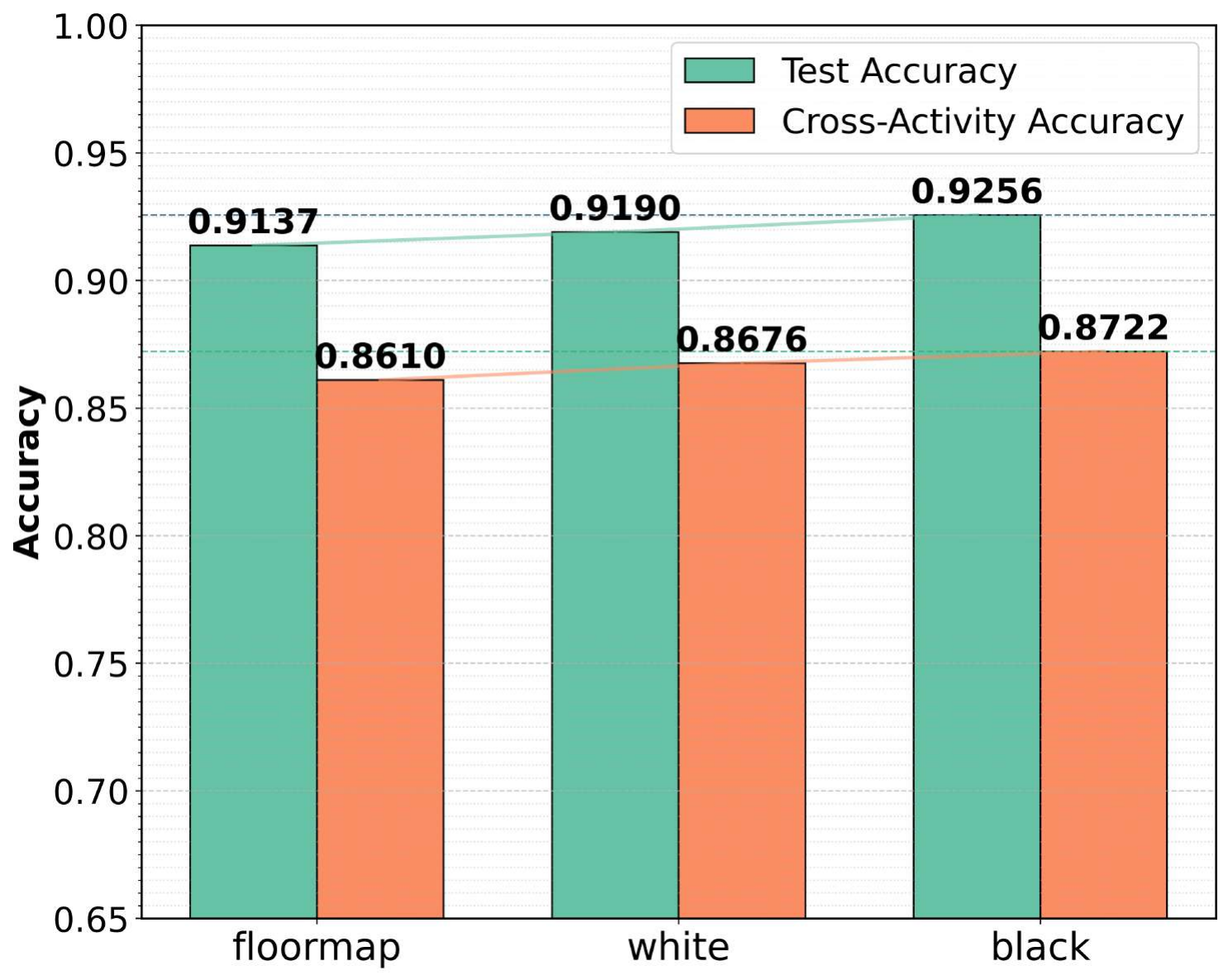}
            \caption{Background Types}
            \label{fig:backtype}
        \end{subfigure}
        \hfill
        \begin{subfigure}[t]{0.3\linewidth}
            \centering
            \includegraphics[width=\linewidth]{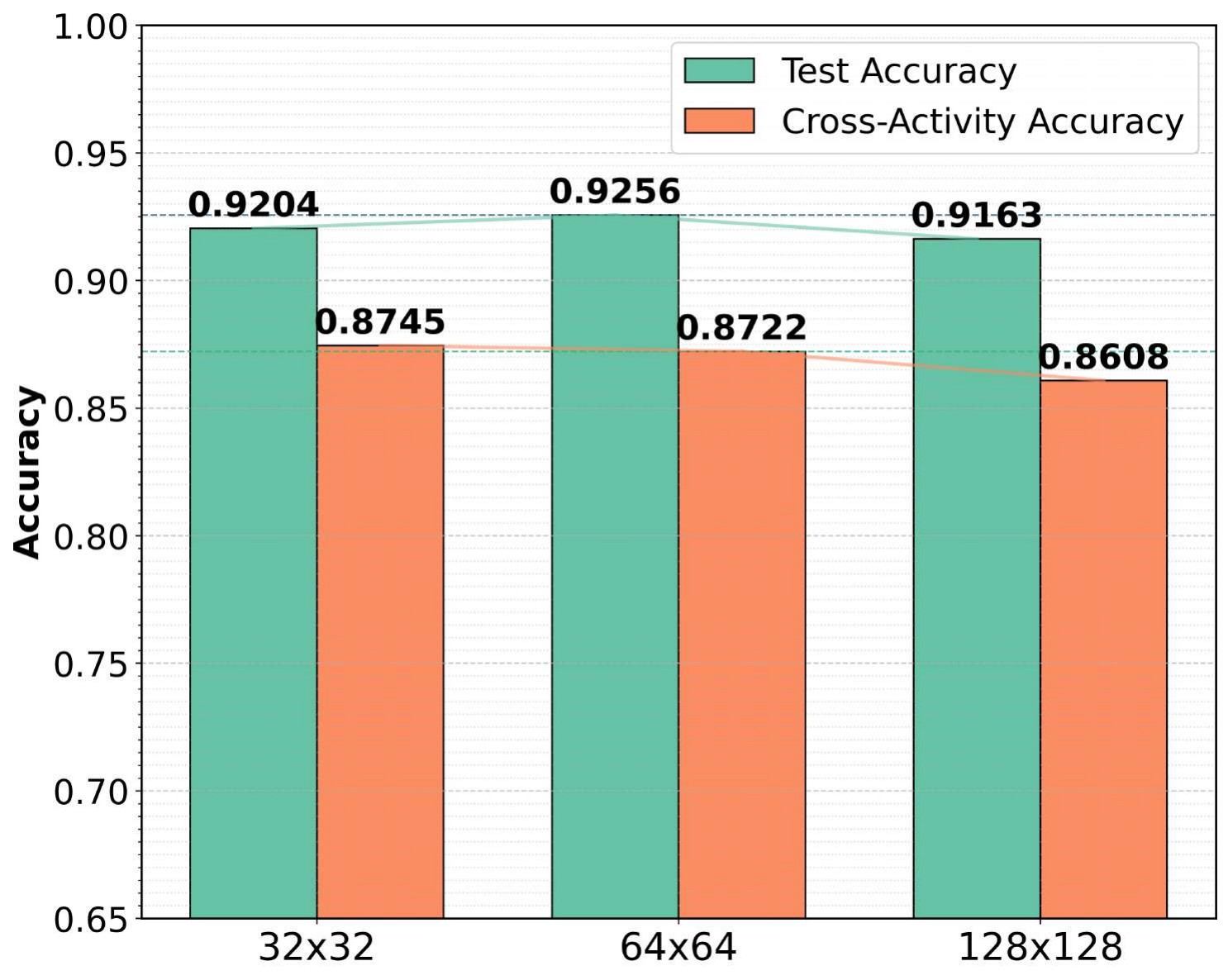}
            \caption{Image Sizes}
            \label{fig:imgsize}
        \end{subfigure} 
        \hfill
        \begin{subfigure}[t]{0.3\linewidth}
            \centering
            \includegraphics[width=\linewidth]{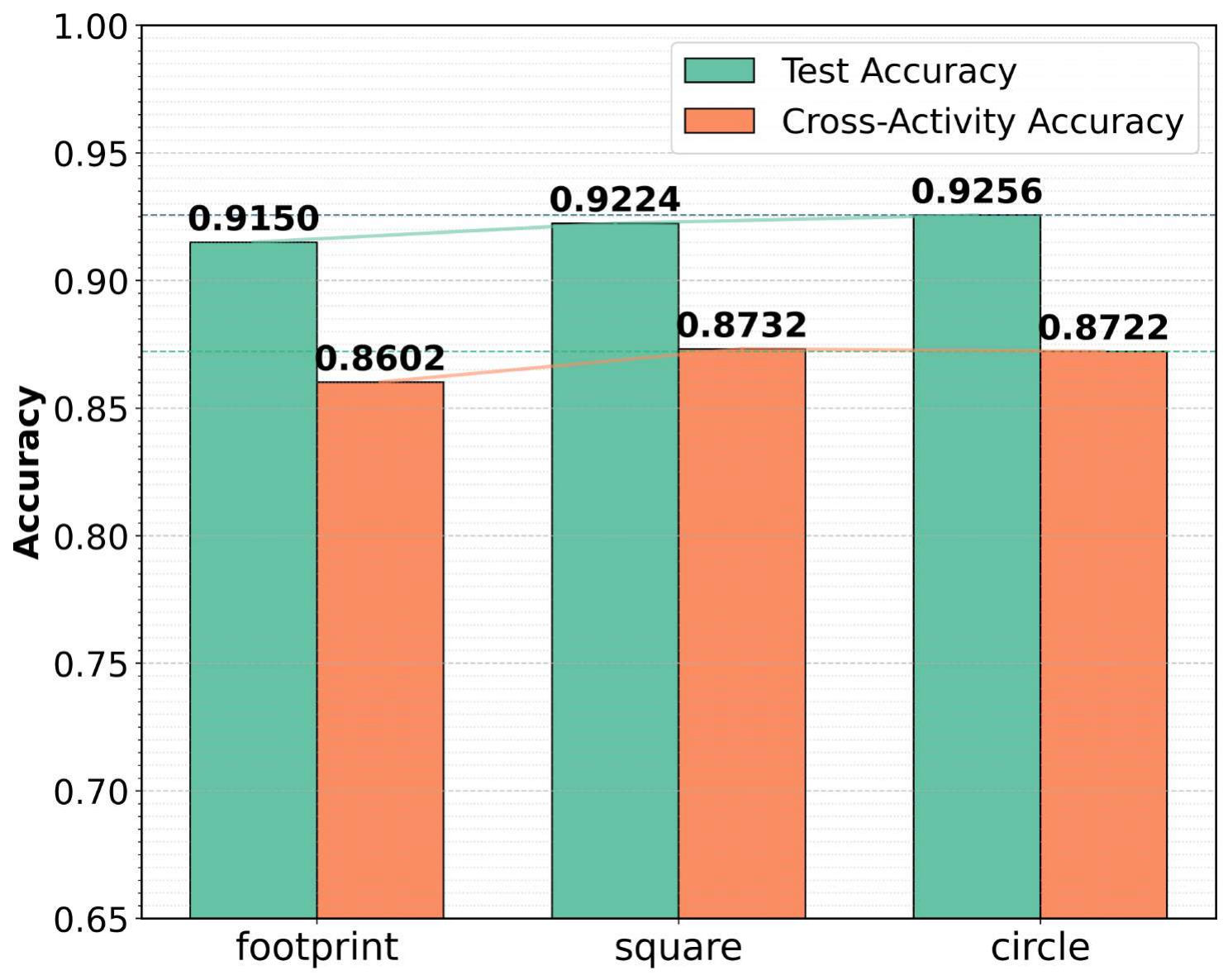}
            \caption{Spot Types}
            \label{fig:spottype}
        \end{subfigure} 
    \end{tabular}
    \caption{Comparison of emulator performance across different visual configuration settings. Subfigures show test accuracy and cross-activity accuracy under variations in: (a) background types (floorplan, white, black), (b) image sizes (32×32, 64×64, 128×128), and (c) spot types (footprint, square, circle).}
    \label{fig:emulators}
\end{figure*}
To evaluate the role of background aesthetics, we compared three types: a default black background, a plain white background, and a realistic floorplan incorporating architectural details such as walls and furniture. Sensor locations were fixed across all variants to ensure that only the visual styling differed. As illustrated in Fig.~\ref{fig:backtype}, the use of a realistic floorplan did not yield notable performance gains compared to the plain black background. This indicates that contextual cues from the background contribute minimally to the model's decision-making; instead, the trajectory patterns themselves appear to carry the core discriminative information. Notably, a performance drop was observed with the white background. We attribute this to reduced contrast between trajectory lines and the background, which may obscure sensor activation paths and impair the model’s ability to extract salient features. In contrast, the black background offers high visual contrast that better highlights movement patterns. These findings suggest that while background design can enhance interpretability for user-facing applications, it has limited impact on classification accuracy, making the black background a practical default for deployment.

We further investigated whether the geometric form of sensor activations influences performance by comparing circle, square, and footprint-shaped markers. All other parameters, including background style and sensor layout, were held constant to isolate the effect of spot shape. As shown in Fig.~\ref{fig:spottype}, circles and squares achieved comparable results, suggesting the model is largely invariant to basic geometric shapes. However, the footprint markers—despite being more semantically expressive—slightly reduced performance. This could be due to their increased visual complexity, which may distract the model from learning spatial trajectories effectively. Overall, our results indicate that shape semantics play a minimal role in recognition accuracy, and simpler marker shapes are preferable when clarity and consistency are priorities. That said, footprint markers may still be valuable in scenarios emphasizing user interpretability, such as elder care interfaces or visual explanation tools.

To understand how spatial granularity affects model learning, we evaluated three image sizes: 32×32, 64×64 (default), and 128×128 pixels. All experiments used fixed-size circular markers to isolate the impact of resolution. As shown in Fig.~\ref{fig:imgsize}, classification performance peaked at the default 64×64 resolution. Increasing the resolution to 128×128 led to marginally lower accuracy, likely due to increased sparsity—sensor spots become more dispersed, diluting temporal coherence and spatial density, thus complicating trajectory learning. Conversely, reducing the resolution to 32×32 increased visual density but compromised spatial precision, causing sensor activations to cluster and blur, which undermines the model’s ability to distinguish fine-grained movement patterns. These results highlight a trade-off between spatial clarity and semantic resolution: too high a resolution may obscure temporal consistency, while too low a resolution reduces spatial fidelity. Therefore, selecting an optimal image resolution is essential for balancing learnability and efficiency, and should be adapted to the scale and complexity of the deployment environment.

In summary, our analyses demonstrate that design choices related to visual encoding influence the interpretability and performance of activity recognition models to varying degrees. While background and marker shape offer flexibility with minimal performance trade-offs, resolution must be carefully tuned to preserve both spatial structure and temporal continuity. These insights guide the practical design of trajectory-encoded HAR systems for diverse deployment settings.

\subsubsection{Comparing Temporal Information Combination Effectiveness}
To investigate the impact of temporal features on activity classification, we design a series of controlled experiments by selectively including different combinations of time-related components. Specifically, we consider six temporal dimensions: Month, Day, Weekday, Hour, and Minute, each representing different granularities of temporal context. As shown in Table \ref{tab:timeperf}, we start with the baseline model without any temporal input and progressively incorporate individual components to assess their standalone contributions. We then examine combinations of temporal features to capture complementary effects between coarse-grained (e.g., Month, Weekday) and fine-grained (e.g., Hour, Minute) temporal cues. This setup enables us to isolate which temporal signals are most informative for improving classification performance in the Milan smart home dataset, measured across Accuracy, F1-score, Precision, and Recall.
\begin{table*}[h!]
    \centering
    \caption{Comparison of Different Temporal Information Combinations on Milan Dataset (Accuracy, F1-Score, Precision, and Recall)}
    \resizebox{0.8\textwidth}{!}{
    \begin{tabular}{ccccc|cccc}
        \toprule
        \multicolumn{5}{c|}{Temporal Components} & \multicolumn{4}{c}{Milan} \\
        \cmidrule(lr){1-5} \cmidrule(lr){6-9}
        Month & Day & Weekday & Hour & Minute & Acc & F1 & Prec & Rec \\
        \midrule
        \textcolor{red}{X} & \textcolor{red}{X} & \textcolor{red}{X} & \textcolor{red}{X} & \textcolor{red}{X} & 89.97 $\pm$ 0.89 & 83.53 $\pm$ 1.41 & 85.12 $\pm$ 1.18 & 83.01 $\pm$ 1.49 \\
        \textcolor{green}{Y} & \textcolor{red}{X} & \textcolor{red}{X} & \textcolor{red}{X} & \textcolor{red}{X} & 90.86 $\pm$ 0.18 & 86.08 $\pm$ 0.78$^\ddagger$ & \underline{87.15 $\pm$ 1.10} & 85.54 $\pm$ 1.02 \\
        \textcolor{red}{X} & \textcolor{green}{Y} & \textcolor{red}{X} & \textcolor{red}{X} & \textcolor{red}{X} & 90.62 $\pm$ 0.53 & 84.94 $\pm$ 0.81 & 86.57 $\pm$ 1.01 & 84.29 $\pm$ 1.10 \\
        \textcolor{red}{X} & \textcolor{red}{X} & \textcolor{green}{Y} & \textcolor{red}{X} & \textcolor{red}{X} & 90.40 $\pm$ 0.60 & 83.75 $\pm$ 1.41 & 85.08 $\pm$ 0.08 & 83.75 $\pm$ 1.97 \\
        \textcolor{red}{X} & \textcolor{red}{X} & \textcolor{red}{X} & \textcolor{green}{Y} & \textcolor{red}{X} & 90.75 $\pm$ 1.70 & 84.44 $\pm$ 2.75 & 86.33 $\pm$ 1.82 & 83.61 $\pm$ 2.89 \\
        \textcolor{red}{X} & \textcolor{red}{X} & \textcolor{red}{X} & \textcolor{red}{X} & \textcolor{green}{Y} & 90.91 $\pm$ 0.44 & 84.41 $\pm$ 1.29 & 85.95 $\pm$ 0.71 & 83.86 $\pm$ 2.19 \\
        \cmidrule(lr){1-5} \cmidrule(lr){6-9}
        \textcolor{green}{Y} & \textcolor{green}{Y} & \textcolor{red}{X} & \textcolor{red}{X} & \textcolor{red}{X} & \underline{92.64 $\pm$ 0.37} & 85.97 $\pm$ 0.75 & 87.10 $\pm$ 1.09$^\ddagger$ & 85.48 $\pm$ 2.04 \\
        \rowcolor{LightGreen} \textcolor{red}{X} & \textcolor{red}{X} & \textcolor{green}{Y} & \textcolor{green}{Y} & \textcolor{red}{X} & 92.55 $\pm$ 0.07$^\ddagger$ & \textbf{87.00 $\pm$ 0.67} & 86.97 $\pm$ 0.92 & \textbf{87.53 $\pm$ 0.65} \\
        \textcolor{red}{X} & \textcolor{red}{X} & \textcolor{green}{Y} & \textcolor{red}{X} & \textcolor{green}{Y} & 91.99 $\pm$ 0.57 & 85.61 $\pm$ 1.82 & \textbf{87.15 $\pm$ 0.73} & 85.06 $\pm$ 2.87 \\
        \textcolor{red}{X} & \textcolor{red}{X} & \textcolor{red}{X} & \textcolor{green}{Y} & \textcolor{green}{Y} & 92.02 $\pm$ 0.47 & 85.68 $\pm$ 0.60 & 86.89 $\pm$ 0.98 & 85.59 $\pm$ 1.50 \\
        \cmidrule(lr){1-5} \cmidrule(lr){6-9}
        \textcolor{red}{X} & \textcolor{red}{X} & \textcolor{green}{Y} & \textcolor{green}{Y} & \textcolor{green}{Y} & \textbf{92.68 $\pm$ 0.20} & 86.01 $\pm$ 0.96 & 85.98 $\pm$ 0.82 & \underline{86.66 $\pm$ 1.24} \\
        \textcolor{green}{Y} & \textcolor{red}{X} & \textcolor{green}{Y} & \textcolor{green}{Y} & \textcolor{green}{Y} & 90.91 $\pm$ 0.58  & 84.00 $\pm$ 1.32 & 86.42 $\pm$ 0.95 & 82.55 $\pm$ 1.64 \\
        \textcolor{green}{Y} & \textcolor{green}{Y} & \textcolor{green}{Y} & \textcolor{green}{Y} & \textcolor{green}{Y} & 92.25 $\pm$ 0.70 & \underline{86.23 $\pm$ 1.60} & 86.47 $\pm$ 0.79 & 86.54 $\pm$ 2.40$^\ddagger$ \\
        \bottomrule
    \end{tabular}}
    \label{tab:timeperf}
\end{table*}

As shown in Table~\ref{tab:timeperf}, without any temporal information, the model achieves an accuracy of 89.97\%, already outperforming the strong baseline DeepCasas, which records an accuracy of 75.93\% (see Table~\ref{tab:overallperf}). Incorporating simple temporal cues, such as the weekday or hour alone, leads to consistent performance improvements, with the model reaching 90.75\% accuracy when using only hour encoding. When combining multiple temporal components, performance improves further. The best overall configuration—weekday, hour, and minute—achieves the highest accuracy of 92.68\% and an F1-score of 86.01\%. This improvement is particularly meaningful given the class imbalance in the dataset, where F1 score serves as a more reliable metric than accuracy. Notably, the combination of just weekday and hour yields a slightly lower accuracy (92.55\%) but a highest F1 score (87.00\%) and the best recall (87.53\%), making it a compelling alternative when balancing between model complexity and performance. These results highlight the value of incorporating temporally rich features and suggest that even moderately fine-grained time embeddings can substantially enhance recognition accuracy and robustness in real-world smart home data.
\begin{figure}[h!]
    \centering
    \begin{tabular}{ccc}
        \begin{subfigure}[t]{0.33\linewidth}
            \centering
            \includegraphics[width=\linewidth]{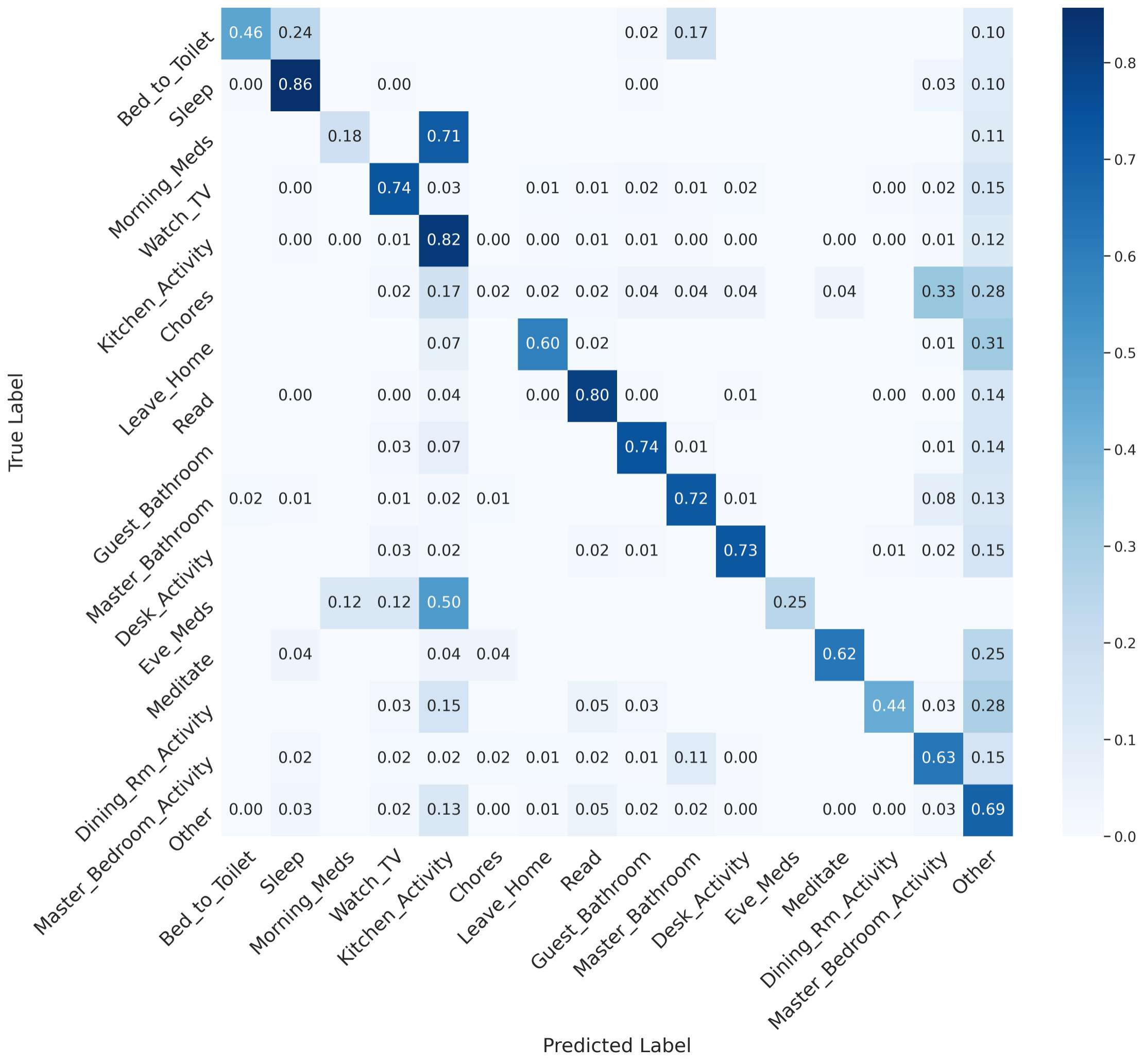}
            \caption{Deepcasas}
            \label{fig:lstm_cm}
        \end{subfigure}
        \hfill
        \begin{subfigure}[t]{0.33\linewidth}
            \centering
            \includegraphics[width=\linewidth]{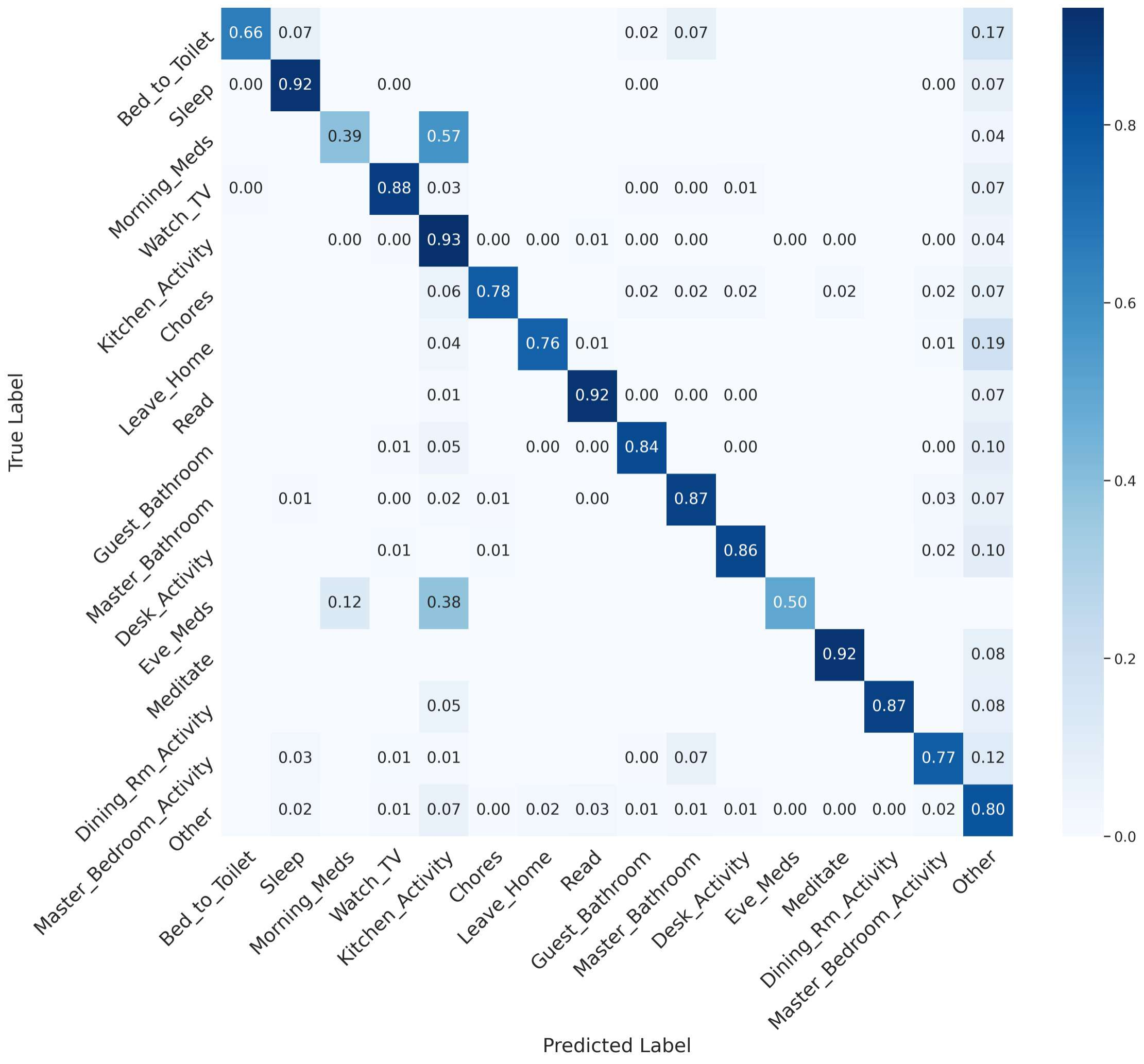}
            \caption{Marauder's Map w/o Time Embedding}
            \label{fig:wotime_cm}
        \end{subfigure}
        \hfill
        \begin{subfigure}[t]{0.33\linewidth}
            \centering
            \includegraphics[width=\linewidth]{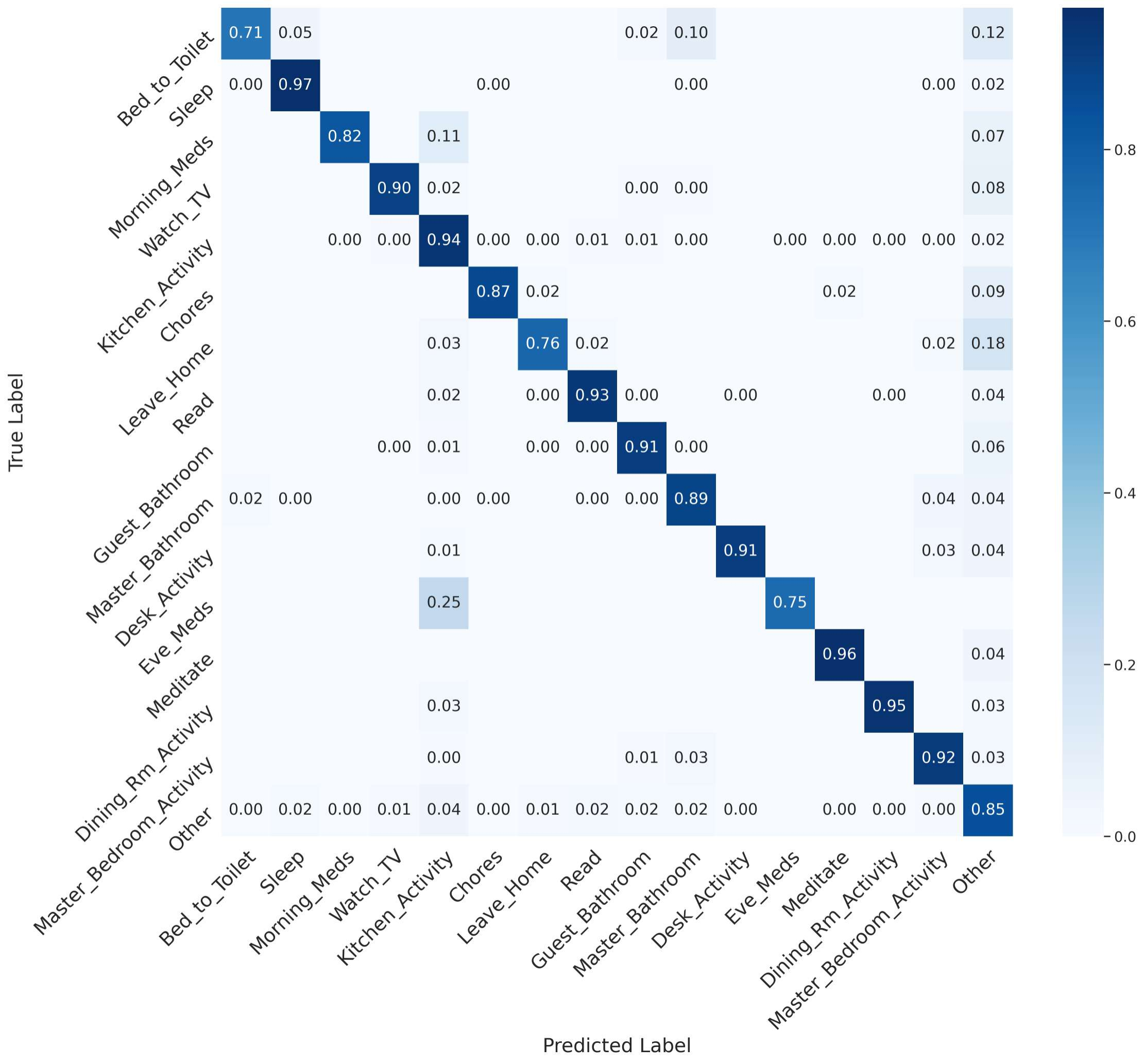}
            \caption{Marauder's map encoded with Weekday, Hour, Minute}
            \label{fig:time_cm}
        \end{subfigure} 
    \end{tabular}
    \caption{Confusion matrices of classification results on cross-activity test data, comparing models without (a) and with (b) the Temporal Encoding. The attention-based model shows improved diagonal dominance, indicating enhanced accuracy and reduced misclassification across most activity classes.}
    \label{fig:com_time_cm}
\end{figure}

The confusion matrices in Fig.\ref{fig:com_time_cm} further illustrate these trends. Without time embedding (Fig.\ref{fig:wotime_cm}), the model struggles to distinguish temporally similar activities occurring in the same spatial regions, such as "Morning Meds" and "Evening Meds," which are both kitchen-centered activities. With temporal encoding (Fig.\ref{fig:time_cm}), the confusion between such activities is substantially reduced, leading to more confident and accurate predictions. Compared to DeepCasas (Fig.\ref{fig:lstm_cm}), our proposed model with temporal embeddings exhibits sharper diagonal dominance in the confusion matrix, confirming enhanced class separability.

These results validate that capturing periodic and contextual temporal information—especially through weekday, hour, and minute encoding—substantially improves the discriminative ability of the model, highlighting the necessity of explicit temporal modeling in realistic smart home ADL settings.

\subsubsection{Attention}
To assess the impact of the attention mechanism on classification performance, we evaluate Marauder’s model under both test and cross-activity test settings, comparing results with and without attention. As shown in Table \ref{tab:attn_perf}, the attention-enhanced model consistently outperforms the baseline across all metrics. Specifically, under the standard test setting, attention improves accuracy from 91.31\% to 92.74\% and F1-score from 85.68 to 86.27. More notably, under the more challenging cross-activity setting, attention leads to a substantial gain in both accuracy (from 86.08\% to 88.43\%) and F1-score (from 82.40 to 83.85), indicating enhanced robustness against mixed or ambiguous activity windows.
\begin{table}[h!]
\centering
\caption{Performance Comparison of Marauder's Model With and Without Attention (\%)}
\resizebox{1\columnwidth}{!}{
    \setlength{\tabcolsep}{1mm}{
\begin{tabular}{l|cccc|cccc}
\toprule
\multicolumn{1}{c|}{\textbf{Model}} & \multicolumn{4}{c|}{\textbf{Test Setting}} & \multicolumn{4}{c}{\textbf{Cross-Activity Test Setting}} \\
\cmidrule(lr){2-5} \cmidrule(lr){6-9}
& Accuracy & F1 & Precision & Recall & Accuracy & F1 & Precision & Recall \\
\midrule
Marauder's (w/o Attn) &
91.31 $\pm$ 0.54 & 85.68 $\pm$ 1.02 & 86.14 $\pm$ 1.53 & 85.74 $\pm$ 0.84 &
86.08 $\pm$ 0.62 & 82.40 $\pm$ 1.19 & 83.30 $\pm$ 1.82 & 82.30 $\pm$ 0.99 \\
Marauder's (with Attn) &
\textbf{92.74 $\pm$ 0.90} & \textbf{86.27 $\pm$ 2.51} & \textbf{87.04 $\pm$ 1.34} & \textbf{86.08 $\pm$ 3.36} &
\textbf{88.43 $\pm$ 0.93} & \textbf{83.85 $\pm$ 2.67} & \textbf{84.65 $\pm$ 1.67} & \textbf{83.78 $\pm$ 3.46} \\
\bottomrule
\end{tabular}}}
\label{tab:attn_perf}
\end{table}

\begin{figure}[h!]
    \centering
    \begin{tabular}{cc}
        \begin{subfigure}[t]{0.40\linewidth}
            \centering
            \includegraphics[width=\linewidth]{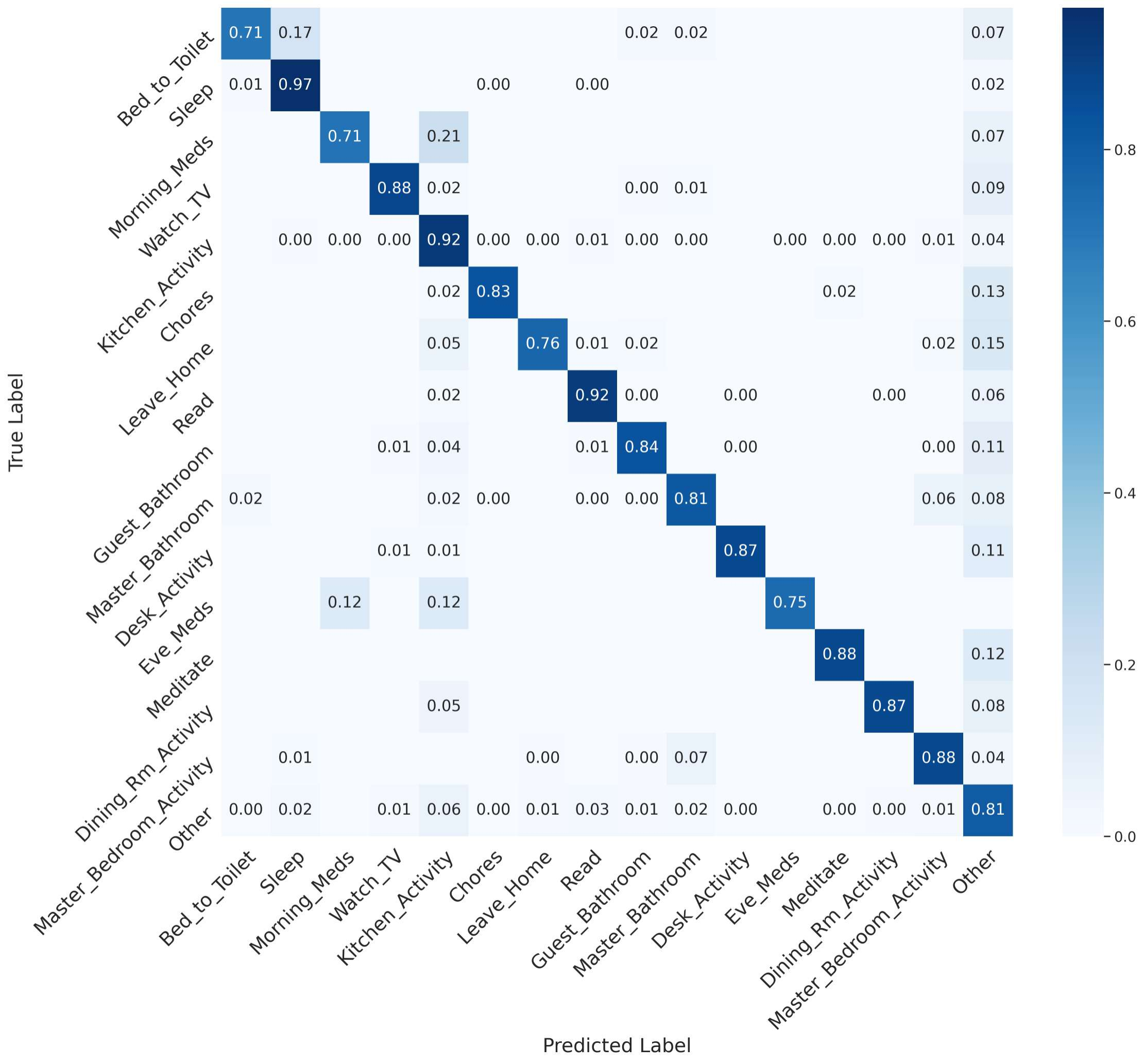}
            \caption{w/o Attn}
            \label{fig:woatt_cm}
        \end{subfigure}
        \hfill
        \begin{subfigure}[t]{0.40\linewidth}
            \centering
            \includegraphics[width=\linewidth]{figures/cm_cross_attn_normalized.pdf}
            \caption{Attn}
            \label{fig:attn_cm}
        \end{subfigure} 
    \end{tabular}
    \caption{Confusion matrices of classification results on cross-activity test data, comparing models without (a) and with (b) the attention mechanism. The attention-based model shows improved diagonal dominance, indicating enhanced accuracy and reduced misclassification across most activity classes.}
    \label{fig:com_attn_cm}
\end{figure}

\begin{figure}[h!]
    \centering
    \begin{tabular}{cc}
        \begin{subfigure}[t]{0.4\linewidth}
            \centering
            \includegraphics[width=\linewidth]{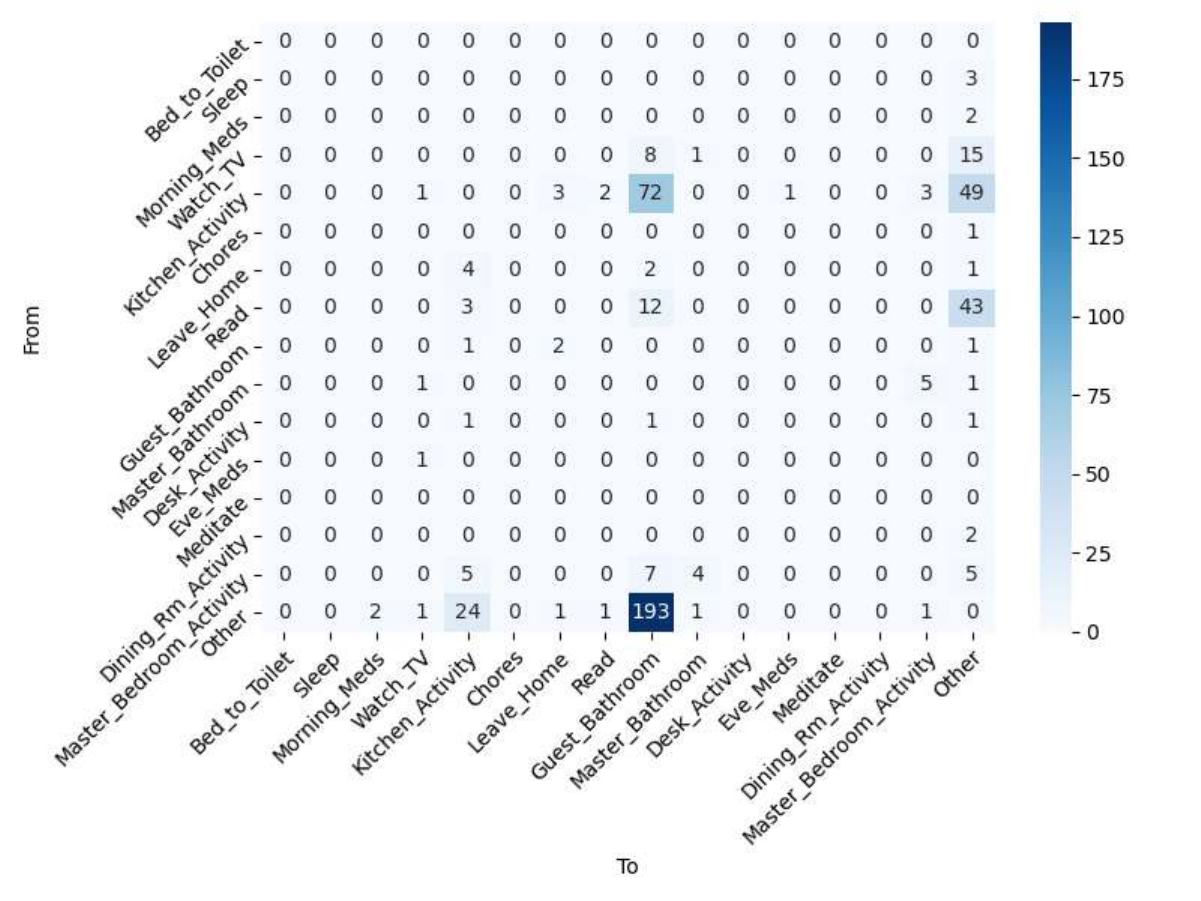}
            \caption{Guest Bathroom Activity}
            \label{fig:guestbath_trans}
        \end{subfigure}
        \hfill
        \begin{subfigure}[t]{0.4\linewidth}
            \centering
            \includegraphics[width=\linewidth]{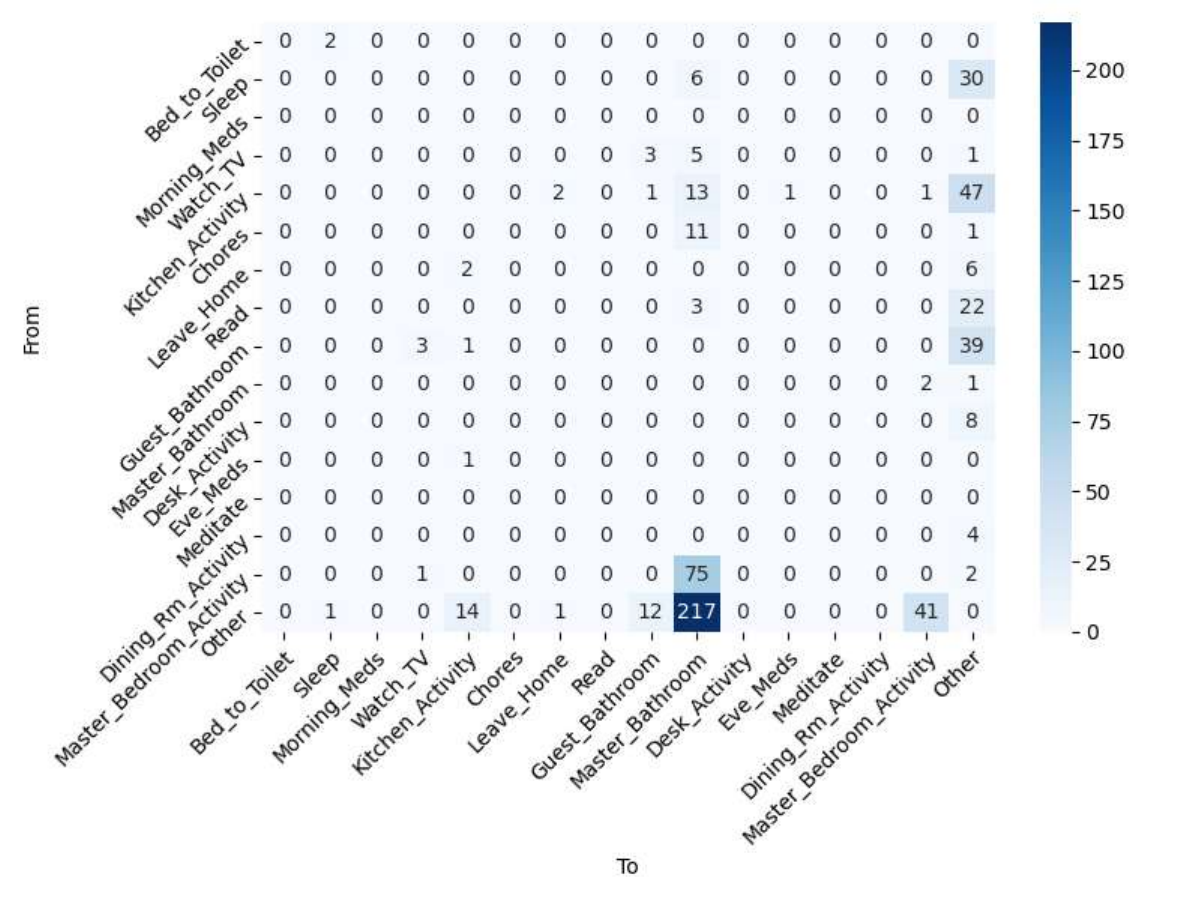}
            \caption{Master Bathroom Activity}
            \label{fig:masterbath_trans}
        \end{subfigure} 
    \end{tabular}
    \caption{Transition matrices for Guest Bathroom (a) and Master Bathroom (b) activities. The matrices highlight the frequent transitions between these activities and related activities, illustrating the complex temporal dependencies that the attention mechanism helps to disambiguate.}
    \label{fig:transition}
\end{figure}

The confusion matrices in Fig. \ref{fig:com_attn_cm} further illustrate these improvements. With attention, the model exhibits stronger diagonal dominance and reduced misclassification rates, particularly for activities that frequently co-occur with others in the window as shown in Fig.~\ref{fig:lable_unique}, such as Guest Bathroom and Master Bathroom. Without attention, these activities are often confused with adjacent contexts, whereas the attention mechanism enables the model to better distinguish fine-grained activity patterns. Fig. \ref{fig:transition} presents the transition matrices for Guest Bathroom and Master Bathroom, showing frequent transitions that could cause ambiguity. Overall, the quantitative and qualitative results confirm that attention significantly enhances the discriminative capacity and generalization ability of our model, especially in complex real-world smart home scenarios.

\subsection{Sliding Window Study}
\begin{figure}
    \centering
    \includegraphics[width=\linewidth]{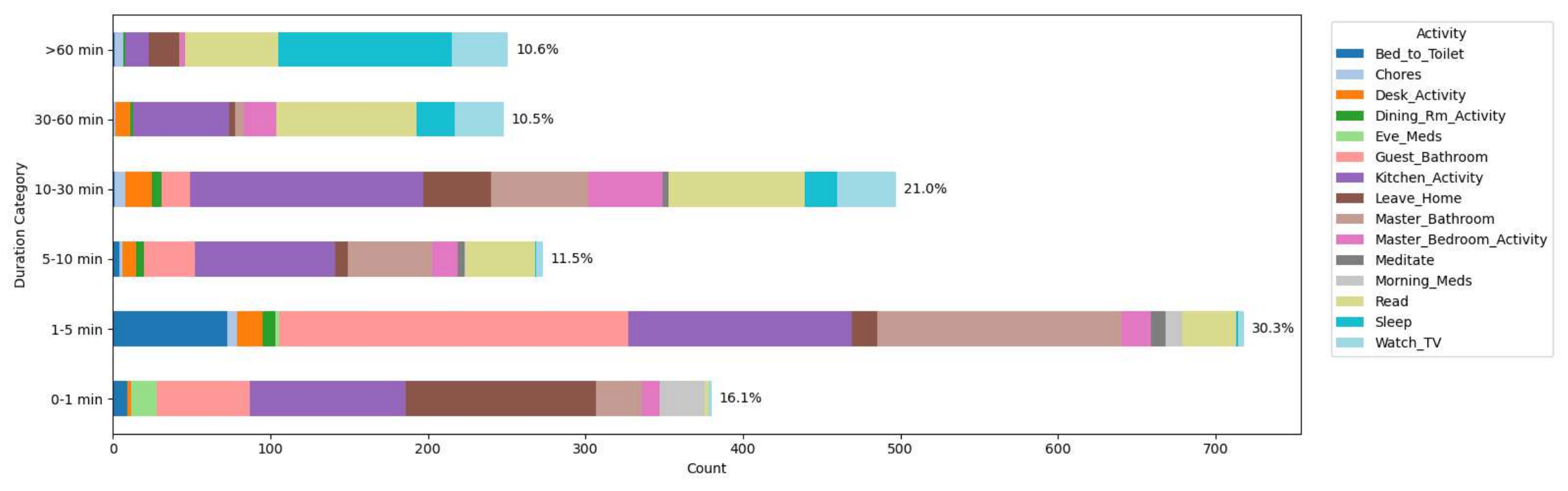}
    \caption{Activity duration distribution categorized into six time intervals on the dataset. Each bar represents the proportion of different activities within a specific duration range. The majority of activities fall within the 1–5 minute interval, while fewer instances are observed in durations exceeding 30 minutes. The color segments indicate the contribution of each activity type, revealing the dominance of short-term activities such as Bed\_to\_Toilet, Kitchen\_Activity, and Leave\_Home, and the concentration of longer-term activities such as Sleep and Watch\_TV in extended time categories.}
    \label{fig:overall_duration_distr}
\end{figure}

\begin{figure}[h!]
  \centering
   \includegraphics[width=0.95\linewidth]{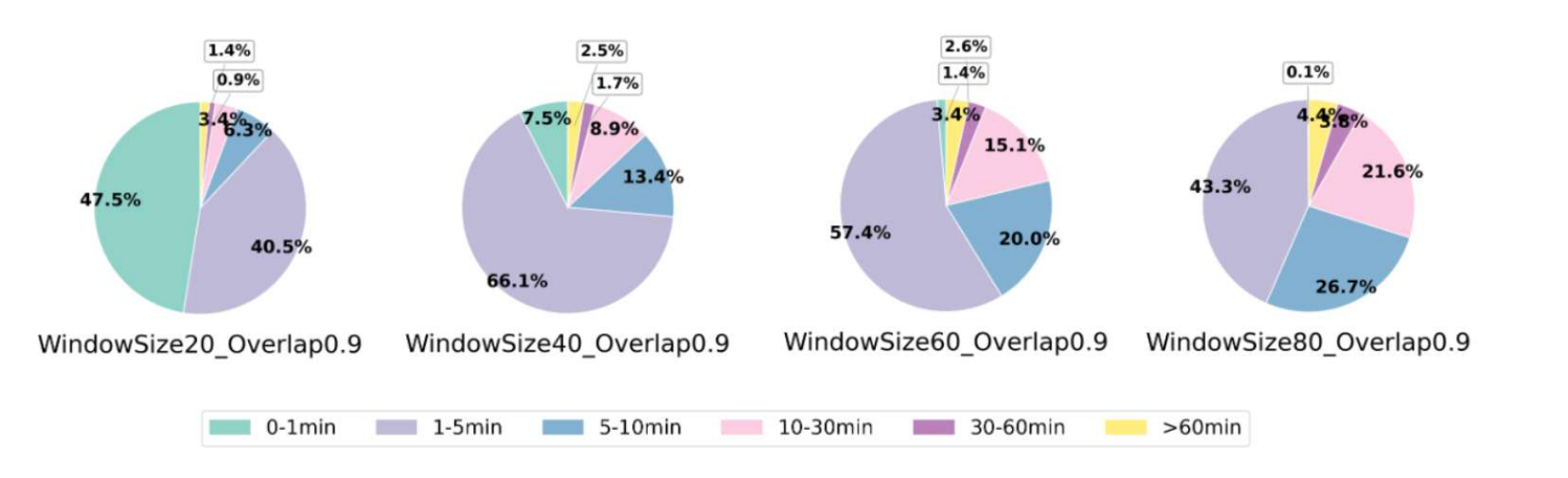}
   \caption{Window duration distributions under window sizes 20, 40 60, 80. The pie charts show how the window durations are categorized across six duration ranges (0–1 min, 1–5 min, 5–10 min, 10–30 min, 30–60 min, >60 min). Smaller window sizes result in shorter window durations, while larger window sizes capture longer activities but may increase cross-activity risk.}
   \label{fig:window_duration_distribution}
\end{figure}

We conduct a series of experiments varying window size and step size parameters used during the construction of input sequences to explore the influence of temporal segmentation on model performance. The window size determines the temporal span captured in each segment, while the step size controls the degree of overlap between consecutive windows. To better guide our window size selection, we analyze the overall activity duration distribution alongside the window duration distributions across different configurations. In Fig.\ref{fig:overall_duration_distr}, the overall statistics show that the majority of activity segments are relatively short, with approximately 30\% lasting 1–5 minutes and around 16\% lasting less than 1 minute, while longer activities (10–30 minutes and beyond) account for a smaller but non-negligible portion. Consistent with this, we can see from Fig. \ref{fig:window_duration_distribution} that when using a small window size (e.g., 20), the generated windows are predominantly short, with nearly 88\% falling within the 0–5 minute range. As the window size increases to 40 or 60, the window durations shift, capturing a higher proportion of medium-length activities (5–10 minutes and 10–30 minutes), aligning better with the broader range of activities present in the dataset. With the largest window size (80), an even larger fraction of windows fall into the 5–30 minute range, though the risk of crossing multiple activities increases.

\begin{figure}[h!]
    \centering
    \includegraphics[width=0.40\linewidth]{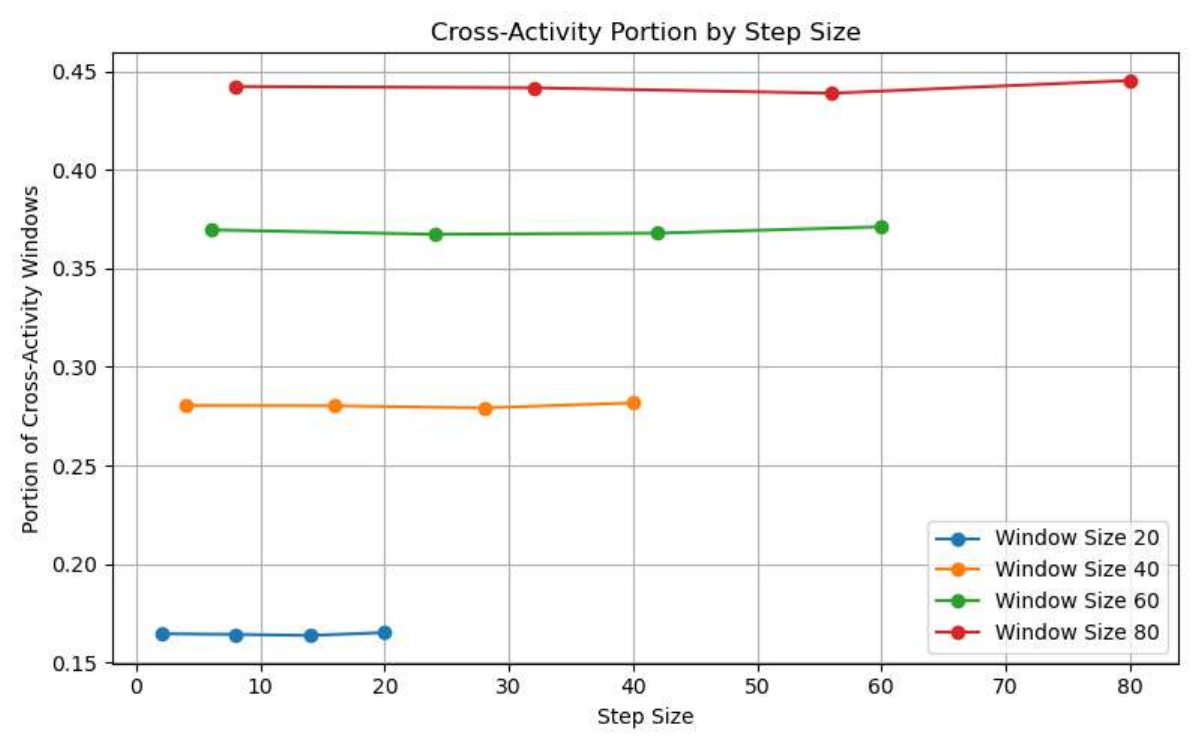}
    \caption{Portion of cross-activity windows under different window sizes and step sizes. The figure shows that larger window sizes consistently lead to higher proportions of cross-activity windows. Step size (overlapping) has limited effect on the proportion, indicating that window size is the primary factor influencing cross-activity contamination.}
    \label{fig:cross-portion}
\end{figure}

This observation is supported by the analysis of cross-activity window portions, where larger window sizes correspond to higher proportions of windows containing transitions between multiple activities. Specifically, as shown in Fig. \ref{fig:cross-portion}, the proportion of cross-activity windows steadily increases with window size: window size 20 exhibits the lowest cross-activity portion (17\%), while window size 80 reaches around 44\%. Additionally, the step size (i.e., the degree of overlap) has minimal influence on the cross-activity portion within each window size, further suggesting that overlap primarily affects sample augmentation rather than the purity of windows. Based on these findings, we conclude that moderate window sizes (e.g., 40 or 60) strike a better balance between capturing meaningful activity durations and minimizing cross-activity contamination, which is crucial for maintaining classification performance and robustness to label noise.

\begin{figure}[h!]
    \centering
    \begin{tabular}{cccc}
        \begin{subfigure}{0.25\linewidth}
            \centering
            \includegraphics[width=\linewidth]{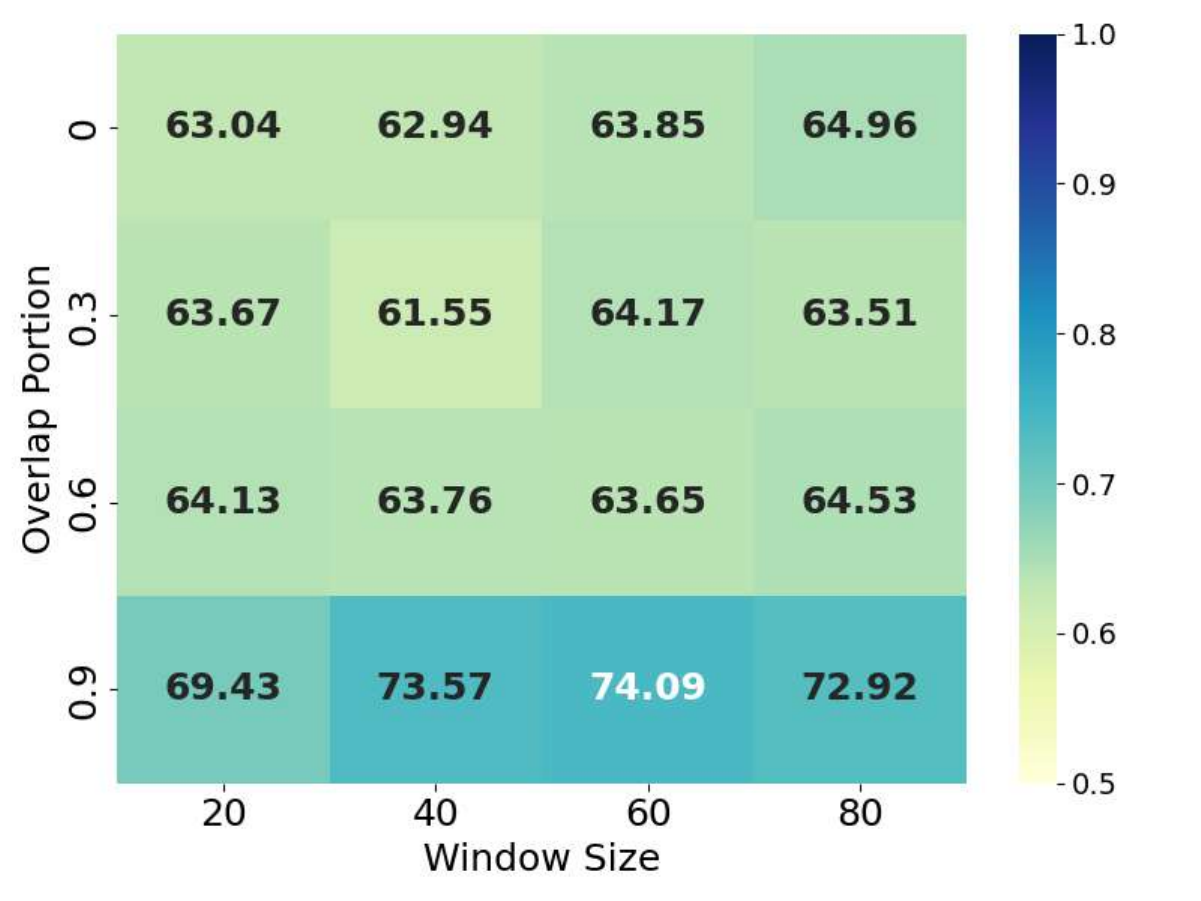}
            \caption{DeepCasas}
        \end{subfigure}
        \hfill
        \begin{subfigure}{0.25\linewidth}
            \centering
            \includegraphics[width=\linewidth]{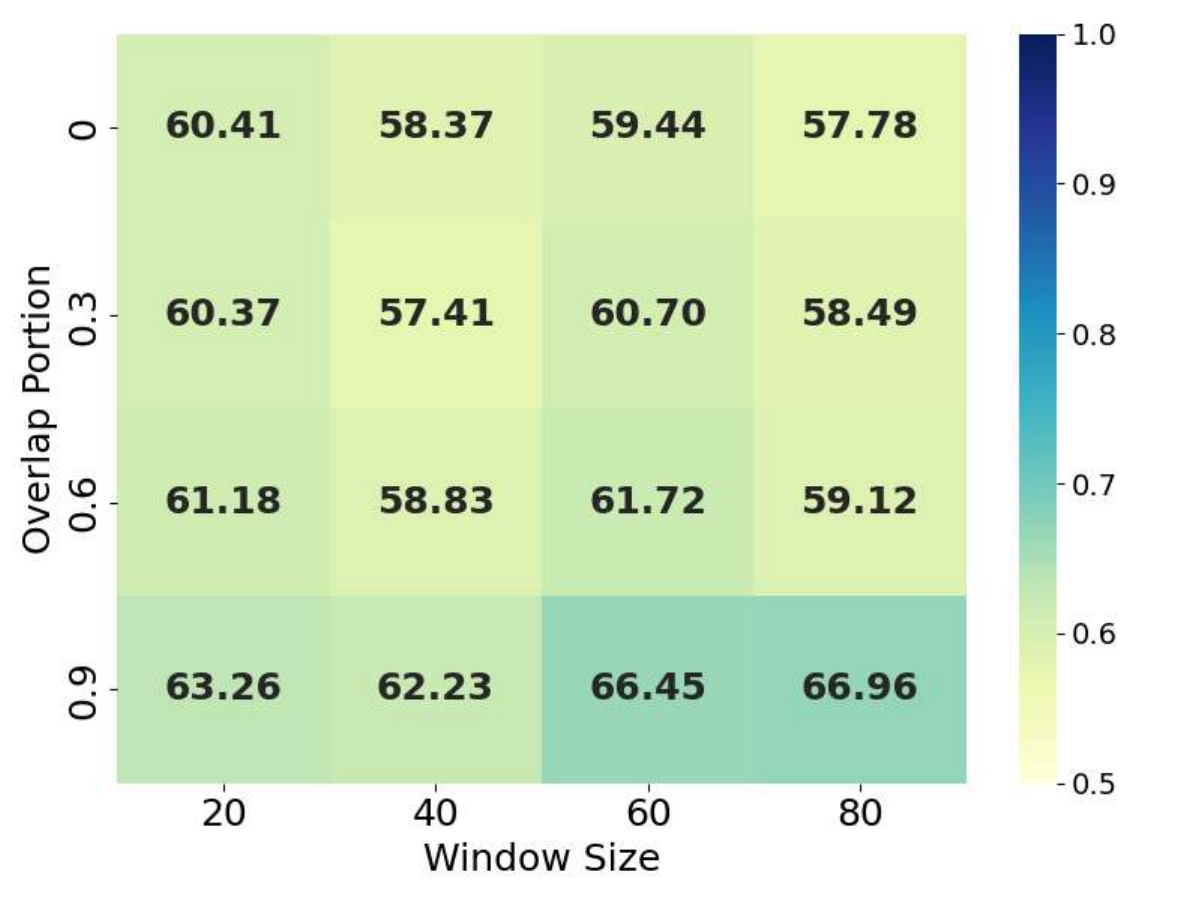}
            \caption{DCNN}
        \end{subfigure}
        \hfill
        \begin{subfigure}{0.25\linewidth}
            \centering
            \includegraphics[width=\linewidth]{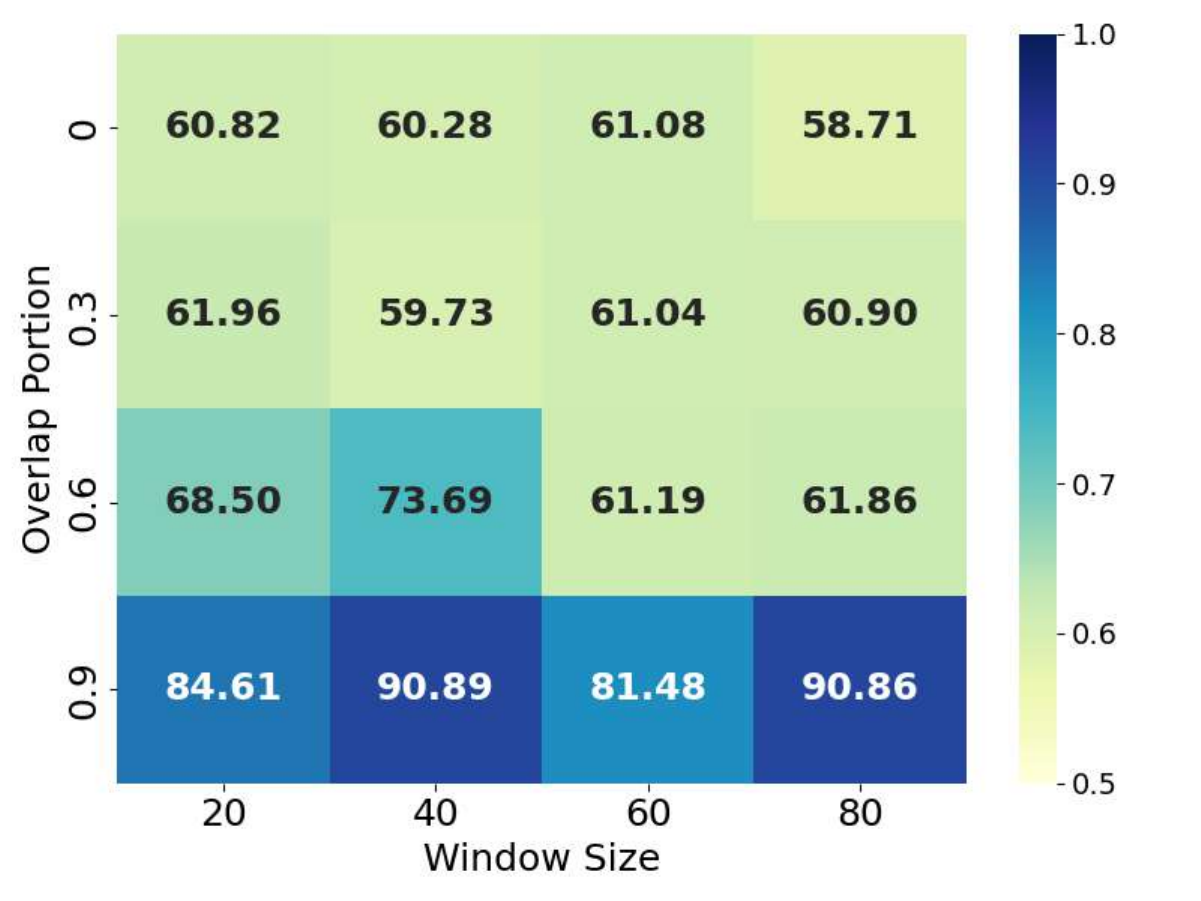}
            \caption{TCN}
        \end{subfigure} 
        \hfill
        \begin{subfigure}{0.25\linewidth}
            \centering
            \includegraphics[width=\linewidth]{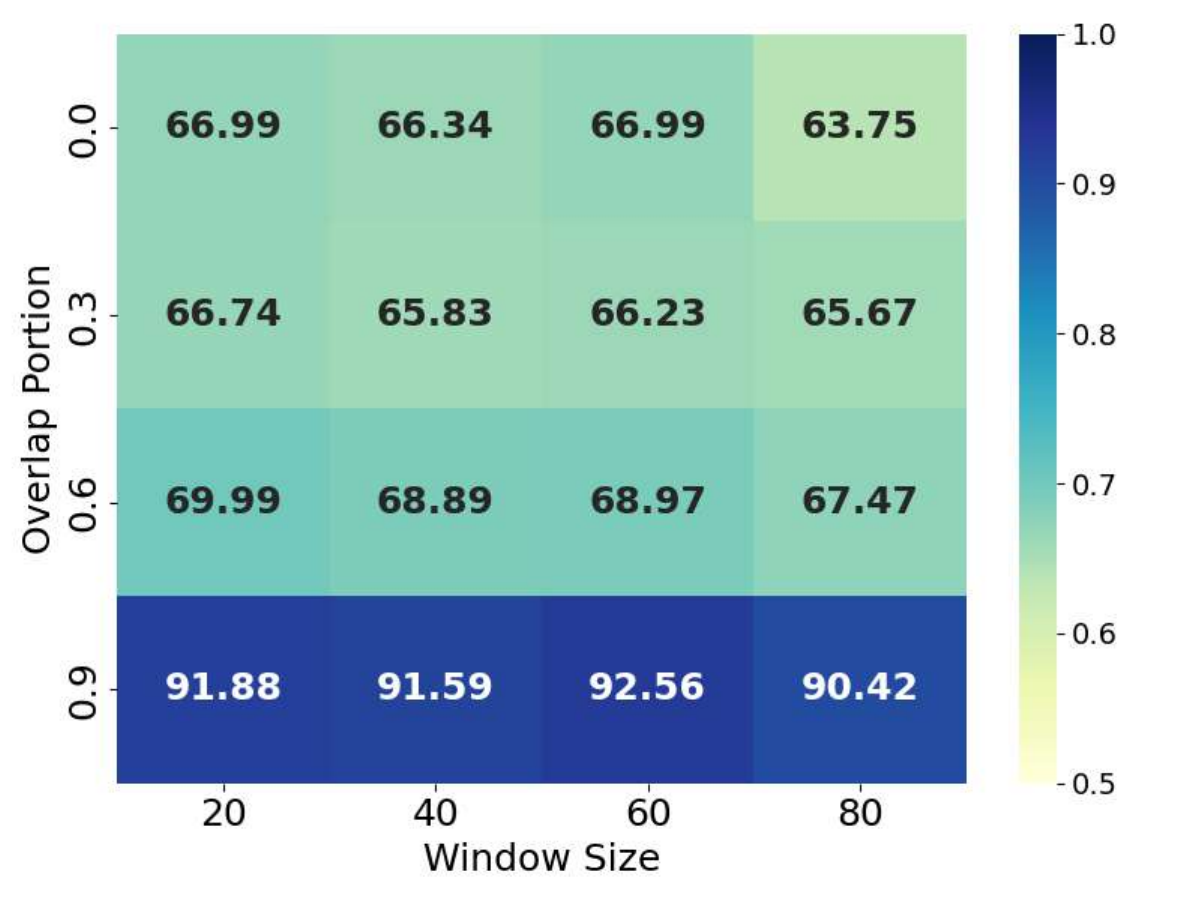}
            \caption{Marauder's Map}
        \end{subfigure} \\
        \begin{subfigure}{0.25\linewidth}
            \centering
            \includegraphics[width=\linewidth]{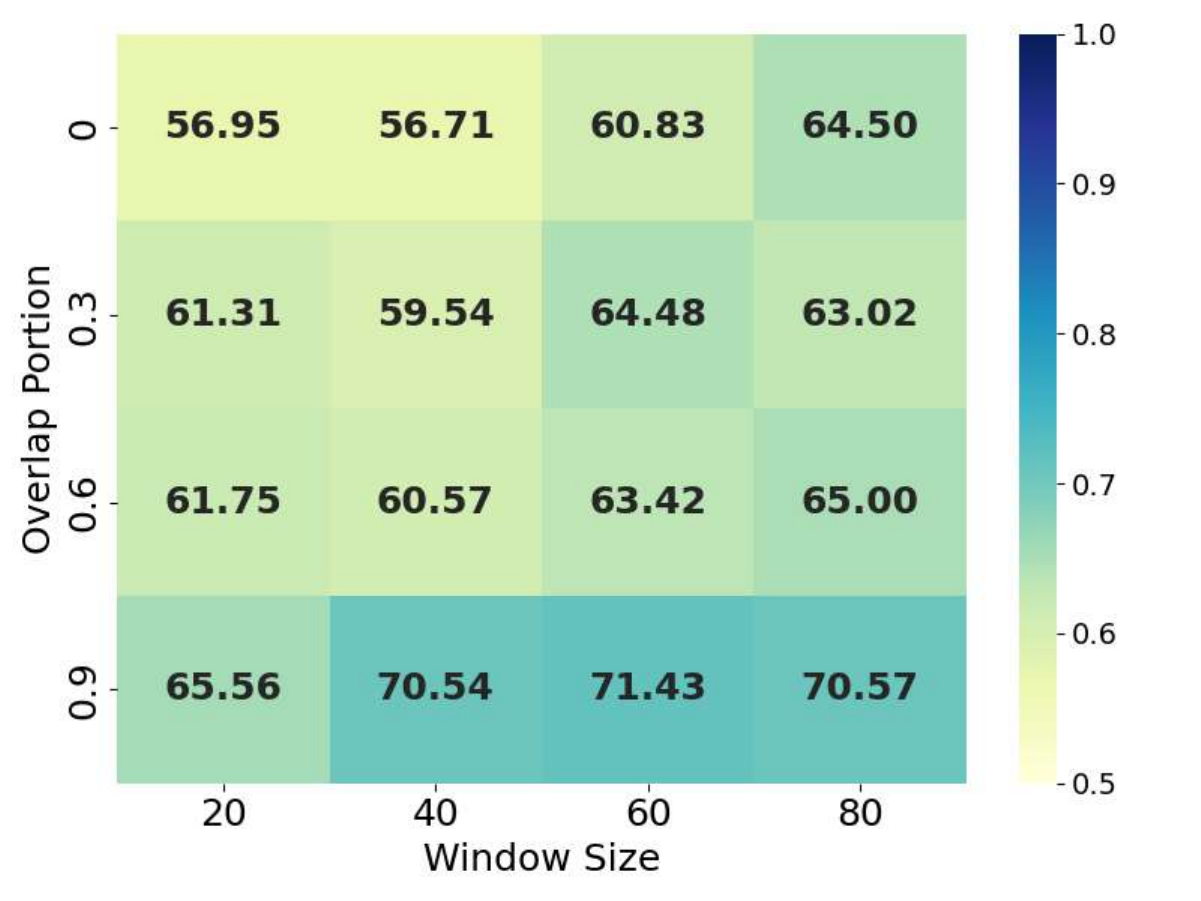}
            \caption{DeepCasas}
        \end{subfigure}
        \hfill
        \begin{subfigure}{0.25\linewidth}
            \centering
            \includegraphics[width=\linewidth]{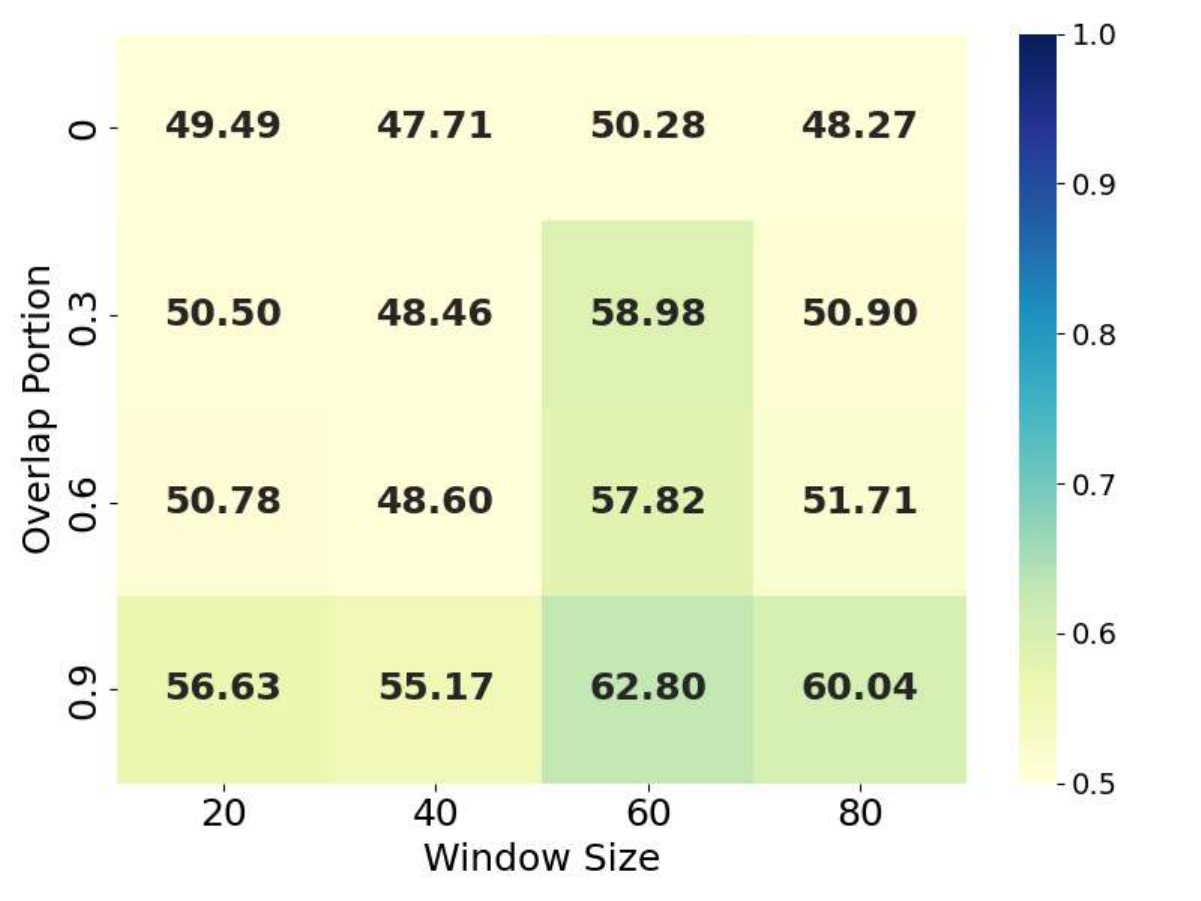}
            \caption{DCNN}
        \end{subfigure}
        \hfill
        \begin{subfigure}{0.25\linewidth}
            \centering
            \includegraphics[width=\linewidth]{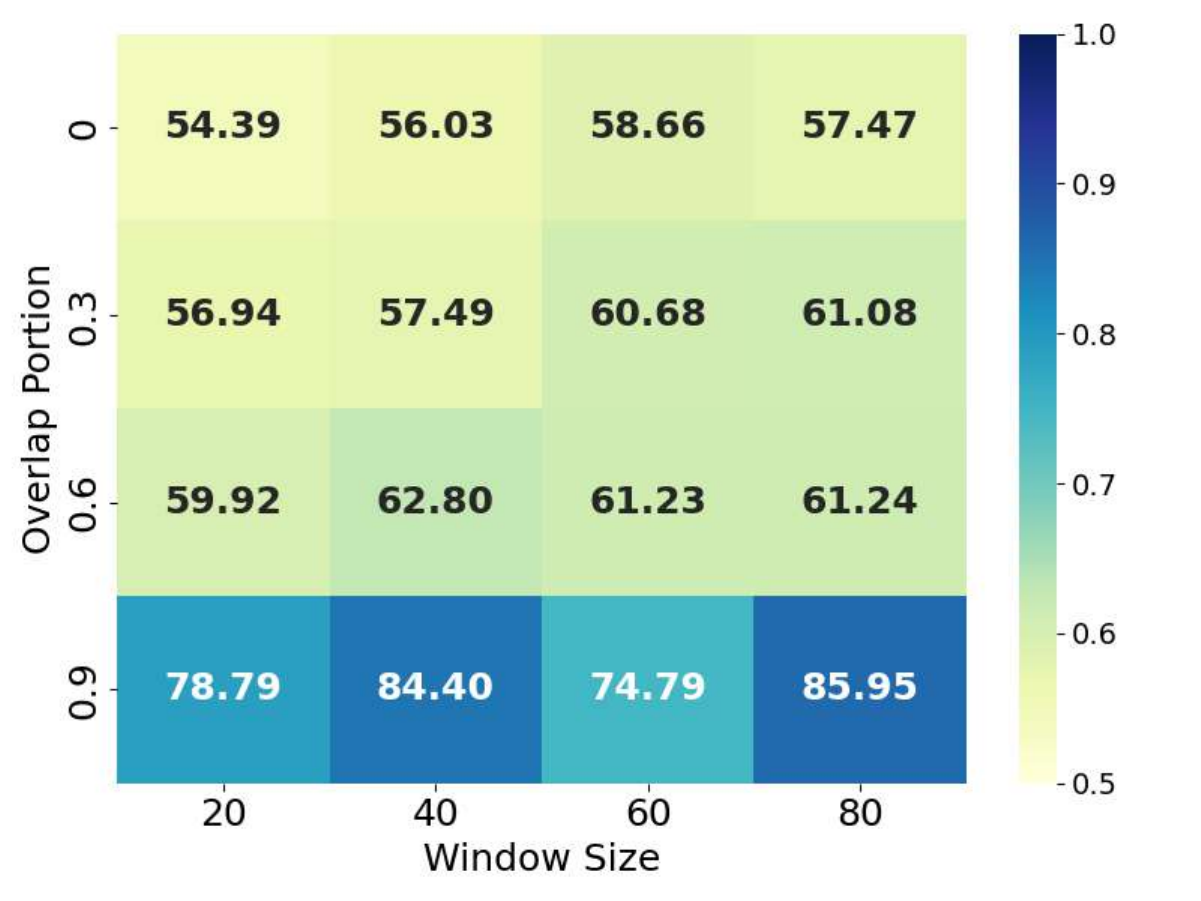}
            \caption{TCN}
        \end{subfigure} 
        \hfill
        \begin{subfigure}{0.25\linewidth}
            \centering
            \includegraphics[width=\linewidth]{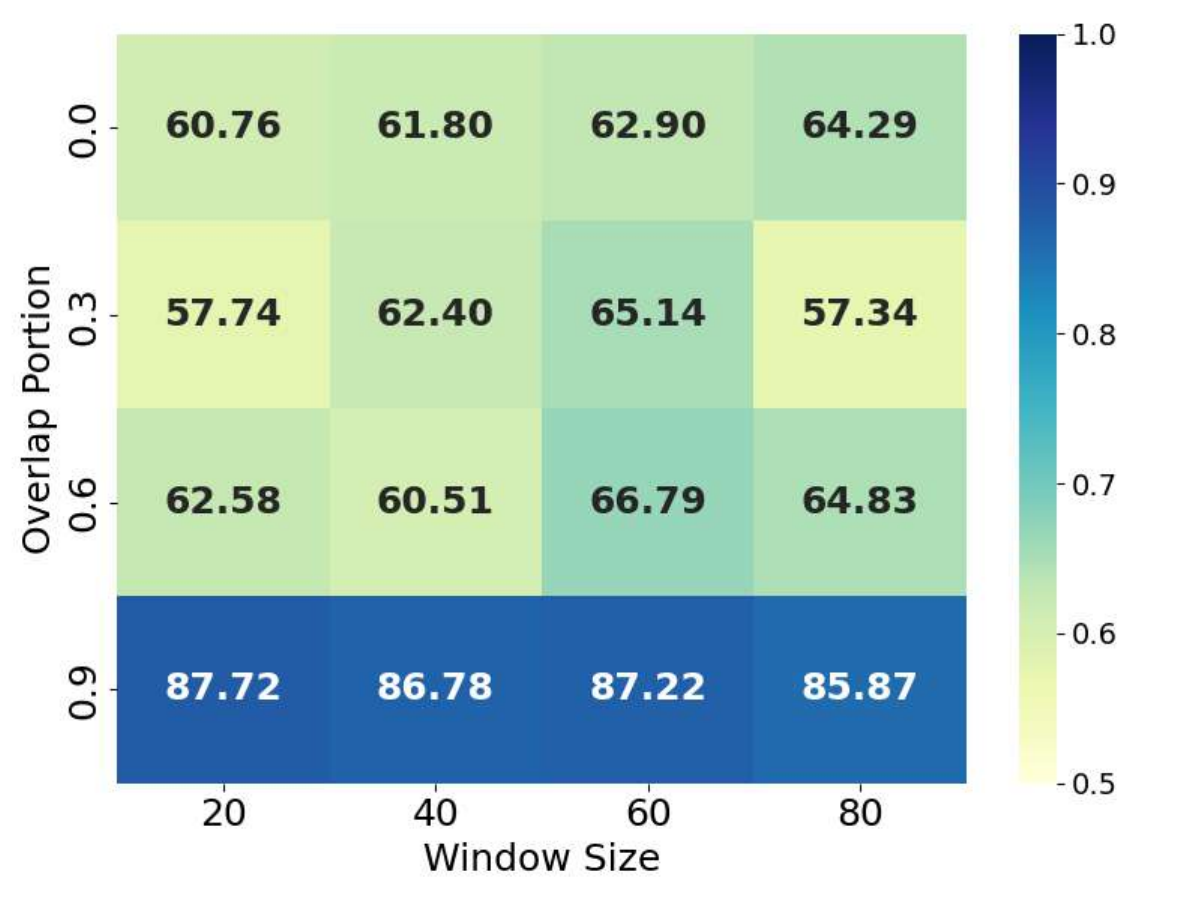}
            \caption{Marauder's Map}
        \end{subfigure}
    \end{tabular}
    \caption{Performance comparison of different models on the Milan dataset across varying window sizes and overlap portions. The first row (a–d) shows the accuracy on the original test set, while the second row (e–h) shows the performance on cross-activity windows. Models compared include DeepCasas (a, e), DCNN (b, f), TCN (c, g), and our proposed Marauder’s Map (d, h). Each heatmap cell represents classification accuracy for a specific combination of window size and overlap ratio.}
    \label{fig:win-step}
\end{figure}

To further evaluate model robustness across different temporal configurations, we present the accuracy of DeepCasas, DCNN, TCN, and Marauder’s Map on both the original test set and the cross-activity windows (Fig. \ref{fig:win-step}). Consistent with the prior analysis of window durations and cross-activity proportions, all models tend to achieve their highest performance with larger overlap ratios (i.e., 0.9), benefiting from denser sampling of the activity sequences. However, significant differences are observed across models. On the original test set (first row), Marauder’s Map consistently outperforms all baselines in most configurations, especially at window sizes 40 and 60 with high overlap, where it achieves peak accuracies above 92\%. TCN also performs competitively, though slightly lower than Marauder’s Map, while DeepCasas and DCNN exhibit noticeably lower accuracies overall.DeepCasas and DCNN consistently show lower accuracies across all configurations, with particularly poor performance at low overlap levels. These results highlight Marauder’s Map’s general superiority in modeling temporal dependencies, while also acknowledging that TCN remains competitive in specific settings.

The trend becomes more pronounced in the cross-activity setting (second row), where handling overlapping and ambiguous activity windows becomes more challenging. Baseline models suffer a clear drop in accuracy, especially DCNN, which struggles under cross-activity contamination. In contrast, Marauder’s Map maintains robust performance across most configurations. These results validate that our method's time-aware and attention-guided design effectively mitigates the effects of label noise and temporal ambiguity introduced by cross-activity windows. Furthermore, the performance trends corroborate our earlier findings: moderate window sizes (40 or 60) combined with high overlap yield the best balance between capturing activity context and minimizing cross-activity confusion.

\section{Conclusion}
In this paper, we introduce MARAuder’s Map, a novel framework for real-time activity of daily living recognition in smart home environments. Our approach addresses key limitations of traditional winthin-activity recognition by embracing the spatial-temporal complexity of real-world sensor data, where observation windows often contain multiple or transitional behaviors. By transforming raw sensor sequences into spatially grounded trajectory images mapped onto the physical home layout, our method preserves the contextual relationships among triggered sensors. A learnable time embedding module further captures time-sensitive cues such as hour of day and day of week, enabling the model to differentiate between temporally distinct routines. Additionally, our attention-based sequential architecture enables the model to selectively focus on informative temporal regions, improving robustness to transitional or composite activity windows that often arise in real-time settings. Through comprehensive studies, we demonstrated the effectiveness of our method on real-word datasets.

In general, MARAuder's Map provides a unified and extensible solution for trajectory-driven activity recognition, with practical implications for smart home monitoring, assistive living, and ubiquitous computing. Future work will explore online adaptation for personalization, multi-resident activity disambiguation, and cross-home generalization to enhance scalability and robustness in diverse deployment scenarios.

\begin{thebibliography}{77}


\ifx \showCODEN    \undefined \def \showCODEN     #1{\unskip}     \fi
\ifx \showISBNx    \undefined \def \showISBNx     #1{\unskip}     \fi
\ifx \showISBNxiii \undefined \def \showISBNxiii  #1{\unskip}     \fi
\ifx \showISSN     \undefined \def \showISSN      #1{\unskip}     \fi
\ifx \showLCCN     \undefined \def \showLCCN      #1{\unskip}     \fi
\ifx \shownote     \undefined \def \shownote      #1{#1}          \fi
\ifx \showarticletitle \undefined \def \showarticletitle #1{#1}   \fi
\ifx \showURL      \undefined \def \showURL       {\relax}        \fi
\providecommand\bibfield[2]{#2}
\providecommand\bibinfo[2]{#2}
\providecommand\natexlab[1]{#1}
\providecommand\showeprint[2][]{arXiv:#2}

\bibitem[Al~Machot and Mayr(2016)]%
        {al2016improving}
\bibfield{author}{\bibinfo{person}{Fadi Al~Machot} {and} \bibinfo{person}{Heinrich~C Mayr}.} \bibinfo{year}{2016}\natexlab{}.
\newblock \showarticletitle{Improving human activity recognition by smart windowing and spatio-temporal feature analysis}. In \bibinfo{booktitle}{\emph{Proceedings of the 9th ACM International Conference on PErvasive Technologies Related to Assistive Environments}}. \bibinfo{pages}{1--7}.
\newblock


\bibitem[Al~Machot et~al\mbox{.}(2016)]%
        {al2016windowing}
\bibfield{author}{\bibinfo{person}{Fadi Al~Machot}, \bibinfo{person}{Heinrich~C Mayr}, {and} \bibinfo{person}{Suneth Ranasinghe}.} \bibinfo{year}{2016}\natexlab{}.
\newblock \showarticletitle{A windowing approach for activity recognition in sensor data streams}. In \bibinfo{booktitle}{\emph{2016 Eighth International Conference on Ubiquitous and Future Networks (ICUFN)}}. IEEE, \bibinfo{pages}{951--953}.
\newblock


\bibitem[Al~Machot et~al\mbox{.}(2017)]%
        {al2017activity}
\bibfield{author}{\bibinfo{person}{Fadi Al~Machot}, \bibinfo{person}{Ahmad~Haj Mosa}, \bibinfo{person}{Mouhannad Ali}, {and} \bibinfo{person}{Kyandoghere Kyamakya}.} \bibinfo{year}{2017}\natexlab{}.
\newblock \showarticletitle{Activity recognition in sensor data streams for active and assisted living environments}.
\newblock \bibinfo{journal}{\emph{IEEE Transactions on Circuits and Systems for Video Technology}} \bibinfo{volume}{28}, \bibinfo{number}{10} (\bibinfo{year}{2017}), \bibinfo{pages}{2933--2945}.
\newblock


\bibitem[Ali et~al\mbox{.}(2022)]%
        {ali2022human}
\bibfield{author}{\bibinfo{person}{Ashraf Ali}, \bibinfo{person}{Weam Samara}, \bibinfo{person}{Doaa Alhaddad}, \bibinfo{person}{Andrew Ware}, {and} \bibinfo{person}{Omar~A Saraereh}.} \bibinfo{year}{2022}\natexlab{}.
\newblock \showarticletitle{Human activity and motion pattern recognition within indoor environment using convolutional neural networks clustering and naive bayes classification algorithms}.
\newblock \bibinfo{journal}{\emph{Sensors}} \bibinfo{volume}{22}, \bibinfo{number}{3} (\bibinfo{year}{2022}), \bibinfo{pages}{1016}.
\newblock


\bibitem[Alshammari et~al\mbox{.}(2018)]%
        {alshammari2018evaluating}
\bibfield{author}{\bibinfo{person}{Talal Alshammari}, \bibinfo{person}{Nasser Alshammari}, \bibinfo{person}{Mohamed Sedky}, {and} \bibinfo{person}{Chris Howard}.} \bibinfo{year}{2018}\natexlab{}.
\newblock \showarticletitle{Evaluating machine learning techniques for activity classification in smart home environments}.
\newblock \bibinfo{journal}{\emph{Int. J. Inf. Commun. Eng}}  \bibinfo{volume}{12} (\bibinfo{year}{2018}), \bibinfo{pages}{72--78}.
\newblock


\bibitem[Arrotta et~al\mbox{.}(2022)]%
        {arrotta2022dexar}
\bibfield{author}{\bibinfo{person}{Luca Arrotta}, \bibinfo{person}{Gabriele Civitarese}, {and} \bibinfo{person}{Claudio Bettini}.} \bibinfo{year}{2022}\natexlab{}.
\newblock \showarticletitle{Dexar: Deep explainable sensor-based activity recognition in smart-home environments}.
\newblock \bibinfo{journal}{\emph{Proceedings of the ACM on Interactive, Mobile, Wearable and Ubiquitous Technologies}} \bibinfo{volume}{6}, \bibinfo{number}{1} (\bibinfo{year}{2022}), \bibinfo{pages}{1--30}.
\newblock


\bibitem[Asghari et~al\mbox{.}(2020)]%
        {asghari2020online}
\bibfield{author}{\bibinfo{person}{Parviz Asghari}, \bibinfo{person}{Elnaz Soleimani}, {and} \bibinfo{person}{Ehsan Nazerfard}.} \bibinfo{year}{2020}\natexlab{}.
\newblock \showarticletitle{Online human activity recognition employing hierarchical hidden Markov models}.
\newblock \bibinfo{journal}{\emph{Journal of Ambient Intelligence and Humanized Computing}}  \bibinfo{volume}{11} (\bibinfo{year}{2020}), \bibinfo{pages}{1141--1152}.
\newblock


\bibitem[Bai et~al\mbox{.}(2018)]%
        {bai2018empirical}
\bibfield{author}{\bibinfo{person}{Shaojie Bai}, \bibinfo{person}{J~Zico Kolter}, {and} \bibinfo{person}{Vladlen Koltun}.} \bibinfo{year}{2018}\natexlab{}.
\newblock \showarticletitle{An empirical evaluation of generic convolutional and recurrent networks for sequence modeling}.
\newblock \bibinfo{journal}{\emph{arXiv preprint arXiv:1803.01271}} (\bibinfo{year}{2018}).
\newblock


\bibitem[Banos et~al\mbox{.}(2014)]%
        {banos2014window}
\bibfield{author}{\bibinfo{person}{Oresti Banos}, \bibinfo{person}{Juan-Manuel Galvez}, \bibinfo{person}{Miguel Damas}, \bibinfo{person}{Hector Pomares}, {and} \bibinfo{person}{Ignacio Rojas}.} \bibinfo{year}{2014}\natexlab{}.
\newblock \showarticletitle{Window size impact in human activity recognition}.
\newblock \bibinfo{journal}{\emph{Sensors}} \bibinfo{volume}{14}, \bibinfo{number}{4} (\bibinfo{year}{2014}), \bibinfo{pages}{6474--6499}.
\newblock


\bibitem[Bari et~al\mbox{.}(2024)]%
        {bari2024advancements}
\bibfield{author}{\bibinfo{person}{Ahsanul Bari}, \bibinfo{person}{Fahmid Al~Farid}, \bibinfo{person}{Md~Tanjil Sarker}, \bibinfo{person}{Sarina Mansor}, \bibinfo{person}{Hezerul~Abdul Karim}, \bibinfo{person}{Md~Roman Bhuiyan}, {and} \bibinfo{person}{Hasanul Bannah}.} \bibinfo{year}{2024}\natexlab{}.
\newblock \showarticletitle{Advancements in Multi-View Human Activity Recognition for Ambient Assisted Living}. In \bibinfo{booktitle}{\emph{2024 Multimedia University Engineering Conference (MECON)}}. IEEE, \bibinfo{pages}{1--6}.
\newblock


\bibitem[Bhattacharya et~al\mbox{.}(2022)]%
        {bhattacharya2022ensem}
\bibfield{author}{\bibinfo{person}{Debarshi Bhattacharya}, \bibinfo{person}{Deepak Sharma}, \bibinfo{person}{Wonjoon Kim}, \bibinfo{person}{Muhammad~Fazal Ijaz}, {and} \bibinfo{person}{Pawan~Kumar Singh}.} \bibinfo{year}{2022}\natexlab{}.
\newblock \showarticletitle{Ensem-HAR: An ensemble deep learning model for smartphone sensor-based human activity recognition for measurement of elderly health monitoring}.
\newblock \bibinfo{journal}{\emph{Biosensors}} \bibinfo{volume}{12}, \bibinfo{number}{6} (\bibinfo{year}{2022}), \bibinfo{pages}{393}.
\newblock


\bibitem[Bibb{\`o} et~al\mbox{.}(2025)]%
        {bibbo2025mems}
\bibfield{author}{\bibinfo{person}{Luigi Bibb{\`o}}, \bibinfo{person}{Giovanni Angiulli}, \bibinfo{person}{Filippo Lagan{\`a}}, \bibinfo{person}{Danilo Prattic{\`o}}, \bibinfo{person}{Francesco Cotroneo}, \bibinfo{person}{Fabio La~Foresta}, {and} \bibinfo{person}{Mario Versaci}.} \bibinfo{year}{2025}\natexlab{}.
\newblock \showarticletitle{MEMS and IoT in HAR: Effective Monitoring for the Health of Older People}.
\newblock \bibinfo{journal}{\emph{Applied Sciences}} \bibinfo{volume}{15}, \bibinfo{number}{8} (\bibinfo{year}{2025}), \bibinfo{pages}{4306}.
\newblock


\bibitem[Bibb{\`o} and Vellasco(2023)]%
        {bibbo2023human}
\bibfield{author}{\bibinfo{person}{Luigi Bibb{\`o}} {and} \bibinfo{person}{Marley~MBR Vellasco}.} \bibinfo{year}{2023}\natexlab{}.
\newblock \bibinfo{title}{Human activity recognition (har) in healthcare}.
\newblock \bibinfo{numpages}{13009}~pages.
\newblock


\bibitem[Bloom et~al\mbox{.}(2016)]%
        {bloom2016demography}
\bibfield{author}{\bibinfo{person}{David~E Bloom}, \bibinfo{person}{Elizabeth Mitgang}, {and} \bibinfo{person}{Benjamin Osher}.} \bibinfo{year}{2016}\natexlab{}.
\newblock \showarticletitle{Demography of global ageing}.
\newblock  (\bibinfo{year}{2016}).
\newblock


\bibitem[Bock et~al\mbox{.}(2021)]%
        {bock2021improving}
\bibfield{author}{\bibinfo{person}{Marius Bock}, \bibinfo{person}{Alexander H{\"o}lzemann}, \bibinfo{person}{Michael Moeller}, {and} \bibinfo{person}{Kristof Van~Laerhoven}.} \bibinfo{year}{2021}\natexlab{}.
\newblock \showarticletitle{Improving deep learning for HAR with shallow LSTMs}. In \bibinfo{booktitle}{\emph{Proceedings of the 2021 ACM International Symposium on Wearable Computers}}. \bibinfo{pages}{7--12}.
\newblock


\bibitem[Bouchabou et~al\mbox{.}(2021a)]%
        {bouchabou2021survey}
\bibfield{author}{\bibinfo{person}{Damien Bouchabou}, \bibinfo{person}{Sao~Mai Nguyen}, \bibinfo{person}{Christophe Lohr}, \bibinfo{person}{Benoit LeDuc}, {and} \bibinfo{person}{Ioannis Kanellos}.} \bibinfo{year}{2021}\natexlab{a}.
\newblock \showarticletitle{A survey of human activity recognition in smart homes based on IoT sensors algorithms: Taxonomies, challenges, and opportunities with deep learning}.
\newblock \bibinfo{journal}{\emph{Sensors}} \bibinfo{volume}{21}, \bibinfo{number}{18} (\bibinfo{year}{2021}), \bibinfo{pages}{6037}.
\newblock


\bibitem[Bouchabou et~al\mbox{.}(2021b)]%
        {bouchabou2021using}
\bibfield{author}{\bibinfo{person}{Damien Bouchabou}, \bibinfo{person}{Sao~Mai Nguyen}, \bibinfo{person}{Christophe Lohr}, \bibinfo{person}{Benoit LeDuc}, {and} \bibinfo{person}{Ioannis Kanellos}.} \bibinfo{year}{2021}\natexlab{b}.
\newblock \showarticletitle{Using language model to bootstrap human activity recognition ambient sensors based in smart homes}.
\newblock \bibinfo{journal}{\emph{Electronics}} \bibinfo{volume}{10}, \bibinfo{number}{20} (\bibinfo{year}{2021}), \bibinfo{pages}{2498}.
\newblock


\bibitem[Chan et~al\mbox{.}(2008)]%
        {chan2008review}
\bibfield{author}{\bibinfo{person}{Marie Chan}, \bibinfo{person}{Daniel Est{\`e}ve}, \bibinfo{person}{Christophe Escriba}, {and} \bibinfo{person}{Eric Campo}.} \bibinfo{year}{2008}\natexlab{}.
\newblock \showarticletitle{A review of smart homes—Present state and future challenges}.
\newblock \bibinfo{journal}{\emph{Computer methods and programs in biomedicine}} \bibinfo{volume}{91}, \bibinfo{number}{1} (\bibinfo{year}{2008}), \bibinfo{pages}{55--81}.
\newblock


\bibitem[ChatGPT(2025)]%
        {chatgpt_image2025}
\bibfield{author}{\bibinfo{person}{OpenAI ChatGPT}.} \bibinfo{year}{2025}\natexlab{}.
\newblock \bibinfo{title}{Custom Image Generated by ChatGPT for Smart Home Sensor Data Visualization}.
\newblock \bibinfo{howpublished}{\url{https://chat.openai.com/}}.
\newblock
\newblock
\shownote{Accessed via ChatGPT on April 21, 2025}.


\bibitem[Chen et~al\mbox{.}(2023)]%
        {chen2023digital}
\bibfield{author}{\bibinfo{person}{Junxin Chen}, \bibinfo{person}{Wei Wang}, \bibinfo{person}{Bo Fang}, \bibinfo{person}{Yu Liu}, \bibinfo{person}{Keping Yu}, \bibinfo{person}{Victor~CM Leung}, {and} \bibinfo{person}{Xiping Hu}.} \bibinfo{year}{2023}\natexlab{}.
\newblock \showarticletitle{Digital twin empowered wireless healthcare monitoring for smart home}.
\newblock \bibinfo{journal}{\emph{IEEE Journal on Selected Areas in Communications}} \bibinfo{volume}{41}, \bibinfo{number}{11} (\bibinfo{year}{2023}), \bibinfo{pages}{3662--3676}.
\newblock


\bibitem[Chen et~al\mbox{.}(2008)]%
        {chen2008logical}
\bibfield{author}{\bibinfo{person}{Luke Chen}, \bibinfo{person}{Chris~D Nugent}, \bibinfo{person}{Maurice Mulvenna}, \bibinfo{person}{Dewar Finlay}, \bibinfo{person}{Xin Hong}, {and} \bibinfo{person}{Michael Poland}.} \bibinfo{year}{2008}\natexlab{}.
\newblock \showarticletitle{A logical framework for behaviour reasoning and assistance in a smart home}.
\newblock \bibinfo{journal}{\emph{International Journal of Assistive Robotics and Mechatronics}} \bibinfo{volume}{9}, \bibinfo{number}{4} (\bibinfo{year}{2008}), \bibinfo{pages}{20--34}.
\newblock


\bibitem[Chen et~al\mbox{.}(2024)]%
        {chen2024towards}
\bibfield{author}{\bibinfo{person}{Xi Chen}, \bibinfo{person}{Julien Cumin}, \bibinfo{person}{Fano Ramparany}, {and} \bibinfo{person}{Dominique Vaufreydaz}.} \bibinfo{year}{2024}\natexlab{}.
\newblock \showarticletitle{Towards llm-powered ambient sensor based multi-person human activity recognition}. In \bibinfo{booktitle}{\emph{2024 IEEE 30th International Conference on Parallel and Distributed Systems (ICPADS)}}. IEEE, \bibinfo{pages}{609--616}.
\newblock


\bibitem[Civitarese(2017)]%
        {civitarese2017behavioral}
\bibfield{author}{\bibinfo{person}{Gabriele Civitarese}.} \bibinfo{year}{2017}\natexlab{}.
\newblock \showarticletitle{Behavioral monitoring in smart-home environments for health-care applications}. In \bibinfo{booktitle}{\emph{2017 IEEE International Conference on Pervasive Computing and Communications Workshops (PerCom Workshops)}}. IEEE, \bibinfo{pages}{105--106}.
\newblock


\bibitem[Civitarese et~al\mbox{.}(2024)]%
        {civitarese2024large}
\bibfield{author}{\bibinfo{person}{Gabriele Civitarese}, \bibinfo{person}{Michele Fiori}, \bibinfo{person}{Priyankar Choudhary}, {and} \bibinfo{person}{Claudio Bettini}.} \bibinfo{year}{2024}\natexlab{}.
\newblock \showarticletitle{Large language models are zero-shot recognizers for activities of daily living}.
\newblock \bibinfo{journal}{\emph{arXiv preprint arXiv:2407.01238}} (\bibinfo{year}{2024}).
\newblock


\bibitem[Cook et~al\mbox{.}(2012)]%
        {cook2012casas}
\bibfield{author}{\bibinfo{person}{Diane~J Cook}, \bibinfo{person}{Aaron~S Crandall}, \bibinfo{person}{Brian~L Thomas}, {and} \bibinfo{person}{Narayanan~C Krishnan}.} \bibinfo{year}{2012}\natexlab{}.
\newblock \showarticletitle{CASAS: A smart home in a box}.
\newblock \bibinfo{journal}{\emph{Computer}} \bibinfo{volume}{46}, \bibinfo{number}{7} (\bibinfo{year}{2012}), \bibinfo{pages}{62--69}.
\newblock


\bibitem[Davis et~al\mbox{.}(2016)]%
        {davis2016activity}
\bibfield{author}{\bibinfo{person}{Kadian Davis}, \bibinfo{person}{Evans Owusu}, \bibinfo{person}{Vahid Bastani}, \bibinfo{person}{Lucio Marcenaro}, \bibinfo{person}{Jun Hu}, \bibinfo{person}{Carlo Regazzoni}, {and} \bibinfo{person}{Loe Feijs}.} \bibinfo{year}{2016}\natexlab{}.
\newblock \showarticletitle{Activity recognition based on inertial sensors for ambient assisted living}. In \bibinfo{booktitle}{\emph{2016 19th international conference on information fusion (fusion)}}. Ieee, \bibinfo{pages}{371--378}.
\newblock


\bibitem[Demongivert et~al\mbox{.}(2021)]%
        {demongivert2021handling}
\bibfield{author}{\bibinfo{person}{C{\'e}dric Demongivert}, \bibinfo{person}{K{\'e}vin Bouchard}, \bibinfo{person}{S{\'e}bastien Gaboury}, \bibinfo{person}{Maxime Lussier}, \bibinfo{person}{Hubert Kenfack-Ngankam}, \bibinfo{person}{M{\'e}lanie Couture}, \bibinfo{person}{Nathalie Bier}, {and} \bibinfo{person}{Sylvain Giroux}.} \bibinfo{year}{2021}\natexlab{}.
\newblock \showarticletitle{Handling of labeling uncertainty in smart homes using generalizable fuzzy features}. In \bibinfo{booktitle}{\emph{Proceedings of the Conference on Information Technology for Social Good}}. \bibinfo{pages}{248--253}.
\newblock


\bibitem[Do et~al\mbox{.}(2018)]%
        {do2018rish}
\bibfield{author}{\bibinfo{person}{Ha~Manh Do}, \bibinfo{person}{Minh Pham}, \bibinfo{person}{Weihua Sheng}, \bibinfo{person}{Dan Yang}, {and} \bibinfo{person}{Meiqin Liu}.} \bibinfo{year}{2018}\natexlab{}.
\newblock \showarticletitle{RiSH: A robot-integrated smart home for elderly care}.
\newblock \bibinfo{journal}{\emph{Robotics and Autonomous Systems}}  \bibinfo{volume}{101} (\bibinfo{year}{2018}), \bibinfo{pages}{74--92}.
\newblock


\bibitem[Dosovitskiy et~al\mbox{.}(2020)]%
        {dosovitskiy2020image}
\bibfield{author}{\bibinfo{person}{Alexey Dosovitskiy}, \bibinfo{person}{Lucas Beyer}, \bibinfo{person}{Alexander Kolesnikov}, \bibinfo{person}{Dirk Weissenborn}, \bibinfo{person}{Xiaohua Zhai}, \bibinfo{person}{Thomas Unterthiner}, \bibinfo{person}{Mostafa Dehghani}, \bibinfo{person}{Matthias Minderer}, \bibinfo{person}{Georg Heigold}, \bibinfo{person}{Sylvain Gelly}, {et~al\mbox{.}}} \bibinfo{year}{2020}\natexlab{}.
\newblock \showarticletitle{An image is worth 16x16 words: Transformers for image recognition at scale}.
\newblock \bibinfo{journal}{\emph{arXiv preprint arXiv:2010.11929}} (\bibinfo{year}{2020}).
\newblock


\bibitem[Enshaeifar et~al\mbox{.}(2018)]%
        {enshaeifar2018health}
\bibfield{author}{\bibinfo{person}{Shirin Enshaeifar}, \bibinfo{person}{Ahmed Zoha}, \bibinfo{person}{Andreas Markides}, \bibinfo{person}{Severin Skillman}, \bibinfo{person}{Sahr~Thomas Acton}, \bibinfo{person}{Tarek Elsaleh}, \bibinfo{person}{Masoud Hassanpour}, \bibinfo{person}{Alireza Ahrabian}, \bibinfo{person}{Mark Kenny}, \bibinfo{person}{Stuart Klein}, {et~al\mbox{.}}} \bibinfo{year}{2018}\natexlab{}.
\newblock \showarticletitle{Health management and pattern analysis of daily living activities of people with dementia using in-home sensors and machine learning techniques}.
\newblock \bibinfo{journal}{\emph{PloS one}} \bibinfo{volume}{13}, \bibinfo{number}{5} (\bibinfo{year}{2018}), \bibinfo{pages}{e0195605}.
\newblock


\bibitem[Fritz et~al\mbox{.}(2022)]%
        {fritz2022nurse}
\bibfield{author}{\bibinfo{person}{Roschelle Fritz}, \bibinfo{person}{Katherine Wuestney}, \bibinfo{person}{Gordana Dermody}, {and} \bibinfo{person}{Diane~J Cook}.} \bibinfo{year}{2022}\natexlab{}.
\newblock \showarticletitle{Nurse-in-the-loop smart home detection of health events associated with diagnosed chronic conditions: A case-event series}.
\newblock \bibinfo{journal}{\emph{International Journal of Nursing Studies Advances}}  \bibinfo{volume}{4} (\bibinfo{year}{2022}), \bibinfo{pages}{100081}.
\newblock


\bibitem[Fuster(2017)]%
        {fuster2017changing}
\bibfield{author}{\bibinfo{person}{Valentin Fuster}.} \bibinfo{year}{2017}\natexlab{}.
\newblock \bibinfo{title}{Changing demographics: a new approach to global health care due to the aging population}.
\newblock \bibinfo{numpages}{3002--3005}~pages.
\newblock


\bibitem[Ghadi et~al\mbox{.}(2022)]%
        {ghadi2022improving}
\bibfield{author}{\bibinfo{person}{Yazeed~Yasin Ghadi}, \bibinfo{person}{Mouazma Batool}, \bibinfo{person}{Munkhjargal Gochoo}, \bibinfo{person}{Suliman~A Alsuhibany}, \bibinfo{person}{Tamara Al~Shloul}, \bibinfo{person}{Ahmad Jalal}, {and} \bibinfo{person}{Jeongmin Park}.} \bibinfo{year}{2022}\natexlab{}.
\newblock \showarticletitle{Improving the ambient intelligence living using deep learning classifier}.
\newblock \bibinfo{journal}{\emph{Comput. Mater. Contin}} \bibinfo{volume}{73}, \bibinfo{number}{1} (\bibinfo{year}{2022}), \bibinfo{pages}{1037--1053}.
\newblock


\bibitem[Ghayvat et~al\mbox{.}(2018)]%
        {ghayvat2018smart}
\bibfield{author}{\bibinfo{person}{Hemant Ghayvat}, \bibinfo{person}{S Mukhopadhyay}, \bibinfo{person}{B Shenjie}, \bibinfo{person}{Arpita Chouhan}, {and} \bibinfo{person}{W Chen}.} \bibinfo{year}{2018}\natexlab{}.
\newblock \showarticletitle{Smart home based ambient assisted living: Recognition of anomaly in the activity of daily living for an elderly living alone}. In \bibinfo{booktitle}{\emph{2018 IEEE international instrumentation and measurement technology conference (I2MTC)}}. IEEE, \bibinfo{pages}{1--5}.
\newblock


\bibitem[Ghods and Cook(2019)]%
        {ghods2019activity2vec}
\bibfield{author}{\bibinfo{person}{Alireza Ghods} {and} \bibinfo{person}{Diane~J Cook}.} \bibinfo{year}{2019}\natexlab{}.
\newblock \showarticletitle{Activity2vec: Learning adl embeddings from sensor data with a sequence-to-sequence model}.
\newblock \bibinfo{journal}{\emph{arXiv preprint arXiv:1907.05597}} (\bibinfo{year}{2019}).
\newblock


\bibitem[Gochoo et~al\mbox{.}(2018)]%
        {gochoo2018unobtrusive}
\bibfield{author}{\bibinfo{person}{Munkhjargal Gochoo}, \bibinfo{person}{Tan-Hsu Tan}, \bibinfo{person}{Shing-Hong Liu}, \bibinfo{person}{Fu-Rong Jean}, \bibinfo{person}{Fady~S Alnajjar}, {and} \bibinfo{person}{Shih-Chia Huang}.} \bibinfo{year}{2018}\natexlab{}.
\newblock \showarticletitle{Unobtrusive activity recognition of elderly people living alone using anonymous binary sensors and DCNN}.
\newblock \bibinfo{journal}{\emph{IEEE journal of biomedical and health informatics}} \bibinfo{volume}{23}, \bibinfo{number}{2} (\bibinfo{year}{2018}), \bibinfo{pages}{693--702}.
\newblock


\bibitem[Gu et~al\mbox{.}(2021)]%
        {gu2021survey}
\bibfield{author}{\bibinfo{person}{Fuqiang Gu}, \bibinfo{person}{Mu-Huan Chung}, \bibinfo{person}{Mark Chignell}, \bibinfo{person}{Shahrokh Valaee}, \bibinfo{person}{Baoding Zhou}, {and} \bibinfo{person}{Xue Liu}.} \bibinfo{year}{2021}\natexlab{}.
\newblock \showarticletitle{A survey on deep learning for human activity recognition}.
\newblock \bibinfo{journal}{\emph{ACM Computing Surveys (CSUR)}} \bibinfo{volume}{54}, \bibinfo{number}{8} (\bibinfo{year}{2021}), \bibinfo{pages}{1--34}.
\newblock


\bibitem[Guerra et~al\mbox{.}(2023)]%
        {guerra2023ambient}
\bibfield{author}{\bibinfo{person}{Bruna Maria~Vittoria Guerra}, \bibinfo{person}{Emanuele Torti}, \bibinfo{person}{Elisa Marenzi}, \bibinfo{person}{Micaela Schmid}, \bibinfo{person}{Stefano Ramat}, \bibinfo{person}{Francesco Leporati}, {and} \bibinfo{person}{Giovanni Danese}.} \bibinfo{year}{2023}\natexlab{}.
\newblock \showarticletitle{Ambient assisted living for frail people through human activity recognition: state-of-the-art, challenges and future directions}.
\newblock \bibinfo{journal}{\emph{Frontiers in neuroscience}}  \bibinfo{volume}{17} (\bibinfo{year}{2023}), \bibinfo{pages}{1256682}.
\newblock


\bibitem[Hamad et~al\mbox{.}(2019)]%
        {hamad2019efficient}
\bibfield{author}{\bibinfo{person}{Rebeen~Ali Hamad}, \bibinfo{person}{Alberto~Salguero Hidalgo}, \bibinfo{person}{Mohamed-Rafik Bouguelia}, \bibinfo{person}{Macarena~Espinilla Estevez}, {and} \bibinfo{person}{Javier~Medina Quero}.} \bibinfo{year}{2019}\natexlab{}.
\newblock \showarticletitle{Efficient activity recognition in smart homes using delayed fuzzy temporal windows on binary sensors}.
\newblock \bibinfo{journal}{\emph{IEEE journal of biomedical and health informatics}} \bibinfo{volume}{24}, \bibinfo{number}{2} (\bibinfo{year}{2019}), \bibinfo{pages}{387--395}.
\newblock


\bibitem[Hamad et~al\mbox{.}(2021)]%
        {hamad2021dilated}
\bibfield{author}{\bibinfo{person}{Rebeen~Ali Hamad}, \bibinfo{person}{Masashi Kimura}, \bibinfo{person}{Longzhi Yang}, \bibinfo{person}{Wai~Lok Woo}, {and} \bibinfo{person}{Bo Wei}.} \bibinfo{year}{2021}\natexlab{}.
\newblock \showarticletitle{Dilated causal convolution with multi-head self attention for sensor human activity recognition}.
\newblock \bibinfo{journal}{\emph{Neural Computing and Applications}}  \bibinfo{volume}{33} (\bibinfo{year}{2021}), \bibinfo{pages}{13705--13722}.
\newblock


\bibitem[Hamad et~al\mbox{.}(2020)]%
        {hamad2020joint}
\bibfield{author}{\bibinfo{person}{Rebeen~Ali Hamad}, \bibinfo{person}{Longzhi Yang}, \bibinfo{person}{Wai~Lok Woo}, {and} \bibinfo{person}{Bo Wei}.} \bibinfo{year}{2020}\natexlab{}.
\newblock \showarticletitle{Joint learning of temporal models to handle imbalanced data for human activity recognition}.
\newblock \bibinfo{journal}{\emph{Applied Sciences}} \bibinfo{volume}{10}, \bibinfo{number}{15} (\bibinfo{year}{2020}), \bibinfo{pages}{5293}.
\newblock


\bibitem[Hwang et~al\mbox{.}(2021)]%
        {hwang2021deep}
\bibfield{author}{\bibinfo{person}{Yu~Min Hwang}, \bibinfo{person}{Sangjun Park}, \bibinfo{person}{Hyung~Ok Lee}, \bibinfo{person}{Seok-Kap Ko}, {and} \bibinfo{person}{Byung-Tak Lee}.} \bibinfo{year}{2021}\natexlab{}.
\newblock \showarticletitle{Deep learning for human activity recognition based on causality feature extraction}.
\newblock \bibinfo{journal}{\emph{IEEE Access}}  \bibinfo{volume}{9} (\bibinfo{year}{2021}), \bibinfo{pages}{112257--112275}.
\newblock


\bibitem[Islam et~al\mbox{.}(2020)]%
        {islam2020development}
\bibfield{author}{\bibinfo{person}{Md~Milon Islam}, \bibinfo{person}{Ashikur Rahaman}, {and} \bibinfo{person}{Md~Rashedul Islam}.} \bibinfo{year}{2020}\natexlab{}.
\newblock \showarticletitle{Development of smart healthcare monitoring system in IoT environment}.
\newblock \bibinfo{journal}{\emph{SN computer science}}  \bibinfo{volume}{1} (\bibinfo{year}{2020}), \bibinfo{pages}{1--11}.
\newblock


\bibitem[Kanasi et~al\mbox{.}(2016)]%
        {kanasi2016aging}
\bibfield{author}{\bibinfo{person}{Eleni Kanasi}, \bibinfo{person}{Srinivas Ayilavarapu}, {and} \bibinfo{person}{Judith Jones}.} \bibinfo{year}{2016}\natexlab{}.
\newblock \showarticletitle{The aging population: demographics and the biology of aging}.
\newblock \bibinfo{journal}{\emph{Periodontology 2000}} \bibinfo{volume}{72}, \bibinfo{number}{1} (\bibinfo{year}{2016}), \bibinfo{pages}{13--18}.
\newblock


\bibitem[Kiranyaz et~al\mbox{.}(2021)]%
        {kiranyaz20211d}
\bibfield{author}{\bibinfo{person}{Serkan Kiranyaz}, \bibinfo{person}{Onur Avci}, \bibinfo{person}{Osama Abdeljaber}, \bibinfo{person}{Turker Ince}, \bibinfo{person}{Moncef Gabbouj}, {and} \bibinfo{person}{Daniel~J Inman}.} \bibinfo{year}{2021}\natexlab{}.
\newblock \showarticletitle{1D convolutional neural networks and applications: A survey}.
\newblock \bibinfo{journal}{\emph{Mechanical systems and signal processing}}  \bibinfo{volume}{151} (\bibinfo{year}{2021}), \bibinfo{pages}{107398}.
\newblock


\bibitem[Krishnan and Cook(2014)]%
        {krishnan2014activity}
\bibfield{author}{\bibinfo{person}{Narayanan~C Krishnan} {and} \bibinfo{person}{Diane~J Cook}.} \bibinfo{year}{2014}\natexlab{}.
\newblock \showarticletitle{Activity recognition on streaming sensor data}.
\newblock \bibinfo{journal}{\emph{Pervasive and mobile computing}}  \bibinfo{volume}{10} (\bibinfo{year}{2014}), \bibinfo{pages}{138--154}.
\newblock


\bibitem[Kwon et~al\mbox{.}(2019)]%
        {kwon2019handling}
\bibfield{author}{\bibinfo{person}{Hyeokhyen Kwon}, \bibinfo{person}{Gregory~D Abowd}, {and} \bibinfo{person}{Thomas Pl{\"o}tz}.} \bibinfo{year}{2019}\natexlab{}.
\newblock \showarticletitle{Handling annotation uncertainty in human activity recognition}. In \bibinfo{booktitle}{\emph{Proceedings of the 2019 ACM International Symposium on Wearable Computers}}. \bibinfo{pages}{109--117}.
\newblock


\bibitem[Lesani et~al\mbox{.}(2021)]%
        {lesani2021smart}
\bibfield{author}{\bibinfo{person}{Fatemeh~Sadat Lesani}, \bibinfo{person}{Faranak Fotouhi~Ghazvini}, {and} \bibinfo{person}{Hossein Amirkhani}.} \bibinfo{year}{2021}\natexlab{}.
\newblock \showarticletitle{Smart home resident identification based on behavioral patterns using ambient sensors}.
\newblock \bibinfo{journal}{\emph{Personal and Ubiquitous Computing}}  \bibinfo{volume}{25} (\bibinfo{year}{2021}), \bibinfo{pages}{151--162}.
\newblock


\bibitem[Li et~al\mbox{.}(2019)]%
        {li2019relation}
\bibfield{author}{\bibinfo{person}{Linjie Li}, \bibinfo{person}{Zhe Gan}, \bibinfo{person}{Yu Cheng}, {and} \bibinfo{person}{Jingjing Liu}.} \bibinfo{year}{2019}\natexlab{}.
\newblock \showarticletitle{Relation-aware graph attention network for visual question answering}. In \bibinfo{booktitle}{\emph{Proceedings of the IEEE/CVF international conference on computer vision}}. \bibinfo{pages}{10313--10322}.
\newblock


\bibitem[Liciotti et~al\mbox{.}(2020)]%
        {liciotti2020sequential}
\bibfield{author}{\bibinfo{person}{Daniele Liciotti}, \bibinfo{person}{Michele Bernardini}, \bibinfo{person}{Luca Romeo}, {and} \bibinfo{person}{Emanuele Frontoni}.} \bibinfo{year}{2020}\natexlab{}.
\newblock \showarticletitle{A sequential deep learning application for recognising human activities in smart homes}.
\newblock \bibinfo{journal}{\emph{Neurocomputing}}  \bibinfo{volume}{396} (\bibinfo{year}{2020}), \bibinfo{pages}{501--513}.
\newblock


\bibitem[Medina-Quero et~al\mbox{.}(2018)]%
        {medina2018ensemble}
\bibfield{author}{\bibinfo{person}{Javier Medina-Quero}, \bibinfo{person}{Shuai Zhang}, \bibinfo{person}{Chris Nugent}, {and} \bibinfo{person}{Macarena Espinilla}.} \bibinfo{year}{2018}\natexlab{}.
\newblock \showarticletitle{Ensemble classifier of long short-term memory with fuzzy temporal windows on binary sensors for activity recognition}.
\newblock \bibinfo{journal}{\emph{Expert Systems with Applications}}  \bibinfo{volume}{114} (\bibinfo{year}{2018}), \bibinfo{pages}{441--453}.
\newblock


\bibitem[Mohmed et~al\mbox{.}(2020)]%
        {mohmed2020employing}
\bibfield{author}{\bibinfo{person}{Gadelhag Mohmed}, \bibinfo{person}{Ahmad Lotfi}, {and} \bibinfo{person}{Amir Pourabdollah}.} \bibinfo{year}{2020}\natexlab{}.
\newblock \showarticletitle{Employing a deep convolutional neural network for human activity recognition based on binary ambient sensor data}. In \bibinfo{booktitle}{\emph{Proceedings of the 13th ACM international conference on pervasive technologies related to assistive environments}}. \bibinfo{pages}{1--7}.
\newblock


\bibitem[Mustafa et~al\mbox{.}(2021)]%
        {mustafa2021iot}
\bibfield{author}{\bibinfo{person}{Mustafa~A Mustafa}, \bibinfo{person}{Alexandros Konios}, {and} \bibinfo{person}{Matias Garcia-Constantino}.} \bibinfo{year}{2021}\natexlab{}.
\newblock \showarticletitle{IoT-based activities of daily living for abnormal behavior detection: Privacy issues and potential countermeasures}.
\newblock \bibinfo{journal}{\emph{IEEE Internet of Things Magazine}} \bibinfo{volume}{4}, \bibinfo{number}{3} (\bibinfo{year}{2021}), \bibinfo{pages}{90--95}.
\newblock


\bibitem[Najeh et~al\mbox{.}(2022)]%
        {najeh2022dynamic}
\bibfield{author}{\bibinfo{person}{Houda Najeh}, \bibinfo{person}{Christophe Lohr}, {and} \bibinfo{person}{Benoit Leduc}.} \bibinfo{year}{2022}\natexlab{}.
\newblock \showarticletitle{Dynamic segmentation of sensor events for real-time human activity recognition in a smart home context}.
\newblock \bibinfo{journal}{\emph{Sensors}} \bibinfo{volume}{22}, \bibinfo{number}{14} (\bibinfo{year}{2022}), \bibinfo{pages}{5458}.
\newblock


\bibitem[Nweke et~al\mbox{.}(2018)]%
        {nweke2018deep}
\bibfield{author}{\bibinfo{person}{Henry~Friday Nweke}, \bibinfo{person}{Ying~Wah Teh}, \bibinfo{person}{Mohammed~Ali Al-Garadi}, {and} \bibinfo{person}{Uzoma~Rita Alo}.} \bibinfo{year}{2018}\natexlab{}.
\newblock \showarticletitle{Deep learning algorithms for human activity recognition using mobile and wearable sensor networks: State of the art and research challenges}.
\newblock \bibinfo{journal}{\emph{Expert Systems with Applications}}  \bibinfo{volume}{105} (\bibinfo{year}{2018}), \bibinfo{pages}{233--261}.
\newblock


\bibitem[Oguntala et~al\mbox{.}(2021)]%
        {oguntala2021passive}
\bibfield{author}{\bibinfo{person}{George~A Oguntala}, \bibinfo{person}{Yim~Fun Hu}, \bibinfo{person}{Ali~AS Alabdullah}, \bibinfo{person}{Raed~A Abd-Alhameed}, \bibinfo{person}{Muhammad Ali}, {and} \bibinfo{person}{Doanh~K Luong}.} \bibinfo{year}{2021}\natexlab{}.
\newblock \showarticletitle{Passive RFID module with LSTM recurrent neural network activity classification algorithm for ambient-assisted living}.
\newblock \bibinfo{journal}{\emph{IEEE Internet of Things Journal}} \bibinfo{volume}{8}, \bibinfo{number}{13} (\bibinfo{year}{2021}), \bibinfo{pages}{10953--10962}.
\newblock


\bibitem[Park et~al\mbox{.}(2018)]%
        {park2018deep}
\bibfield{author}{\bibinfo{person}{Jiho Park}, \bibinfo{person}{Kiyoung Jang}, {and} \bibinfo{person}{Sung-Bong Yang}.} \bibinfo{year}{2018}\natexlab{}.
\newblock \showarticletitle{Deep neural networks for activity recognition with multi-sensor data in a smart home}. In \bibinfo{booktitle}{\emph{2018 IEEE 4th World Forum on Internet of Things (WF-IoT)}}. IEEE, \bibinfo{pages}{155--160}.
\newblock


\bibitem[Pl{\"o}tz et~al\mbox{.}(2024)]%
        {plotz2024using}
\bibfield{author}{\bibinfo{person}{Thomas Pl{\"o}tz} {et~al\mbox{.}}} \bibinfo{year}{2024}\natexlab{}.
\newblock \showarticletitle{Using Graphs to Perform Effective Sensor-Based Human Activity Recognition in Smart Homes}.
\newblock \bibinfo{journal}{\emph{Sensors}} \bibinfo{volume}{24}, \bibinfo{number}{12} (\bibinfo{year}{2024}), \bibinfo{pages}{3944}.
\newblock


\bibitem[Quigley et~al\mbox{.}(2018)]%
        {quigley2018comparative}
\bibfield{author}{\bibinfo{person}{Bronagh Quigley}, \bibinfo{person}{Mark Donnelly}, \bibinfo{person}{George Moore}, {and} \bibinfo{person}{Leo Galway}.} \bibinfo{year}{2018}\natexlab{}.
\newblock \showarticletitle{A comparative analysis of windowing approaches in dense sensing environments}. In \bibinfo{booktitle}{\emph{Proceedings}}, Vol.~\bibinfo{volume}{2}. MDPI, \bibinfo{pages}{1245}.
\newblock


\bibitem[Ranieri et~al\mbox{.}(2021)]%
        {ranieri2021activity}
\bibfield{author}{\bibinfo{person}{Caetano~Mazzoni Ranieri}, \bibinfo{person}{Scott MacLeod}, \bibinfo{person}{Mauro Dragone}, \bibinfo{person}{Patricia~Amancio Vargas}, {and} \bibinfo{person}{Roseli Aparecida~Francelin Romero}.} \bibinfo{year}{2021}\natexlab{}.
\newblock \showarticletitle{Activity recognition for ambient assisted living with videos, inertial units and ambient sensors}.
\newblock \bibinfo{journal}{\emph{Sensors}} \bibinfo{volume}{21}, \bibinfo{number}{3} (\bibinfo{year}{2021}), \bibinfo{pages}{768}.
\newblock


\bibitem[Riboni et~al\mbox{.}(2016)]%
        {riboni2016smartfaber}
\bibfield{author}{\bibinfo{person}{Daniele Riboni}, \bibinfo{person}{Claudio Bettini}, \bibinfo{person}{Gabriele Civitarese}, \bibinfo{person}{Zaffar~Haider Janjua}, {and} \bibinfo{person}{Rim Helaoui}.} \bibinfo{year}{2016}\natexlab{}.
\newblock \showarticletitle{SmartFABER: Recognizing fine-grained abnormal behaviors for early detection of mild cognitive impairment}.
\newblock \bibinfo{journal}{\emph{Artificial intelligence in medicine}}  \bibinfo{volume}{67} (\bibinfo{year}{2016}), \bibinfo{pages}{57--74}.
\newblock


\bibitem[Schrader et~al\mbox{.}(2020)]%
        {schrader2020advanced}
\bibfield{author}{\bibinfo{person}{Lisa Schrader}, \bibinfo{person}{Agust{\'\i}n Vargas~Toro}, \bibinfo{person}{Sebastian Konietzny}, \bibinfo{person}{Stefan R{\"u}ping}, \bibinfo{person}{Barbara Sch{\"a}pers}, \bibinfo{person}{Martina Steinb{\"o}ck}, \bibinfo{person}{Carmen Krewer}, \bibinfo{person}{Friedemann M{\"u}ller}, \bibinfo{person}{J{\"o}rg G{\"u}ttler}, {and} \bibinfo{person}{Thomas Bock}.} \bibinfo{year}{2020}\natexlab{}.
\newblock \showarticletitle{Advanced sensing and human activity recognition in early intervention and rehabilitation of elderly people}.
\newblock \bibinfo{journal}{\emph{Journal of Population Ageing}}  \bibinfo{volume}{13} (\bibinfo{year}{2020}), \bibinfo{pages}{139--165}.
\newblock


\bibitem[Singh et~al\mbox{.}(2017)]%
        {singh2017human}
\bibfield{author}{\bibinfo{person}{Deepika Singh}, \bibinfo{person}{Erinc Merdivan}, \bibinfo{person}{Ismini Psychoula}, \bibinfo{person}{Johannes Kropf}, \bibinfo{person}{Sten Hanke}, \bibinfo{person}{Matthieu Geist}, {and} \bibinfo{person}{Andreas Holzinger}.} \bibinfo{year}{2017}\natexlab{}.
\newblock \showarticletitle{Human activity recognition using recurrent neural networks}. In \bibinfo{booktitle}{\emph{Machine Learning and Knowledge Extraction: First IFIP TC 5, WG 8.4, 8.9, 12.9 International Cross-Domain Conference, CD-MAKE 2017, Reggio, Italy, August 29--September 1, 2017, Proceedings 1}}. Springer, \bibinfo{pages}{267--274}.
\newblock


\bibitem[Tan et~al\mbox{.}(2018)]%
        {tan2018multi}
\bibfield{author}{\bibinfo{person}{Tan-Hsu Tan}, \bibinfo{person}{Munkhjargal Gochoo}, \bibinfo{person}{Shih-Chia Huang}, \bibinfo{person}{Yi-Hung Liu}, \bibinfo{person}{Shing-Hong Liu}, {and} \bibinfo{person}{Yun-Fa Huang}.} \bibinfo{year}{2018}\natexlab{}.
\newblock \showarticletitle{Multi-resident activity recognition in a smart home using RGB activity image and DCNN}.
\newblock \bibinfo{journal}{\emph{IEEE Sensors Journal}} \bibinfo{volume}{18}, \bibinfo{number}{23} (\bibinfo{year}{2018}), \bibinfo{pages}{9718--9727}.
\newblock


\bibitem[Tekemetieu et~al\mbox{.}(2021)]%
        {tekemetieu2021context}
\bibfield{author}{\bibinfo{person}{Armel~Ayimdji Tekemetieu}, \bibinfo{person}{Corentin Haidon}, \bibinfo{person}{Fr{\'e}d{\'e}ric Bergeron}, \bibinfo{person}{Hubert~Kengfack Ngankam}, \bibinfo{person}{H{\'e}l{\`e}ne Pigot}, \bibinfo{person}{Charles Gouin-Vallerand}, {and} \bibinfo{person}{Sylvain Giroux}.} \bibinfo{year}{2021}\natexlab{}.
\newblock \showarticletitle{Context modelling in ambient assisted living: Trends and lessons}.
\newblock \bibinfo{journal}{\emph{Internet of Things: Cases and Studies}} (\bibinfo{year}{2021}), \bibinfo{pages}{189--225}.
\newblock


\bibitem[Tewell et~al\mbox{.}(2019)]%
        {tewell2019monitoring}
\bibfield{author}{\bibinfo{person}{Jordan Tewell}, \bibinfo{person}{Dympna O’Sullivan}, \bibinfo{person}{Neil Maiden}, \bibinfo{person}{James Lockerbie}, {and} \bibinfo{person}{Simone Stumpf}.} \bibinfo{year}{2019}\natexlab{}.
\newblock \showarticletitle{Monitoring meaningful activities using small low-cost devices in a smart home}.
\newblock \bibinfo{journal}{\emph{Personal and Ubiquitous Computing}}  \bibinfo{volume}{23} (\bibinfo{year}{2019}), \bibinfo{pages}{339--357}.
\newblock


\bibitem[Thukral et~al\mbox{.}(2025)]%
        {thukral2025layout}
\bibfield{author}{\bibinfo{person}{Megha Thukral}, \bibinfo{person}{Sourish~Gunesh Dhekane}, \bibinfo{person}{Shruthi~K Hiremath}, \bibinfo{person}{Harish Haresamudram}, {and} \bibinfo{person}{Thomas Ploetz}.} \bibinfo{year}{2025}\natexlab{}.
\newblock \showarticletitle{Layout-Agnostic Human Activity Recognition in Smart Homes through Textual Descriptions Of Sensor Triggers (TDOST)}.
\newblock \bibinfo{journal}{\emph{Proceedings of the ACM on Interactive, Mobile, Wearable and Ubiquitous Technologies}} \bibinfo{volume}{9}, \bibinfo{number}{1} (\bibinfo{year}{2025}), \bibinfo{pages}{1--38}.
\newblock


\bibitem[Uddin et~al\mbox{.}(2018)]%
        {uddin2018ambient}
\bibfield{author}{\bibinfo{person}{Md~Zia Uddin}, \bibinfo{person}{Weria Khaksar}, {and} \bibinfo{person}{Jim Torresen}.} \bibinfo{year}{2018}\natexlab{}.
\newblock \showarticletitle{Ambient sensors for elderly care and independent living: a survey}.
\newblock \bibinfo{journal}{\emph{Sensors}} \bibinfo{volume}{18}, \bibinfo{number}{7} (\bibinfo{year}{2018}), \bibinfo{pages}{2027}.
\newblock


\bibitem[Vaswani et~al\mbox{.}(2017)]%
        {vaswani2017attention}
\bibfield{author}{\bibinfo{person}{Ashish Vaswani}, \bibinfo{person}{Noam Shazeer}, \bibinfo{person}{Niki Parmar}, \bibinfo{person}{Jakob Uszkoreit}, \bibinfo{person}{Llion Jones}, \bibinfo{person}{Aidan~N Gomez}, \bibinfo{person}{{\L}ukasz Kaiser}, {and} \bibinfo{person}{Illia Polosukhin}.} \bibinfo{year}{2017}\natexlab{}.
\newblock \showarticletitle{Attention is all you need}.
\newblock \bibinfo{journal}{\emph{Advances in neural information processing systems}}  \bibinfo{volume}{30} (\bibinfo{year}{2017}).
\newblock


\bibitem[Xia et~al\mbox{.}(2020)]%
        {xia2020lstm}
\bibfield{author}{\bibinfo{person}{Kun Xia}, \bibinfo{person}{Jianguang Huang}, {and} \bibinfo{person}{Hanyu Wang}.} \bibinfo{year}{2020}\natexlab{}.
\newblock \showarticletitle{LSTM-CNN architecture for human activity recognition}.
\newblock \bibinfo{journal}{\emph{Ieee Access}}  \bibinfo{volume}{8} (\bibinfo{year}{2020}), \bibinfo{pages}{56855--56866}.
\newblock


\bibitem[Yala et~al\mbox{.}(2015)]%
        {yala2015feature}
\bibfield{author}{\bibinfo{person}{Nawel Yala}, \bibinfo{person}{Belkacem Fergani}, {and} \bibinfo{person}{Anthony Fleury}.} \bibinfo{year}{2015}\natexlab{}.
\newblock \showarticletitle{Feature extraction for human activity recognition on streaming data}. In \bibinfo{booktitle}{\emph{2015 International symposium on innovations in intelligent systems and applications (INISTA)}}. IEEE, \bibinfo{pages}{1--6}.
\newblock


\bibitem[Yamada et~al\mbox{.}(2007)]%
        {yamada2007applying}
\bibfield{author}{\bibinfo{person}{Naoharu Yamada}, \bibinfo{person}{Kenji Sakamoto}, \bibinfo{person}{Goro Kunito}, \bibinfo{person}{Yoshinori Isoda}, \bibinfo{person}{Kenichi Yamazaki}, {and} \bibinfo{person}{Satoshi Tanaka}.} \bibinfo{year}{2007}\natexlab{}.
\newblock \showarticletitle{Applying ontology and probabilistic model to human activity recognition from surrounding things}.
\newblock \bibinfo{journal}{\emph{IPSJ Digital Courier}}  \bibinfo{volume}{3} (\bibinfo{year}{2007}), \bibinfo{pages}{506--517}.
\newblock


\bibitem[Yan et~al\mbox{.}(2016)]%
        {yan2016real}
\bibfield{author}{\bibinfo{person}{Surong Yan}, \bibinfo{person}{Yixing Liao}, \bibinfo{person}{Xiaoqing Feng}, {and} \bibinfo{person}{Yanan Liu}.} \bibinfo{year}{2016}\natexlab{}.
\newblock \showarticletitle{Real time activity recognition on streaming sensor data for smart environments}. In \bibinfo{booktitle}{\emph{2016 International Conference on Progress in Informatics and Computing (PIC)}}. IEEE, \bibinfo{pages}{51--55}.
\newblock


\bibitem[Ye et~al\mbox{.}(2023)]%
        {ye2023graph}
\bibfield{author}{\bibinfo{person}{Jiancong Ye}, \bibinfo{person}{Hongjie Jiang}, {and} \bibinfo{person}{Junpei Zhong}.} \bibinfo{year}{2023}\natexlab{}.
\newblock \showarticletitle{A graph-attention-based method for single-resident daily activity recognition in smart homes}.
\newblock \bibinfo{journal}{\emph{Sensors}} \bibinfo{volume}{23}, \bibinfo{number}{3} (\bibinfo{year}{2023}), \bibinfo{pages}{1626}.
\newblock


\bibitem[Zakka et~al\mbox{.}(2024)]%
        {zakka2024action}
\bibfield{author}{\bibinfo{person}{Vincent~Gbouna Zakka}, \bibinfo{person}{Zhuangzhuang Dai}, {and} \bibinfo{person}{Luis~J Manso}.} \bibinfo{year}{2024}\natexlab{}.
\newblock \showarticletitle{Action recognition for privacy-preserving ambient assisted living}. In \bibinfo{booktitle}{\emph{International Conference on AI in Healthcare}}. Springer, \bibinfo{pages}{203--217}.
\newblock


\bibitem[Zhang et~al\mbox{.}(2021)]%
        {zhang2021elderly}
\bibfield{author}{\bibinfo{person}{Xinyu Zhang}, \bibinfo{person}{Qammer~H Abbasi}, \bibinfo{person}{Francesco Fioranelli}, \bibinfo{person}{Olivier Romain}, {and} \bibinfo{person}{Julien Le~Kernec}.} \bibinfo{year}{2021}\natexlab{}.
\newblock \showarticletitle{Elderly care-human activity recognition using radar with an open dataset and hybrid maps}. In \bibinfo{booktitle}{\emph{EAI International Conference on Body Area Networks}}. Springer, \bibinfo{pages}{39--51}.
\newblock


\bibitem[Zin et~al\mbox{.}(2021)]%
        {zin2021real}
\bibfield{author}{\bibinfo{person}{Thi~Thi Zin}, \bibinfo{person}{Ye Htet}, \bibinfo{person}{Yuya Akagi}, \bibinfo{person}{Hiroki Tamura}, \bibinfo{person}{Kazuhiro Kondo}, \bibinfo{person}{Sanae Araki}, {and} \bibinfo{person}{Etsuo Chosa}.} \bibinfo{year}{2021}\natexlab{}.
\newblock \showarticletitle{Real-time action recognition system for elderly people using stereo depth camera}.
\newblock \bibinfo{journal}{\emph{Sensors}} \bibinfo{volume}{21}, \bibinfo{number}{17} (\bibinfo{year}{2021}), \bibinfo{pages}{5895}.
\newblock


\end{thebibliography}

\end{document}